\documentclass[format=acmsmall, screen=false, review=false, authordraft=false]{acmart}

\setcopyright{acmcopyright}
\copyrightyear{2018}
\acmYear{2018}
\acmDOI{XXXXXXX.XXXXXXX}

\acmJournal{JACM}
\acmVolume{37}
\acmNumber{4}
\acmArticle{111}
\acmMonth{8}

\usepackage[linesnumbered,ruled,noend]{algorithm2e}
\usepackage{bm}
\usepackage{multirow}
\usepackage{subfigure}
\usepackage{siunitx}
\usepackage{textcomp}
\usepackage{wrapfig}
\newcommand{\std}[1]{\scriptsize{±{#1}}}

\begin{document}

\title{Graph Neural Rough Differential Equations for Traffic Forecasting}


\author{Jeongwhan Choi}
\affiliation{%
  \institution{Yonsei University}
  \streetaddress{50 Yonsei-ro}
  \city{Seoul}
  \country{South Korea}}
\email{jeongwhan.choi@yonsei.ac.kr}

\author{Noseong Park}
\affiliation{%
  \institution{Yonsei University}
  \streetaddress{50 Yonsei-ro}
  \city{Seoul}
  \country{South Korea}}
\email{noseong@yonsei.ac.kr}

\renewcommand{\shortauthors}{Choi et al.}

\begin{abstract}
  Traffic forecasting is one of the most popular spatio-temporal tasks in the field of machine learning. A prevalent approach in the field is to combine graph convolutional networks and recurrent neural networks for the spatio-temporal processing. There has been fierce competition and many novel methods have been proposed. In this paper, we present the method of spatio-temporal graph neural rough differential equation (STG-NRDE). Neural rough differential equations (NRDEs) are a breakthrough concept for processing time-series data. Their main concept is to use the log-signature transform to convert a time-series sample into a relatively shorter series of feature vectors. We extend the concept and design two NRDEs: one for the temporal processing and the other for the spatial processing. After that, we combine them into a single framework. We conduct experiments with 6 benchmark datasets and 27 baselines. STG-NRDE shows the best accuracy in all cases, outperforming all those 27 baselines by non-trivial margins.
\end{abstract}




\maketitle

\section{Introduction}
The spatio-temporal graph data frequently happens in real-world applications, ranging from traffic to climate forecasting~\cite{zaytar2016sequence,shi2015convolutional,shi2017deep,liu2016application,hwang2021climate, racah2016extremeweather,kurth2018exascale,cheng2018ensemble,cheng2018neural,hossain2015forecasting,ren2021deep, tekin2021spatio,li2018dcrnn_traffic,bing2018stgcn,wu2019graphwavenet,guo2019astgcn,bai2019STG2Seq,song2020stsgcn,huang2020lsgcn,NEURIPS2020_ce1aad92,li2021stfgnn,chen2021ZGCNET,fang2021STODE,choi2023climate}. For instance, the traffic forecasting task launched by California Performance of Transportation (PeMS) is one of the most popular problems in the area of spatio-temporal processing~\cite{chen2001freeway,bing2018stgcn,guo2019astgcn}.

Given a time-series of graphs $\{\mathcal{G}_{t_i} \stackrel{\text{def}}{=} (\mathcal{V},\mathcal{E},\bm{F}_{i}, t_i)\}_{i=0}^{N}$, where $\mathcal{V}$ is a fixed set of nodes, $\mathcal{E}$ is a fixed set of edges, $t_i$ is a time-point when $\mathcal{G}_{t_i}$ is observed, and $\bm{F}_{i} \in \mathbb{R}^{|\mathcal{V}| \times D}$ is a feature matrix at time $t_i$ which contains $D$-dimensional input features of the nodes, the spatio-temporal forecasting is to predict $\hat{\bm{Y}} \in \mathbb{R}^{|\mathcal{V}| \times S  \times M}$, e.g., predicting the traffic volume for each location of a road network for the next $S$ time-points (or horizons) given past $N+1$ historical traffic patterns, where $|\mathcal{V}|$ is the number of locations to predict and $M=1$ because the volume is a scalar, i.e., the number of vehicles. We note that $\mathcal{V}$ and $\mathcal{E}$ do not change over time --- in other words, the graph topology is fixed --- whereas the node input features can change over time. We use upper boldface to denote matrices and lower boldface for vectors.

\begin{figure}[!t]
    \centering
    \includegraphics[width=0.6\textwidth]{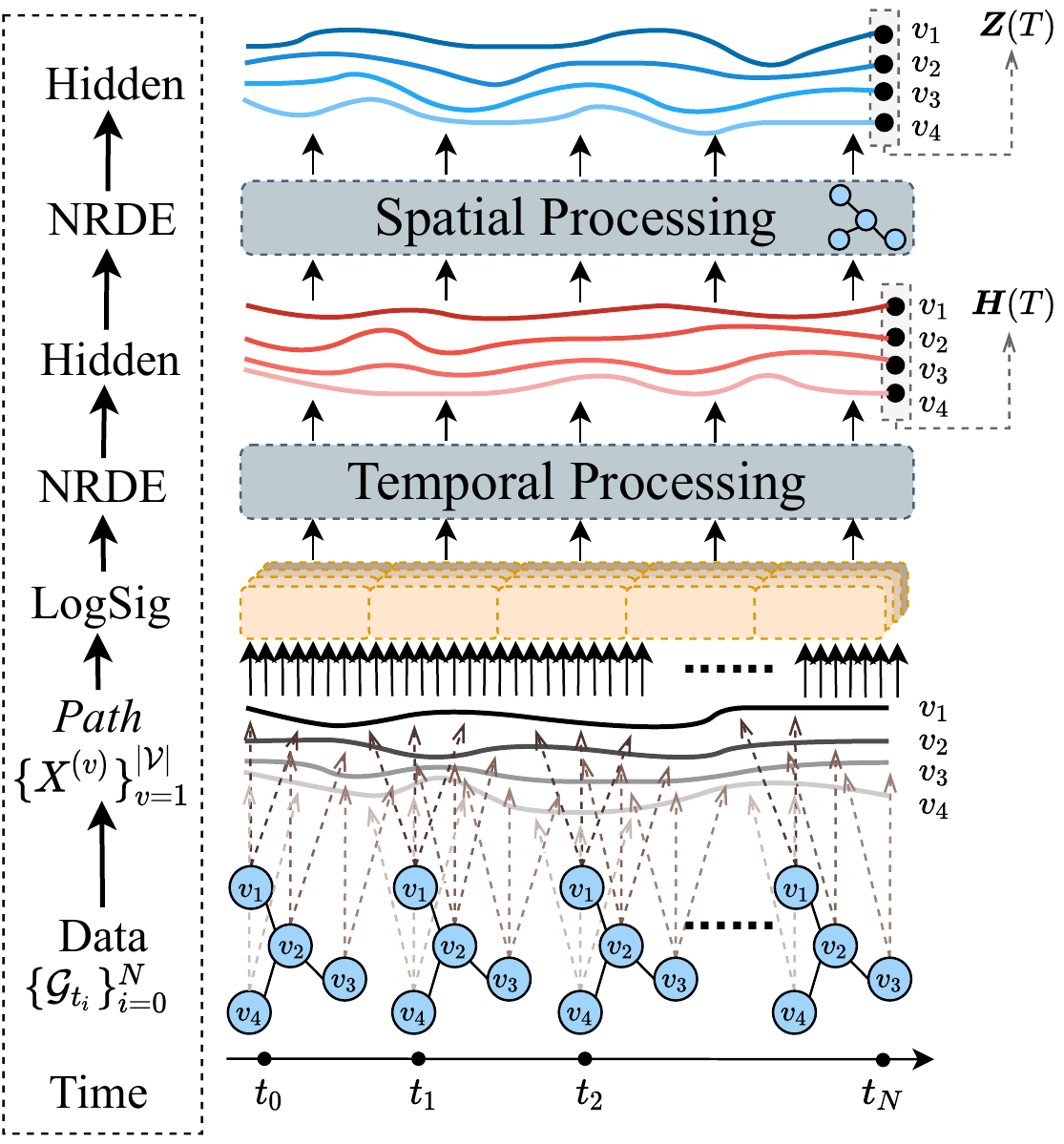}
    \caption{The overall workflow in our proposed STG-NRDE}
    \label{fig:stgnrde}
\end{figure}

% NRDEs are based on the rough path theory which was established to make sense of the controlled differential equation:
% \begin{align}
%     d\mathbf{z}(t) = f(\mathbf{z}(t))dX(t),
% \end{align}where $X$ is a continuous control path, and $\mathbf{z}(t)$ is a hidden vector at time $t$. A prevalent example of $X$ is a (semimartingale) Wiener process coupled with time (i.e. $dX_t = d(t, W_t))$, in which case the equation reduces to a stochastic differential equation. In this sense, the rough path theory extends stochastic differential equations beyond the semimartingale environments~\cite{lyons2004differential}.

\begin{figure}[t]
    \centering
    \includegraphics[width=0.7\textwidth]{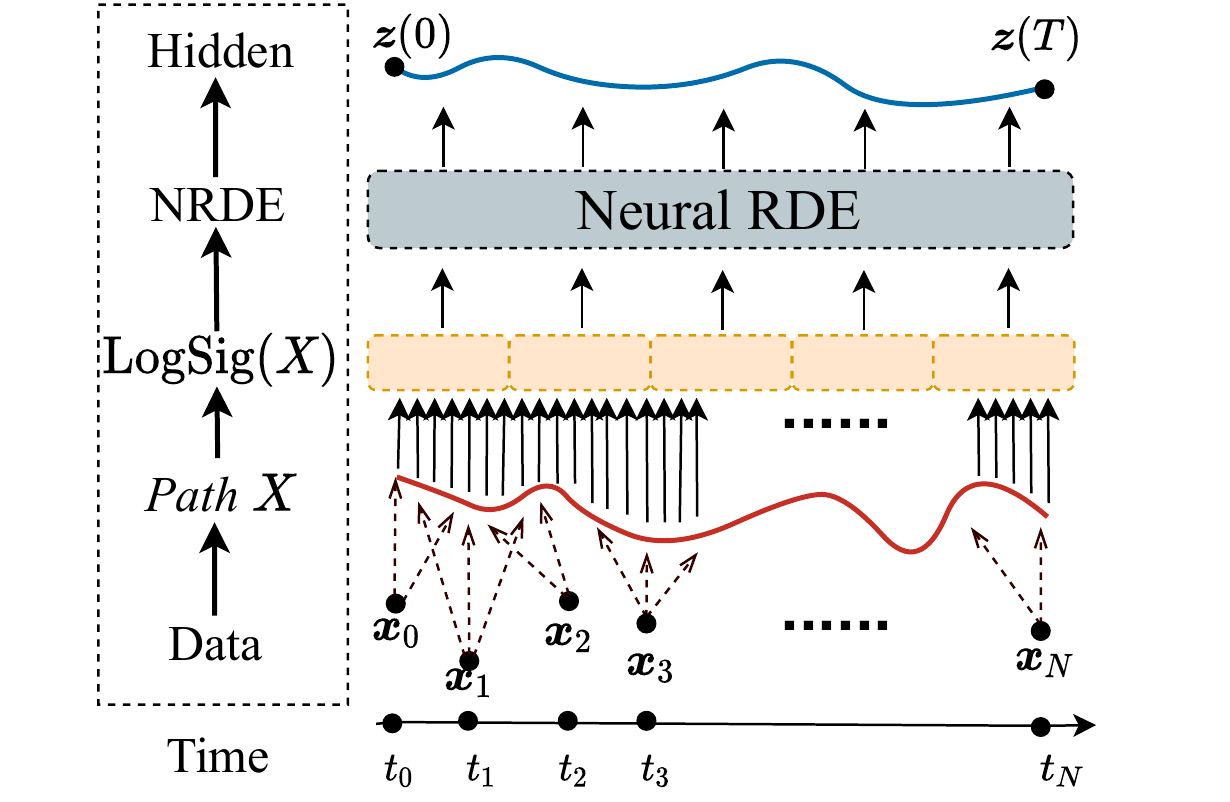}
    \caption{The overall workflow of the original NRDE for processing time-series. The path $X$ is created from $\{(t_i, \bm{x}_{i})\}_{i=0}^N$ by an interpolation algorithm and another time-series of log-signature is created. At the end, the time-series of log-signature is processed for a downstream task.}
    \label{fig:nrde}
\end{figure}

For this task, a diverse set of techniques have been proposed. In this paper, however, we design a method based on neural rough differential equations (NRDEs) --- to our knowledge, we are the first proposing NRDE-based spatio-temporal models. NRDEs are based on the rough path theory designed to make sense of the controlled differential equation. NRDEs, which are considered as a continuous analogue to recurrent neural networks (RNNs), can be written as follows:
\begin{align}\label{eq:nrde}
\bm{z}(T) &= \bm{z}(0) + \int_{0}^{T} f(\bm{z}(t);\bm{\theta}_f) dX(t)\\&= \bm{z}(0) + \int_{0}^{T} f(\bm{z}(t);\bm{\theta}_f) \frac{dX(t)}{dt} dt,\label{eq:nrde2}
\end{align}where $X$ is a continuous path taking values in a Banach space. The entire trajectory of $\bm{z}(t)$ is controlled over time by the path $X$ (cf. Fig.~\ref{fig:nrde}). Learning the RDE function $f$ for a downstream task is a key point in NRDEs.

% A prevalent example of $X$ is a (semimartingale) Wiener process coupled with time (i.e. $dX_t = d(t, W_t))$, in which case the equation reduces to a stochastic differential equation.

% The theory of the controlled differential equation (CDE) had been developed to extend the stochastic differential equation and the It\^{o} calculus far beyond the semimartingale setting of $X$ --- in other words, Eq.~\eqref{eq:nrde} reduces to the stochastic differential equation if and only if $X$ meets the semimartingale requirement. 
Equation~\ref{eq:nrde} reduces to the stochastic differential equation if and only if $X$ meets the semimartingale requirement. For instance, a prevalent example of the path $X$ is a Wiener process in the case of the stochastic differential equation. In this sense, the rough path theory extends stochastic differential equations beyond the semimartingale environments~\cite{lyons2004differential}. 

%%%%%%%%%%%%%%%%%%%%%%%
% In RDEs, however, the path $X$ does not need to be such semimartingale or martingale processes. NODEs are a technology to parameterize such RDEs and learn from data. In addition, Eq.~\eqref{eq:nrde2} continuously reads the values $\frac{dX(t)}{dt}$ and integrates them over time. In this regard, NODEs are equivalent to continuous RNNs and show the state-of-the-art accuracy in many time-series tasks and data.
%%%%%%%%%%%%%%%%%%%%%%%

The \emph{log-signature} of a path is an essential concept in rough path theory. It had been demonstrated that under moderate conditions~\cite{lyons2018inverting,geng2017reconstruction}, the log-signature of a path with bounded variations is unique and most time-series data that happens in the field of deep learning has bounded variations. As a result, the log-signature can be interpreted as a unique feature of the path.

However, no research has yet been done on how to integrate NRDE (temporal processing) and graph convolutional processing (spatial processing). In order to solve the spatio-temporal forecasting problem, we combine them into a single framework.

NRDEs construct a continuous path $X(t)$, where $t \in [0,T]$, with an interpolation algorithm, where $X(t_i) = (\bm{x}_i, t_i)$ for $t_i\in \{t_i\}_{i=0}^N$. In other words, the path has the same value as the observation $(\bm{x}_i, t_i)$, when $t_i$ is one of the observation time-points and otherwise, interpolated values. As shown in Fig.~\ref{fig:nrde}, for each $P$-length sub-path, a log-signature (each dotted orange box in the figure) is calculated, and a new time-series of log-signature, denoted  $\{\text{LogSig}^D_{r_i,r_{i+1}}(X)\}_{i=0}^{\lfloor \frac{T}{P} \rfloor-1}$, is formed. The number and dimensionality of log-signatures are determined by the sub-path length $P$ and the depth $D$, respectively.

In the original setting of NRDEs, there exists a single time-series, denoted $\{(t_i, \bm{x}_{i})\}_{i=0}^N$, where $\bm{x}_{i} \in \mathbb{R}^D$ is a $D$-dimensional vector and $t_i$ is a time-point when $\bm{x}_{i}$ is observed. In our case, however, there are $|\mathcal{V}|$ different time-series patterns to consider, each of which is connected with nearby time-series patterns in some way. The differences between them are shown in Figs.~\ref{fig:stgnrde} and~\ref{fig:nrde}.

In our method, the pre-processing step creates a continuous path $X^{(v)}$ for each node $v \in \mathcal{V}$. We use the same method as in the original NRDE design for this. Given a discrete time-series $\{\bm{x}_{i}\}_{i=0}^N$, the original NRDE runs an interpolation algorithm to build its continuous path and then calculates the log-signature for every sub-path. We repeat the process for each node, resulting in a set of paths, denoted $\{X^{(v)}\}_{v=1}^{|\mathcal{V}|}$, and their log-signatures.

The most important stage is to apply a spatial and temporal processing approach to $\{X^{(v)}\}_{v=1}^{|\mathcal{V}|}$ while taking into acount its its graph connectivity. We create an NRDE model with a graph processing technique for both spatial and temporal processing in our example. The last hidden vector $\bm{z}^{(v)}(T)$ for each node $v$ is then calculated, and the final output layer is used to to predict $\hat{\bm{Y}} \in \mathbb{R}^{|\mathcal{V}| \times S \times M}$.

We use 6 benchmark datasets collected by California Performance of Transportation (PeMS) and 27 baseline methods. In terms of three conventional assessment metrics, our proposed method clearly beats all existing baselines. As a result, we can summarize our contributions as follows:
\begin{enumerate}
    \item We design a single framework by combining two NRDEs for learning the temporal and spatial dependencies of traffic conditions.
    % \item NRDEs are robust to irregular time-series by the design. Owing to this characteristic, our method is also robust to the irregularity of the temporal sequence, i.e., some observations can be missing. 
    \item Our large-scale experiments with 6 datasets and 27 baselines clearly show the efficacy of the proposed method. The accuracy averaged over the 6 datasets is summarized in Table~\ref{tab:average}.
    \item Our method is also robust to the irregularity of the temporal sequence, i.e., some observations can be missing. We perform irregular traffic forecasting to reflect real-world environments where sensing values can be missing (see Table~\ref{tbl:irregular}).
    \item One can download our codes and datasets from this url\footnote{\url{https://github.com/jeongwhanchoi/STG-NRDE}}.
\end{enumerate}

\begin{table}[!t]
\small
    \centering
    \caption{The average error of some selected highly performing models across all the six datasets. Inside the parentheses, we show their performance relative to our method.}
    \label{tab:average}
    \begin{tabular}{c ccc}
    \toprule
        Model                &   MAE                 &   RMSE                &  MAPE\\\midrule
        STGCN                & 14.88 (117.7\%)       & 24.22 (114.1\%)       & 12.30 (124.4\%)\\
        DCRNN                & 14.90 (117.9\%)       & 24.04 (113.2\%)       & 12.75 (128.8\%)\\
        GraphWaveNet         & 15.94 (126.1\%)       & 26.22 (123.5\%)       & 12.96 (131.0\%) \\
        ASTGCN(r)            & 14.86 (117.5\%)       & 23.95 (112.7\%)       & 12.25 (124.7\%) \\
        STSGCN               & 14.45 (114.3\%)       & 23.58 (111.1\%)       & 11.42 (115.5\%) \\
        AGCRN                & 13.32 (105.3\%)       & 22.29 (104.9\%)       & 10.37 (105.5\%) \\
        STFGNN               & 13.92 (110.1\%)       & 22.57 (106.2\%)       & 11.30 (115.0\%) \\
        STGODE               & 13.56 (107.2\%)       & 22.37 (105.2\%)       & 10.77 (109.6\%) \\
        Z-GCNETs             & 13.22 (104.5\%)       & 21.92 (103.1\%)       & 10.44 (106.4\%) \\
        DSTAGNN              & 12.92 (102.1\%)       & 21.63 (101.7\%)       & 10.20 (103.8\%) \\
        STG-NCDE             & 12.72 (100.5\%)       & 21.33 (100.4\%)       & 10.10 (102.8\%)\\\midrule
        \textbf{STG-NRDE}    & \textbf{12.65} (100.0\%) & \textbf{21.25} (100.0\%) & \textbf{9.83} (100.0\%)\\
    \bottomrule
    \end{tabular}
\end{table}

%%%%%%%%%%%%%%%%%%%%%%%%%%%%%%%%%%%%%%%%%%%%%%%%%%%%
%%%%%%%%%%%%%%%%%%%%  SECTION   %%%%%%%%%%%%%%%%%%%%
%%%%%%%%%%%%%%%%%%%%%%%%%%%%%%%%%%%%%%%%%%%%%%%%%%%%
\section{Related Work and Preliminaries}
\noindent In this section, we summarize our literature review related to our NRDE-based spatio-temporal forecasting.
%%%%%%%%%%%%%%%%%%%% SUBSECTION %%%%%%%%%%%%%%%%%%%%
\subsection{Neural Ordinary Differential Equations (NODEs)}
Prior to NRDEs, neural ordinary differential equations (NODEs) shown how to use differential equations to continuously model residual neural networks (ResNets). NODE can be written in the following way:
\begin{align}\label{eq:node}
\bm{z}(T) = \bm{z}(0) + \int_{0}^{T} f(\bm{z}(t), t;\bm{\theta}_f) dt;
\end{align}where the neural network parameterized by $\bm{\theta}_f$ approximates $\frac{\bm{z}(t)}{dt}$. To solve the integral problem, we use a variety of ODE solvers, from the explicit Euler method to the 4th order Runge—Kutta (RK4) method and the Dormand—Prince (DOPRI) method ~\cite{DORMAND198019}.

When using the explicit Euler method to solve Equation~\ref{eq:node}, it reduces to the residual connection. NODEs generalize ResNets in a continuous manner in this regard. This NODE technology is used by STGODE to tackle the spatio-temporal forecasting problem~\cite{fang2021STODE}.

%%%%%%%%%%%%%%%%%%%% SUBSECTION %%%%%%%%%%%%%%%%%%%%
\subsection{Neural Controlled Differential Equations (NCDEs)}
NCDEs in Equation~\ref{eq:nrde} generalize RNNs in a continuous manner, whereas NODEs generalize ResNets. Controlled differential equations (CDEs) are technically more complicated than ordinary differential equations (ODEs) --- Equation~\ref{eq:nrde2} uses the Riemann--Stieltjes integral problem whereas Equation~\ref{eq:node} uses the Rienmann integral. The original CDE formulation in Equation~\ref{eq:nrde} reduces Equation~\ref{eq:nrde2} where $\frac{\bm{z}(t)}{dt}$ is approximated by $f(\bm{z}(t);\bm{\theta}_f) \frac{dX(t)}{dt}$.

We can solve Equation~\ref{eq:nrde2} using existing ODE solvers once $\frac{\bm{z}(t)}{dt}$ is properly formulated. STG-NCDE, which is our previous work, uses two NCDEs for solving spatio-temporal forecasting problems~\cite{choi2022STGNCDE}.

%%%%%%%%%%%%%%%%%%%% SUBSECTION %%%%%%%%%%%%%%%%%%%%
\subsection{Signature Transform}
Let $[r_{i},r_{i+1}]$ be a range of time, where $r_{i} < r_{i+1}$. Then, the signature of a time-series sample in the range is defined as follows:
\begin{align}\begin{split}\label{eq:sigtrans}
S^{i_1,\dots,i_k}_{r_i,r_{i+1}}(X) = &\idotsint\limits_{r_i<t_1<\ldots<t_k<r_{i+1}}  \prod_{j=1}^{k} \frac{dX^{i_j}}{dt}(t_j)dt_j,\\
\text{Sig}^{D}_{r_i,r_{i+1}}(X) = &\Big(
\{S^{i}_{r_i,r_{i+1}}(X)^{(i)}\}_{i=1}^{\dim(X)}, \{S^{i,j}_{r_i,r_{i+1}}(X)\}_{i,j=1}^{\dim(X)}\\
&,\dots,\{S^{i_1,\ldots,i_D}_{r_i,r_{i+1}}(X)\}_{i_1,\ldots,i_D=1}^{\dim(X)}\Big).
\end{split}\end{align}

The signature, on the other hand, has redundancy, e.g., $S^{1,2}_{a,b}(X) + S^{2,1}_{a,b}(X) = S^{1}_{a,b}(X) S^{2}_{a,b}(X)$, where we can know a value if we know three others. After removing the signature's redundancy, the log-signature $\text{LogSig}^D_{r_i,r_{i+1}}(X)$ is created.

%%%%%%%%%%%%%%%%%%%% SUBSECTION %%%%%%%%%%%%%%%%%%%%
\subsection{Neural Rough Differential Equations (NRDEs)}

NRDEs~\cite{morrill2021neuralrough} are defined as follows due to the log-signature transform of time-series:
% \begin{align}\begin{split}\label{eq:nrdes}
% h(X, t) &= \frac{LogSig^D_{r_i,r_{i+1}}(X)}{r_{i+1} - r_i}\textrm{ for } t \in [r_i, r_{i+1}),\\
% \bm{z}(T) &= \bm{z}(0) + \int_{0}^{T} f(\bm{z}(t);\bm{\theta}_f) h(X,t) dt,
% \end{split}\end{align}
\begin{align}\begin{split}\label{eq:nrdes}
\bm{z}(T) &= \bm{z}(0) + \int_{0}^{T} f(\bm{z}(t);\bm{\theta}_f) \frac{\text{LogSig}^D_{r_i,r_{i+1}}(X)}{r_{i+1} - r_i} dt, \textrm{ for } t \in [r_i, r_{i+1}).
\end{split}\end{align} 
The log-signature created by the path  $X$ inside the interval $[r_i, r_{i+1})$ is referred to as $\text{LogSig}^D_{r_i,r_{i+1}}(X)$. $\frac{\text{LogSig}^D_{r_i,r_{i+1}}(X)}{r_{i+1} - r_i}$ is a piecewise approximation of the time-derivative of the log-signature in the short interval $[r_i, r_{i+1})$. $D$ means the depth of the log-signature. Once we define the sub-path length $P$, the intervals $\{[r_i, r_{i+1})\}_{i=0}^{\lfloor \frac{T}{P} \rfloor -1}$ are decided, i.e., $P = r_{i+1} - r_i$ for all $i$. Then, the time-series of $\text{LogSig}^D_{r_i, r_{i+1}}(X)$ constitutes $\{\text{LogSig}^D_{r_i,r_{i+1}}(X)\}_{i=0}^{\lfloor \frac{T}{P} \rfloor -1}$.

Since it continuously reads the time-derivative of the log-signature, NDREs can be thought of as a continuous analogue to RNNs because they use Equation~\ref{eq:nrdes} to deduce $\bm{z}(T)$ from $\bm{z}(0)$. As a result, the initial value z(0) and the sequence of the time-derivative of the log-signature determine $\bm{z}(T)$. The adjoint sensitivity approach can also be used to train NRDEs.

%%%%%%%%%%%%%%%%%%%% SUBSECTION %%%%%%%%%%%%%%%%%%%%
\subsection{Traffic Forecasting}
In spatio-temporal machine learning, the problem of traffic forecasting has been studied extensively for a long time~\cite{lu2022more,li2022sta3d,Zhang2022cSTML,lu2022GSeqAtt}. It has non-trivial effects on our daily life when it is handled correctly. The majority of existing methods, such as auto-regressive integrated moving average (ARIMA)~\cite{van1996combining}, vector autoregressive moving average, and Gaussian process, use a statistical approach. On the other hand, a statistical method forecasts each type of road traffic separately, ignoring the impact of spatial relationships on traffic conditions. 

Graph Neural Networks (GNNs) are a class of deep learning models that can learn from graph-structured data, such as road networks. GNNs have been successfully applied to model spatial dependencies in various domains ranging from social networks to recommender systems~\cite{kipf2017GCN,hamilton2017graphSAGE,liu2019influence,hwang2021climate,do22twostage,choi2023climate,choi2023bspm,choi2023gread}. For traffic forecasting, Gra-TF~\cite{zhang2021graph} proposed the fusion framework of various GNNs and traditional predictor to capture spatial dependencies, but did not consider temporal dependency. Unlike Gra-TF, our proposed method is a single framework that does not require a fusion framework. The temporal dependency is significant in traffic forecasting, but both dependencies were not considered. As a result, in the last four years, academics have opted to model the road network as a graph in order to capture spatio-temporal relationships and improve traffic forecasting accuracy. 

We introduce a number of seminal publications for considering spatio-temporal dependencies. DCRNN~\cite{li2018dcrnn_traffic} combined graph convolution with recurrent neural networks in an encoder-decoder manner. STGCN~\cite{bing2018stgcn} combined graph convolution with a 1D convolution. GraphWaveNet~\cite{wu2019graphwavenet} combined adaptive graph convolution with dilated casual convolution to capture spatial-temporal dependencies. ASTGCN~\cite{guo2019astgcn} proposed the spatial-temporal attention mechanism with the spatial-temporal convolution. STG2Seq~\cite{bai2019STG2Seq} used a multiple gated graph convolutional module and a seq2seq architecture with an attention mechanism to make a multi-step prediction. STSGCN~\cite{song2020stsgcn} used multiple localized spatial-temporal subgraph modules to synchronously capture the localized spatial-temporal correlations directly. LSGCN~\cite{huang2020lsgcn} proposed the framework integrated a novel attention mechanism and graph convolution into a spatial gated block. AGCRN~\cite{NEURIPS2020_ce1aad92} used node adaptive parameter learning to capture node-specific spatial and temporal correlations in time-series data automatically without a pre-defined graph. STFGNN~\cite{li2021stfgnn} proposed a novel fusion operation of various spatial and temporal graphs to capture hidden spatial dependencies treated for different time periods in parallel. Z-GCNETs~\cite{chen2021ZGCNET} proposed the new time-aware zigzag topological-based spatio-temporal model, which is integrated with a time-aware zigzag topological layer into time-conditioned GCNs. {Some studies consider the topology of graphs, such as Z-GCNets, but models that process time-evolution graphs have recently been studied. In particular, LRGCN~\cite{Li2019LRGNN} proposed a framework to handle time-evolving graphs in which the graph's structure changes over time. CoEvolve-GNN~\cite{Wang2021CoEvoGNN} also proposed a framework for processing dynamic property graph sequences. STG-NRDE uses the learned adjacency matrix to learn the edges according to the node features of each graph snapshot, and the structure can be changed. However, our study does not focus on time evolution graphs. In particular, it does not deal with dynamic graphs in which the number of nodes varies. ST-GDN~\cite{zhang2021traffic} used a resolution-aware self-attention network to encode multi-level temporal signals. ST-GDN integrates the local and global traffic dependencies across different regions to enhance the spatial-temporal pattern representations. Our proposed method can improve the temporal and spatial dependencies because it uses the hidden vector evolved with the time derivative amount of the path transformed into a log-signature. In addition, the advantage of our method is that irregular forecasting is possible because NRDEs can handle missing values. DSTGNN~\cite{huang2022dynamical} creates a spatial graph based on the stability of nodes to address the uncertainty of spatial dependency. But our method does not need the pre-defined spatial graph, e.g., adjacency matrix. We adopt the adaptive graph convolution to learn adjacency matrix. STGODE~\cite{fang2021STODE} proposed a tensor-based ODE for capturing spatial-temporal dynamics. STG-ODE requires semantic and pre-defined adjacency matrices, but STG-NRDE uses a learned adjacency matrix and does not require those adjacency matrices. And in the case of STG-ODE, temporal dependency is captured using TCN, but we use it based on NRDE. Recently STG-NCDE~\cite{choi2022STGNCDE} proposed the single framework with two NCDEs for traffic forecasting and shows the state-of-the-art performance.

%%%%%%%%%%%%%%%%%%%%%%%%%%%%%%%%%%%%%%%%%%%%%%%%%%%%
%%%%%%%%%%%%%%%%%%%%  SECTION   %%%%%%%%%%%%%%%%%%%%
%%%%%%%%%%%%%%%%%%%%%%%%%%%%%%%%%%%%%%%%%%%%%%%%%%%%
\section{Proposed Method}
\noindent A time-series of graphs $\{\mathcal{G}_{t_i} \stackrel{\text{def}}{=} (\mathcal{V},\mathcal{E},\bm{F}_{i})\}_{i=0}^{N}$ requires more complex spatio-temporal processing than spatial (i.e., GCNs) or temporal processing (i.e., RNNs). As such, there have been proposed many neural networks combining GCNs and RNNs. Based on the NRDE and adaptive topology generating technologies, this research proposes a unique spatio-temporal model. In this section, we explain our proposed method. We start with its overall design and then move on to details.

%%%%%%%%%%%%%%%%%%%% SUBSECTION %%%%%%%%%%%%%%%%%%%%
\subsection{Overall Design}
Our method includes one pre-processing and one main processing step as follows:
\begin{enumerate}
    \item Its pre-processing step is to create a continuous path $X^{(v)}$ for each node $v$, where $1 \leq v \leq |\mathcal{V}|$, from $\{\bm{F}_{i}^{(v)}\}_{i=0}^{N}$. $\bm{F}_{i}^{(v)} \in \mathbb{R}^D$ means the $v$-th row of $\bm{F}_{i}$, and $\{\bm{F}_{i}^{(v)}\}$ stands for the time-series of the input features of $v$. 
    We use an interpolation method on the discrete time-series $\{\bm{F}_{i}^{(v)}\}$ and building a continuous path. Then, its log-signatures, created by the log-signature transform, contain unique features.
    \item The above pre-processing step occurs prior to training our model. The final hidden vector for each node $v$, denoted $\bm{z}^{(v)}(T)$, is then calculated by our main step, which combines GCN and NRDE technologies.
    \item After that, we have an output layer to predict $\hat{\bm{y}}^{(v)} \in \mathbb{R}^{S \times M}$ for each node $v$. We have the prediction matrix $\hat{\bm{Y}} \in \mathbb{R}^{|\mathcal{V}| \times S \times M}$ after collecting those predictions for all nodes in $\mathcal{V}$.
\end{enumerate}

%%%%%%%%%%%%%%%%%%%% SUBSECTION %%%%%%%%%%%%%%%%%%%%
\subsection{Graph Neural Rough Differential Equations}
The proposed spatio-temporal graph neural rough differential equation (STG-NRDE) consists of two NRDEs: one for temporal processing and one for spatial processing.

\subsubsection{Temporal Processing} The first NRDE for the temporal processing can be written as follows:
\begin{align}
\bm{h}^{(v)}(T) &= \bm{h}^{(v)}(0) + \int_{0}^{T} f(\bm{h}^{(v)}(t);\bm{\theta}_f) \frac{\text{LogSig}^D_{r_i,r_{i+1}}(X)}{r_{i+1} - r_i} dt, \label{eq:type1}
\end{align} where $\bm{h}^{(v)}(t)$ is a hidden trajectory (over time $t \in [0,T]$) of the temporal information of node $v$. After stacking $\bm{h}^{(v)}(t)$ for all $v$, we can define a matrix $\bm{H}(t) \in \mathbb{R}^{|\mathcal{V}| \times \dim(\bm{h}^{(v)})}$. Therefore, the trajectory created by $\bm{H}(t)$ over time $t$ contains the hidden information of the temporal processing results. Equation~\ref{eq:type1} can be equivalently rewritten as follows using the matrix notation:
\begin{align}
\bm{H}(T) &= \bm{H}(0) + \int_{0}^{T} f(\bm{H}(t);\bm{\theta}_f) \frac{\text{LogSig}^D_{r_i,r_{i+1}}(\bm{X})}{r_{i+1} - r_i} dt, \label{eq:type1-2}
\end{align}where $\bm{X}(t)$ is a matrix whose $v$-th row is $X^{(v)}$. The RDE function $f$ separately processes each row in $\bm{H}(t)$. The key in this design is how to define the RDE function $f$ parameterized by $\bm{\theta}_f$. We will describe shortly how to define it. One good thing is that $f$ does not need to be a RNN. By designing it with fully-connected layers only, for instance, Equation~\ref{eq:type1-2} converts it to a \emph{continuous} RNN. 

\subsubsection{Spatial Processing} After that, the second NRDE starts for its spatial processing as follows:
\begin{align}\begin{split}
\bm{Z}(T) = \bm{Z}(0) + \int_{0}^{T} g(\bm{Z}(t);\bm{\theta}_g) \frac{\text{LogSig}^D_{r_i,r_{i+1}}(\bm{H})}{r_{i+1} - r_i} dt,\label{eq:type2}
\end{split}\end{align}where the hidden trajectory $\bm{Z}(t)$ is controlled by $\bm{H}(t)$ which is created by the temporal processing.

After combining Equations~\ref{eq:type1-2} and~\ref{eq:type2}, we have the following single equation which incorporates both the temporal and the spatial processing:
\begin{align}\begin{split}
\bm{Z}(T) &= \bm{Z}(0) + \int_{0}^{T} g(\bm{Z}(t);\bm{\theta}_g)f(\bm{H}(t);\bm{\theta}_f) \frac{\text{LogSig}^D_{r_i,r_{i+1}}(\bm{X})}{r_{i+1} - r_i} dt , \label{eq:type2-2}
\end{split}\end{align}where $\bm{Z}(t) \in \mathbb{R}^{|\mathcal{V}| \times \dim(\bm{z}^{(v)})}$ is a matrix created after stacking the hidden trajectory $\bm{z}^{(v)}$ for all $v$. In this NRDE, a hidden trajectory $\bm{z}^{(v)}$ is created after considering the trajectories of its neighbors --- for ease of writing, we use the matrix notation in Equations~\ref{eq:type2} and~\ref{eq:type2-2}. The key part is how to design the RDE function $g$ parameterized by $\bm{\theta}_g$ for the spatial processing.

\subsubsection{RDE Functions} We now describe the two RDE functions $f$ and $g$. The definition of $f:\mathbb{R}^{|\mathcal{V}| \times \dim(\bm{h}^{(v)})} \rightarrow \mathbb{R}^{|\mathcal{V}| \times \dim(\bm{h}^{(v)})}$ is as as follows:
\begin{align}\begin{split}
f(\bm{H}(t);\bm{\theta}_f) &= \psi(\texttt{FC}_{|\mathcal{V}| \times \dim(\bm{h}^{(v)}) \rightarrow |\mathcal{V}| \times \dim(\bm{h}^{(v)})}(\bm{A}_{K})),\\
&\vdots\\
\bm{A}_1 &= \sigma(\texttt{FC}_{|\mathcal{V}| \times \dim(\bm{h}^{(v)}) \rightarrow |\mathcal{V}| \times \dim(\bm{h}^{(v)})}(\bm{A}_0)),\\
\bm{A}_0 &= \sigma(\texttt{FC}_{|\mathcal{V}| \times \dim(\bm{h}^{(v)}) \rightarrow |\mathcal{V}| \times \dim(\bm{h}^{(v)})}(\bm{H}(t))),\label{eq:fun_f}
\end{split}\end{align}
where $\sigma$ is a rectified linear unit, $\psi$ is a hyperbolic tangent, and $\mathtt{FC}_{input\_size \rightarrow output\_size}$ means a fully-connected layer whose input size is $input\_size$ and output size is also $output\_size$. $\bm{\theta}_f$ refers to the parameters of the fully-connected layers. This function $f$ independently processes each row of $\bm{H}(t)$ with the $K$ fully connected-layers.

For the spatial processing, we need to define one more RDE function $g$. The definition of $g:\mathbb{R}^{|\mathcal{V}| \times \dim(\bm{z}^{(v)})} \rightarrow \mathbb{R}^{|\mathcal{V}| \times \dim(\bm{z}^{(v)})}$ is as follows:
\begin{align}
g(\bm{Z}(t);\bm{\theta}_g) &= \psi(\texttt{FC}_{|\mathcal{V}| \times \dim(\bm{z}^{(v)}) \rightarrow |\mathcal{V}| \times \dim(\bm{z}^{(v)})}(\bm{B}_1)),\label{eq:fun_g1}\\
\bm{B}_1 &= (\bm{I} + \phi(\sigma(\bm{E}\cdot\bm{E}^{\intercal})))\bm{B}_0\bm{W}_{spatial},\label{eq:fun_g2}\\
\bm{B}_0 &= \sigma(\texttt{FC}_{|\mathcal{V}| \times \dim(\bm{z}^{(v)}) \rightarrow |\mathcal{V}| \times \dim(\bm{z}^{(v)})}(\bm{Z}(t))),\label{eq:fun_g3}
\end{align}
where $\bm{I}$ is the $|\mathcal{V}| \times |\mathcal{V}|$ identity matrix, $\phi$ is a softmax activation, $\bm{E} \in \mathbb{R}^{|\mathcal{V}| \times C} $ is a trainable node-embedding matrix, $\bm{E}^{\intercal}$ is its transpose, and $\bm{W}_{spatial}$ is a trainable weight transformation matrix. Conceptually, $\phi(\sigma(\bm{E}\cdot\bm{E}^{\intercal}))$ corresponds to the normalized adjacency matrix $\bm{D}^{-\frac{1}{2}}\bm{A}\bm{D}^{-\frac{1}{2}}$, where $\bm{A} = \sigma(\bm{E}\cdot\bm{E}^{\intercal})$ and the softmax activation plays a role of normalizing the adaptive adjacency matrix ~\cite{wu2019graphwavenet,NEURIPS2020_ce1aad92}. We also note that Equation~\ref{eq:fun_g2} is identical to the first order Chebyshev polynomial expansion of the graph convolution operation~\cite{kipf2017GCN} with the normalized adaptive adjacency matrix. Equation~\ref{eq:fun_g1} and~\ref{eq:fun_g3} do not mix the rows of their input matrices $\bm{Z}(t)$ and $\bm{B}_1$. It is Equation~\ref{eq:fun_g2} where the rows of $\bm{B}_0$ are mixed for the spatial processing.

\subsubsection{Initial Value Generation} The initial value of the temporal processing, i.e., $\bm{H}(0)$, is created from $\bm{F}_{t_0}$ as follows: $\bm{H}(0) = FC_{D \rightarrow \dim(\bm{h}^{(v)})}(\bm{F}_{t_0})$. We also use the following similar strategy to generate $\bm{Z}(0)$: $\bm{Z}(0) = FC_{\dim(\bm{h}^{(v)}) \rightarrow \dim(\bm{z}^{(v)})}(\bm{H}(0))$. After generating these initial values for the two NRDEs, we can calculate $\bm{Z}(T)$ after solving the Riemann--Stieltjes integral problem in Equation~\ref{eq:type2-2}.

%%%%%%%%%%%%%%%%%%%% SUBSECTION %%%%%%%%%%%%%%%%%%%%
\subsection{How to Train}
To implement Equation~\ref{eq:type2-2} --- we do not separately implement Equations~\ref{eq:type1-2} and~\ref{eq:type2} --- we define the following augmented ODE:
\begin{align}
\frac{d}{dt}{\begin{bmatrix}
  \bm{Z}(t) \\
  \bm{H}(t) \\
  \end{bmatrix}\!} = {\begin{bmatrix}
  g(\bm{Z}(t);\bm{\theta}_g)f(\bm{H}(t);\bm{\theta}_f) \frac{\text{LogSig}^D_{r_i,r_{i+1}}(X)}{r_{i+1} - r_i} \\
  f(\bm{H}(t);\bm{\theta}_f) \frac{\text{LogSig}^D_{r_i,r_{i+1}}(X)}{r_{i+1} - r_i}\\
  \end{bmatrix}\!},
\end{align}where the initial values $\bm{Z}(0)$ and $\bm{H}(0)$ are generated in the aforementioned ways. We then train the parameters of the initial value generation layer, the RDE functions, including the node-embedding matrix $\bm{E}$, and the output layer. From $\bm{z}^{(v)}(T)$, i.e., the $v$-th row of $\bm{Z}(T)$, the following output layer produces $\hat{\bm{y}}^{(v)}$:
\begin{align}
    \hat{\bm{y}}^{(v)} = \bm{z}^{(v)}(T)\bm{W}_{output} + \bm{b}_{output}, \label{eq:output}
\end{align}where $\bm{W}_{output} \in \mathbb{R}^{\dim(\bm{z}^{(v)}(T)) \rightarrow S \times M}$ and $\bm{b}_{output} \in \mathbb{R}^{S \times M}$ are a trainable weight and a bias of the output layer. We use the following $L^1$ loss as the training objective, which is defined as:
\begin{align}
\mathcal{L} = \frac{\sum_{\tau \in \mathcal{T}}\sum_{v \in \mathcal{V}} \|\bm{y}^{(\tau,v)} - \hat{\bm{y}}^{(\tau,v)}\|_1}{|\mathcal{V}| \times |\mathcal{T}|}, \label{eq:loss}
\end{align}where $\mathcal{T}$ is a training set, $\tau$ is a training sample, and $\bm{y}^{(\tau,v)}$ is the ground-truth of node $v$ in $\tau$. We also use the standard $L^2$ regularization of the parameters, i.e., weight decay.

The well-posedness\footnote{A well-posed problem means i) its solution uniquely exists, and ii) its solution continuously changes as input data changes.} of NRDEs was already proved in \cite[Theorem 1.3]{lyons2007differential} under the mild condition of the Lipschitz continuity. We show that our NRDE layers are also well-posed problems. Almost all activations, such as ReLU, Leaky ReLU, SoftPlus, Tanh, Sigmoid, ArcTan, and Softsign, have a Lipschitz constant of 1. Other common neural network layers, such as dropout, batch normalization and other pooling methods, have explicit Lipschitz constant values. Therefore, the Lipschitz continuity of $f$ and $g$ can be fulfilled in our case. Therefore, it is a well-posed training problem. Therefore, our training algorithm solves a well-posed problem so its training process is stable in practice.

%%%%%%%%%%%%%%%%%%%%%%%%%%%%%%%%%%%%%%%%%%%%%%%%%%%%
%%%%%%%%%%%%%%%%%%%%  SECTION   %%%%%%%%%%%%%%%%%%%%
%%%%%%%%%%%%%%%%%%%%%%%%%%%%%%%%%%%%%%%%%%%%%%%%%%%%
\section{Experiments}
\noindent We describe our experimental environments and results. We conduct experiments with time-series forecasting. Our software and hardware environments are as follows: \textsc{Ubuntu} 18.04 LTS, \textsc{Python} 3.9.5, \textsc{Numpy} 1.20.3, \textsc{Scipy} 1.7, \textsc{Matplotlib} 3.3.1, \textsc{torchdiffeq} 0.2.2, \textsc{PyTorch} 1.9.0, \textsc{CUDA} 11.4, and \textsc{NVIDIA} Driver 470.42, i9 CPU, and \textsc{NVIDIA RTX A6000}. We use 6 datasets and 27 baseline models, which is one of the largest scale experiments in the field of traffic forecasting.
%%%%%%%%%%%%%%%%%%%% SUBSECTION %%%%%%%%%%%%%%%%%%%%
\subsection{Datasets}
In the experiment, we use six real-world traffic datasets, namely PeMSD7(M), PeMSD7(L), PeMS03, PeMS04, PeMS07, and PeMS08, which were collected by California Performance of Transportation (PeMS)~\cite{chen2001freeway} in real-time every 30 second and widely used in the previous studies~\cite{bing2018stgcn,guo2019astgcn,fang2021STODE,chen2021ZGCNET,song2020stsgcn}. The traffic datasets are aggregated to 5 minutes. The node in the traffic network is represented by the loop detector, which can detect real-time traffic conditions, and the edge is a freeway segment between the two nearest nodes. More details of the datasets are in Table~\ref{tab:dataset}. We note that they contain different types of values: i) the number of vehicles, or ii) velocity.

\begin{table}[t]
    \setlength{\tabcolsep}{2pt}
    \centering
    % \small
    \caption{The summary of the datasets used in our work. We predict either traffic volume (i.e., \# of vehicles) or velocity.}
    \label{tab:dataset}
    \begin{tabular}{ccccc}
    \hline
        Dataset     & $|\mathcal{V}|$  & Time Steps& Time Range & Type \\ \hline
        PeMSD3      & 358       & 26,208    & 09/2018 - 11/2018 & Volume \\
        PeMSD4      & 307       & 16,992    & 01/2018 - 02/2018 & Volume \\
        PeMSD7      & 883       & 28,224    & 05/2017 - 08/2017 & Volume \\ 
        PeMSD8      & 170       & 17,856    & 07/2016 - 08/2016 & Volume \\ 
        PeMSD7(M)   & 228       & 12,672    & 05/2012 - 06/2012 & Velocity\\
        PeMSD7(L)   & 1,026     & 12,672    & 05/2012 - 06/2012 & Velocity\\
    \hline
    \end{tabular}
\end{table}

%%%%%%%%%%%%%%%%%%%% SUBSECTION %%%%%%%%%%%%%%%%%%%%
\subsection{Experimental Settings}
The datasets are already split into training, validation, and testing sets in a 6:2:2 ratio. The time interval between two consecutive time-points in these datasets is 5 minutes. All existing papers, including our paper, use the forecasting settings of $S=12$ and $M=1$ after reading past 12 graph snapshots, i.e., $N=11$ --- note that the graph snapshot index $i$ starts from $0$. In short, we conduct 12-sequence-to-12-sequence forecasting, which is the standard benchmark-setting in this domain.

%%%%%%%%%%%%%%%%%%%% SUBSUBSECTION
\subsubsection{Evaluation Metrics}
We use three widely used metrics to measure the performance of different models, i.e., mean absolute error (MAE),  mean absolute percentage error (MAPE), and root mean squared error (RMSE), which are defined as:
\begin{align}
\text{MAE} &=\frac{1}{|\mathcal{V}| \times |\mathcal{R}|}\sum_{\gamma \in \mathcal{R}}\sum_{v \in \mathcal{V}} |\bm{y}^{(\gamma,v)} - \hat{\bm{y}}^{(\gamma,v)}|,\\
\text{RMSE} &=\sqrt{\frac{1}{|\mathcal{V}| \times |\mathcal{R}|}\sum_{\gamma \in \mathcal{R}}\sum_{v \in \mathcal{V}} (\bm{y}^{(\gamma,v)} - \hat{\bm{y}}^{(\gamma,v)})^2},\\
\text{MAPE} &=\frac{1}{|\mathcal{V}| \times |\mathcal{R}|}\sum_{\gamma \in \mathcal{R}}\sum_{v \in \mathcal{V}} \frac{|\bm{y}^{(\gamma,v)} - \hat{\bm{y}}^{(\gamma,v)}|}{\bm{y}^{(\gamma,v)}},
\end{align}where $\mathcal{R}$ is a test set, $\gamma$ is a test sample, and $\bm{y}^{(\gamma,v)}$ is the ground-truth of node $v$ in $\gamma$.

% We use the mean absolute error (MAE), the mean absolute percentage error (MAPE), and the root mean squared error (RMSE) to measure the performance of different models.
%%%%%%%%%%%%%%%%%%%% SUBSUBSECTION
\subsubsection{Baselines}
We compare our proposed STG-NRDE to the baseline models listed below, as well as the previous models introduced in the related work section —- in total, we use 27 baseline models:
\begin{enumerate}
    \item HA~\cite{hamilton2020time} uses the average value of the last 12 times slices to predict the next value.
    \item ARIMA is a statistical model of time series analysis.
    \item VAR~\cite{hamilton2020time} is a time series model that captures spatial correlations among all traffic series.
    \item TCN~\cite{BaiTCN2018} consists of a stack of causal convolutional layers with exponentially enlarged dilation factors.
    \item FC-LSTM~\cite{sutskever2014sequence} is LSTM with fully connected hidden unit.
    \item GRU-ED~\cite{cho2014grued} is a GRU-based baseline and uses the encoder-decoder framework for multi-step time series prediction.
    \item DSANet~\cite{Huang2019DSANet} is a correlated time series prediction model using CNN networks and a self-attention mechanism for spatial correlations.
    \item MSTGCN~\cite{guo2019astgcn} is another version of ASTGCN, which gets rid of the spatial-temporal attention.
    \item MTGNN~\cite{Wu2020MTGNN} is a multivariate time series forecasting model with graph neural networks, which utilizes a spatial-based GCN and gated dilated causal convolution.
    \item TrafficTransformer~\cite{cai2020TrafficTransformer} is an encoder-decoder architecture that utilizes Transformer to model temporal dependencies and GCN to model spatial dependencies.
    \item S$^2$TAT~\cite{wang2022s2tat} integrates the attention mechanism and graph convolution to capture non-local spatio-temporal features synchronously.
    \item STTN~\cite{xu2020sttn} is a spatio-temporal transformer networks to improve the long-term prediction of traffic flows.
    \item GSMDR~\cite{Liu2022GMSDR} is a novel RNN variant to make use of multiple historical hidden states and current input for capturing the long-range spatial temporal dependencies.
    \item DSTAGNN~\cite{Lan2022DSTAGNN} is a novel framework that leverages spatial-temporal aware distance derived from historical data rather than relying on predefined static adjacency matrix.
\end{enumerate}

\begin{table}[t]
    \centering
    \caption{Forecasting error on PeMSD3, PeMSD4, PeMSD7 and PeMSD8}
    \label{tab:main_exp}
    \setlength{\tabcolsep}{1pt}
    \small
    \begin{tabular}{c ccc ccc ccc ccc}
        \toprule
        \multirow{2}{*}{Model}  & \multicolumn{3}{c}{PeMSD3}    & \multicolumn{3}{c}{PeMSD4}      & \multicolumn{3}{c}{PeMSD7}      & \multicolumn{3}{c}{PeMSD8}\\\cmidrule(lr){2-4} \cmidrule(lr){5-7} \cmidrule(lr){8-10} \cmidrule(lr){11-13}
                                & MAE & RMSE & MAPE             & MAE & RMSE & MAPE               & MAE & RMSE & MAPE               & MAE & RMSE & MAPE \\ \midrule
        HA                      & 31.58 & 52.39 & 33.78\%       & 38.03 & 59.24 & 27.88\%         & 45.12 & 65.64 & 24.51\%         & 34.86 & 59.24 & 27.88\% \\       
        ARIMA                   & 35.41 & 47.59 & 33.78\%       & 33.73 & 48.80 & 24.18\%         & 38.17 & 59.27 & 19.46\%         & 31.09 & 44.32 & 22.73\% \\        
        VAR                     & 23.65 & 38.26 & 24.51\%       & 24.54 & 38.61 & 17.24\%         & 50.22 & 75.63 & 32.22\%         & 19.19 & 29.81 & 13.10\% \\                     
        FC-LSTM                 & 21.33 & 35.11 & 23.33\%       & 26.77 & 40.65 & 18.23\%         & 29.98 & 45.94 & 13.20\%         & 23.09 & 35.17 & 14.99\% \\ 
        TCN                     & 19.32 & 33.55 & 19.93\%       & 23.22 & 37.26 & 15.59\%         & 32.72 & 42.23 & 14.26\%         & 22.72 & 35.79 & 14.03\% \\  
        TCN(w/o causal)         & 18.87 & 32.24 & 18.63\%       & 22.81 & 36.87 & 14.31\%         & 30.53 & 41.02 & 13.88\%         & 21.42 & 34.03 & 13.09\% \\  
        GRU-ED                  & 19.12 & 32.85 & 19.31\%       & 23.68 & 39.27 & 16.44\%         & 27.66 & 43.49 & 12.20\%         & 22.00 & 36.22 & 13.33\% \\       
        DSANet                  & 21.29 & 34.55 & 23.21\%       & 22.79 & 35.77 & 16.03\%         & 31.36 & 49.11 & 14.43\%         & 17.14 & 26.96 & 11.32\% \\       
        STGCN                   & 17.55 & 30.42 & 17.34\%       & 21.16 & 34.89 & 13.83\%         & 25.33 & 39.34 & 11.21\%         & 17.50 & 27.09 & 11.29\% \\ 
        DCRNN                   & 17.99 & 30.31 & 18.34\%       & 21.22 & 33.44 & 14.17\%         & 25.22 & 38.61 & 11.82\%         & 16.82 & 26.36 & 10.92\% \\ 
        GraphWaveNet            & 19.12 & 32.77 & 18.89\%       & 24.89 & 39.66 & 17.29\%         & 26.39 & 41.50 & 11.97\%         & 18.28 & 30.05 & 12.15\% \\ 
        ASTGCN(r)               & 17.34 & 29.56 & 17.21\%       & 22.93 & 35.22 & 16.56\%         & 24.01 & 37.87 & 10.73\%         & 18.25 & 28.06 & 11.64\% \\ 
        MSTGCN                  & 19.54 & 31.93 & 23.86\%       & 23.96 & 37.21 & 14.33\%         & 29.00 & 43.73 & 14.30\%         & 19.00 & 29.15 & 12.38\% \\       
        STG2Seq                 & 19.03 & 29.83 & 21.55\%       & 25.20 & 38.48 & 18.77\%         & 32.77 & 47.16 & 20.16\%         & 20.17 & 30.71 & 17.32\% \\ 
        LSGCN                   & 17.94 & 29.85 & 16.98\%       & 21.53 & 33.86 & 13.18\%         & 27.31 & 41.46 & 11.98\%         & 17.73 & 26.76 & 11.20\% \\       
        STSGCN                  & 17.48 & 29.21 & 16.78\%       & 21.19 & 33.65 & 13.90\%         & 24.26 & 39.03 & 10.21\%         & 17.13 & 26.80 & 10.96\% \\    
        MTGNN             & 16.46 & 28.56 & 16.46\%       & 19.32 & 31.57 & 13.52\%         & 20.82 & 34.09 &  9.03\%         & 15.71 & 24.62 & 10.03\% \\
        AGCRN                   & 15.98 & 28.25 & 15.23\%       & 19.83 & 32.26 & 12.97\%         & 22.37 & 36.55 &  9.12\%         & 15.95 & 25.22 & 10.09\% \\
        STFGNN                  & 16.77 & 28.34 & 16.30\%       & 20.48 & 32.51 & 16.77\%         & 23.46 & 36.60 &  9.21\%         & 16.94 & 26.25 & 10.60\% \\        
        TrafficTransformer & 16.39 & 27.87 & 15.84\%       & 19.16 & 30.57 & 13.70\%         & 23.90 & 36.85 & 10.90\%          & 15.37 & 24.21 &  10.09\% \\
        STTN         & 17.51 & 29.55 & 17.36\%       & 19.48 & 31.91 & 13.63\%         & 21.34 & 34.59 & 9.93\%          & 15.48 & 24.97 & 10.34\% \\
        S$^{2}$TAT   & 16.02 & 27.21 & 15.75\%       & 19.19 & 31.05 & 12.77\%         & 22.42 & 35.81 & 9.75\%          & 15.73 & 25.12 & 10.06\% \\
        STGODE                  & 16.50 & 27.84 & 16.69\%       & 20.84 & 32.82 & 13.77\%         & 22.59 & 37.54 & 10.14\%         & 16.81 & 25.97 & 10.62\% \\
        Z-GCNETs                & 16.64 & 28.15 & 16.39\%       & 19.50 & 31.61 & 12.78\%         & 21.77 & 35.17 &  9.25\%         & 15.76 & 25.11 & 10.01\% \\
        DSTAGNN      & 15.57 & 27.21 & \textbf{14.68}\%       & 19.30 & 31.46 & 12.70\%         & 21.42 & 34.51 & 9.01\%          & 15.67 & 24.77 & 9.94\% \\
        GMSDR        & 15.84 & 27.05 & 15.27\%       & 20.49 & 32.13 & 14.15\%         & 22.27 & 34.94 & 9.86\%          & 16.36 & 25.58 & 10.28\% \\
        STG-NCDE                & 15.57 & 27.09 & 15.06\%       & 19.21 & 31.09 & 12.76\%         & 20.53 & 33.84 & 8.80\%          & 15.45 & 24.81 &  9.92\% \\
        \midrule 
        \textbf{STG-NRDE}       & \textbf{15.50} & 27.06 & 14.90\% & \textbf{19.13} & \textbf{30.94} & \textbf{12.68}\%   & \textbf{20.45}  & \textbf{33.73} & \textbf{8.65}\%  & \textbf{15.32} & \textbf{24.72} & \textbf{8.90}\% \\
        \textbf{Only Temporal}  & 20.44 & 33.22 & 20.10\%       & 26.17 & 40.61 & 18.09\%      & 28.62 & 44.26 & 12.56\%         & 21.28  & 33.02 & 13.23\%         \\
        \textbf{Only Spatial}   & 15.79 & \textbf{26.88} & 15.45\%       & 19.46 & 31.41 & 13.13\%      & 21.01 & 34.09 &  8.91\%         & 17.02 & 26.57 & 11.37\%  \\
        \bottomrule
    \end{tabular}
\end{table}

%%%%%%%%%%%%%%%%%%%% SUBSUBSECTION
\subsubsection{Hyperparameters}~\label{subsubsec:hyperparam}
For our method, we test with the following hyperparameter configurations: we train for 200 epochs using the Adam optimizer, with a batch size of 64 on all datasets. The two dimensionalities of $\dim(\bm{h}^{(v)})$ and $\dim(\bm{z}^{(v)})$ are \{32, 64, 128, 256\}, the node embedding size $C$ is from 1 to 10, and the number of $K$ in Equation~\ref{eq:fun_f} is in \{1, 2, 3\}. The sub-path length $P$ is in \{1,2,3\} and the depth $D$ is in \{1,2,3,4\}. The learning rate in all methods is in \{\num{5e-3}, \num{2e-3}, \num{1e-3}, \num{8e-4}, \num{5e-4}\} and the weight decay coefficient is in \{\num{5e-3}, \num{1e-3}, \num{2e-3}\}. With the validation dataset, an early-stop approach with a patience of 15 iterations is applied. If the accuracy of baselines is unknown for a dataset, we run them through a hyperparameter search procedure based on their recommended settings (See Appendix~\ref{app:detail} for detail). We use their officially reported accuracy if it is known. For reproducibility, we introduce the best hyperparameter configurations for our model in each dataset as follows:
\begin{enumerate}
    \item In PeMSD3, we set the number of $K$ to 1 and the node embedding size $C$ to 2. The depth $D$ is 3 and the sub-path length $P$ is 2. The dimensionality of hidden vector is 64. The learning rate was set to \num{1e-3} and the weight decay coefficient was \num{1e-3}. 
    \item In PeMSD4, we set the number of $K$ to 2 and the node embedding size $C$ to 8. The depth $D$ is 2 and the sub-path length $P$ is 2. The dimensionality of hidden vector is 64. The learning rate was set to \num{1e-3} and the weight decay coefficient was \num{1e-3}.
    \item In PeMSD7, we set the number of $K$ to 2 and the node embedding size $C$ to 10. The depth $D$ is 2 and the sub-path length $P$ is 2. The dimensionality of hidden vector is 64. The learning rate was set to \num{1e-3} and the weight decay coefficient was \num{8e-4}.
    \item In PeMSD8, we set the number of $K$ to 1 and the node embedding size $C$ to 2. The depth $D$ is 2 and the sub-path length $P$ is 2. The dimensionality of hidden vector is 32. The learning rate was set to \num{8e-4} and the weight decay coefficient was \num{2e-3}.
    \item In PeMSD7(M), we set the number of $K$ to 1 and the node embedding size $C$ to 10. The depth $D$ is 3 and the sub-path length $P$ is 2. The dimensionality of hidden vector is 64. The learning rate was set to \num{1e-3} and the weight decay coefficient was \num{1e-3}.
    \item In PeMSD7(L), we set the number of $K$ to 2 and the node embedding size $C$ to 10. The depth $D$ is 4 and the sub-path length $P$ is 2. The dimensionality of hidden vector is 32. The learning rate was set to \num{1e-3} and the weight decay coefficient was \num{1e-3}.
\end{enumerate}

\begin{table}[t]
    \centering
    \caption{Forecasting error on PeMSD7(M) and PeMSD7(L)}
    \label{tab:main_exp_2}
    \setlength{\tabcolsep}{2pt}
    \small
    \begin{tabular}{ccc cccc  ccc}
        \toprule
        \multirow{2}{*}{Model} & \multicolumn{3}{c}{PeMSD7(M)}  & \multicolumn{3}{c}{PeMSD7(L)} \\\cmidrule(lr){2-4} \cmidrule(lr){5-7}
                                & MAE  & RMSE  & MAPE    & MAE  & RMSE  & MAPE    \\ \midrule
        HA                      & 4.59 &  8.63 & 14.35\% & 4.84 &  9.03 & 14.90\% \\       
        ARIMA                   & 7.27 & 13.20 & 15.38\% & 7.51 & 12.39 & 15.83\% \\        
        VAR                     & 4.25 &  7.61 & 10.28\% & 4.45 &  8.09 & 11.62\% \\                     
        FC-LSTM                 & 4.16 &  7.51 & 10.10\% & 4.66 &  8.20 & 11.69\% \\               
        TCN                     & 4.36 &  7.20 &  9.71\% & 4.05 &  7.29 & 10.43\% \\               
        TCN(w/o causal)         & 4.43 &  7.53 &  9.44\% & 4.58 &  7.77 & 11.53\% \\               
        GRU-ED                  & 4.78 &  9.05 & 12.66\% & 3.98 &  7.71 & 10.22\% \\       
        DSANet                  & 3.52 &  6.98 &  8.78\% & 3.66 &  7.20 &  9.02\% \\       
        STGCN                   & 3.86 &  6.79 & 10.06\% & 3.89 &  6.83 & 10.09\% \\       
        DCRNN                   & 3.83 &  7.18 &  9.81\% & 4.33 &  8.33 & 11.41\% \\       
        GraphWaveNet            & 3.19 &  6.24 &  8.02\% & 3.75 &  7.09 &  9.41\% \\       
        ASTGCN(r)               & 3.14 &  6.18 &  8.12\% & 3.51 &  6.81 &  9.24\% \\       
        MSTGCN                  & 3.54 &  6.14 &  9.00\% & 3.58 &  6.43 &  9.01\% \\       
        STG2Seq                 & 3.48 &  6.51 &  8.95\% & 3.78 &  7.12 &  9.50\% \\       
        LSGCN                   & 3.05 &  5.98 &  7.62\% & 3.49 &  6.55 &  8.77\% \\       
        STSGCN                  & 3.01 &  5.93 &  7.55\% & 3.61 &  6.88 &  9.13\% \\        
        MTGNN        & 2.96 &  5.80 &  7.36\% & 3.26 &  6.07 &  7.82\% \\
        AGCRN                   & 2.79 &  5.54 &  7.02\% & 2.99 &  5.92 &  7.59\% \\
        STFGNN                  & 2.90 &  5.79 &  7.23\% & 2.99 &  5.91 &  7.69\% \\        
        TrafficTransformer
                                & 3.45 &  6.50 &  8.38\% & 3.79 &  6.70 &  9.26\% \\
        STTN         & 3.09 &  6.11 &  7.80\% & 3.50 &  6.77 &  8.89\% \\
        S$^{2}$TAT   & 2.72 &  5.46 &  6.81\% & 2.88 &  5.89 &  7.35\% \\
        STGODE                  & 2.97 &  5.66 &  7.36\% & 3.22 &  5.98 &  7.94\% \\
        Z-GCNETs                & 2.75 &  5.62 &  6.89\% & 2.91 &  5.83 &  7.33\% \\
        DSTAGNN      & 2.95 &  5.86 &  7.15\% & 3.05 &  5.98 &  7.70\% \\
        GMSDR        & 2.86 &  5.69 &  7.01\% & 3.10 &  5.96 &  7.68\% \\
        STG-NCDE                & 2.68 &  5.39 &  6.76\% & 2.87 &  5.76 &  7.31\% \\
        \midrule
        \textbf{STG-NRDE}       & \textbf{2.66} & \textbf{5.31}  & \textbf{6.68}\% & \textbf{2.85} &  \textbf{5.76} & \textbf{7.14}\%\\
        \textbf{Only Temporal}  & 3.29 &  6.63 & 8.26\% & 3.48 &  6.99 &  8.72\% \\
        \textbf{Only Spatial}   & 2.67 &  5.37 & 6.73\% & 2.90 &  5.77 &  7.40\% \\
        \bottomrule 
    \end{tabular}
\end{table}
%%%%%%%%%%%%%%%%%%%% SUBSECTION %%%%%%%%%%%%%%%%%%%%
\subsection{Experimental Results}
Tables~\ref{tab:main_exp} and~\ref{tab:main_exp_2} present the detailed prediction performance. Overall, as shown in Table~\ref{tab:average}, our proposed method, STG-NRDE, clearly marks the best average accuracy. We list the average MAE/RMSE/MAPE from the 6 datasets for each notable model. We also show the relative accuracy in comparison to our method within the parentheses. STGCN, for example, has an MAE that is 17.7\% percent worse than our method. All existing methods have higher errors across all metrics than our method by large margins for many baselines.

We now describe experimental results in each dataset. STG-NRDE shows the best accuracy in all cases, followed by STG-NCDE, Z-GCNETs, AGCRN, STGODE, and so on. There are no existing methods that are as stable as STG-NRDE. For instance, STG-NCDE shows reasonably low errors in many cases, e.g., an MAPE of 15.06\% in PeMSD3 by STG-NCDE, which is the second-best result vs. 14.90\% by STG-NRDE. For PeMSD3, STGODE outperforms AGCRN and Z-GCNETS, but not for PeMSD7. Only our method, STG-NRDE, consistently produces accurate forecasts.

We also visualize the ground-truth and the predicted curves by our method, STG-NCDE, and Z-GCNETs in Fig.~\ref{fig:pred_vis}. Node 12 in PeMSD3, Node 111 in PeMSD4, Node 13 in PeMSD7, and Node 150 in PeMSD8 are several of the highest traffic areas for each dataset. Since STG-NCDE and Z-GCNETs show reasonable performance, their predicted curve is similar to that of our method in many time-points. As highlighted with the bounding boxes, however, our method shows much more accurate predictions for challenging cases. In particular, our method significantly outperforms Z-GCNETs for the highlighted time-points for Node 111 in PeMSD4 and Node 13 in PeMSD7, for which Z-GCNETs shows nonsensical predictions, i.e., the prediction curves are straight. Our method also has the strength to predict the traffic flow at the peak points such as Node 70 in PeMSD3 and Node 92 in PeMSD8.

\begin{figure*}[t]
    \centering
    \subfigure[Node 12 in PeMSD3]{\includegraphics[width=0.24\textwidth]{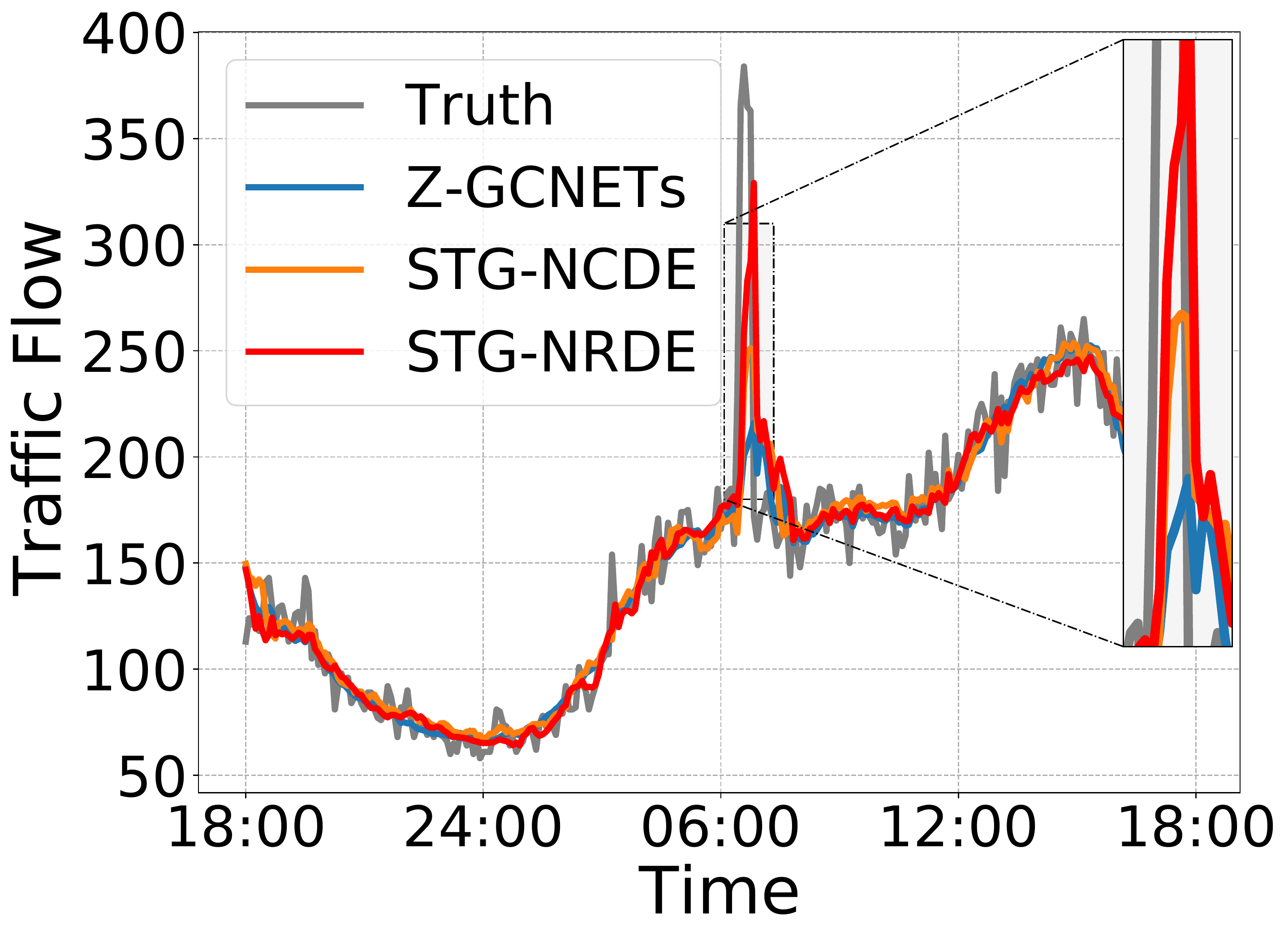}}
    \subfigure[Node 70 in PeMSD3]{\includegraphics[width=0.24\textwidth]{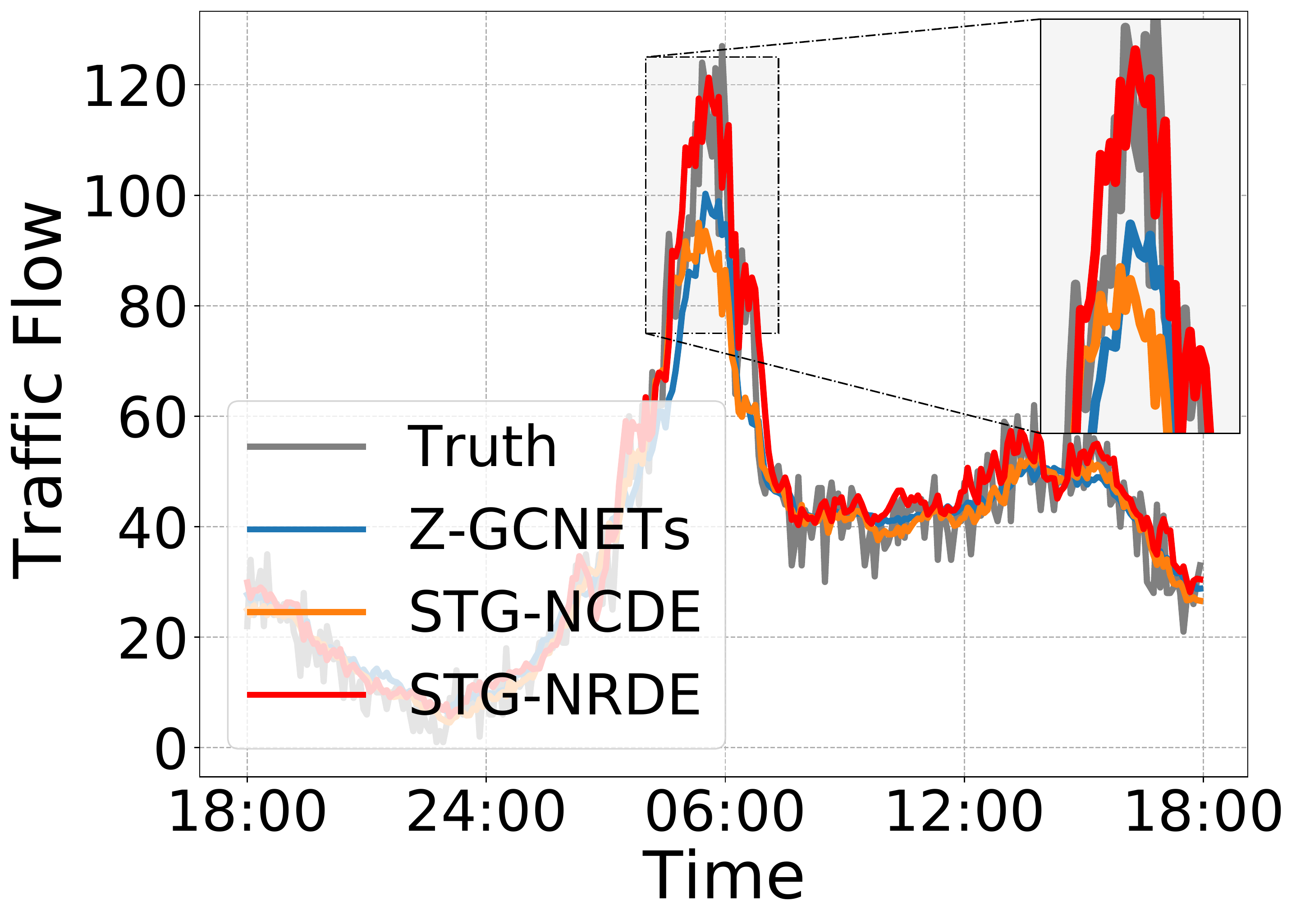}}
    \subfigure[Node 111 in PeMSD4]{\includegraphics[width=0.24\textwidth]{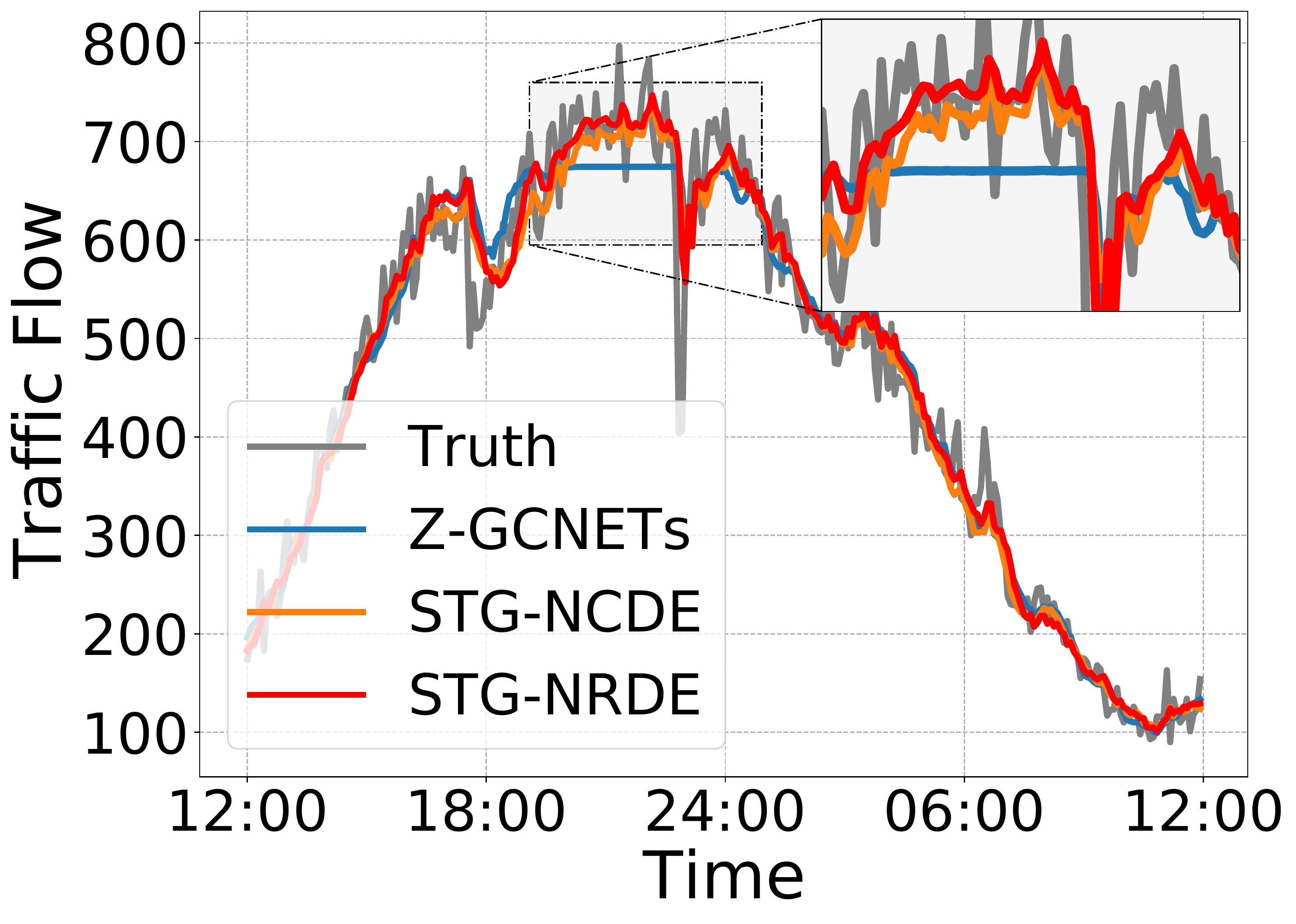}}
    \subfigure[Node 149 in PeMSD4]{\includegraphics[width=0.24\textwidth]{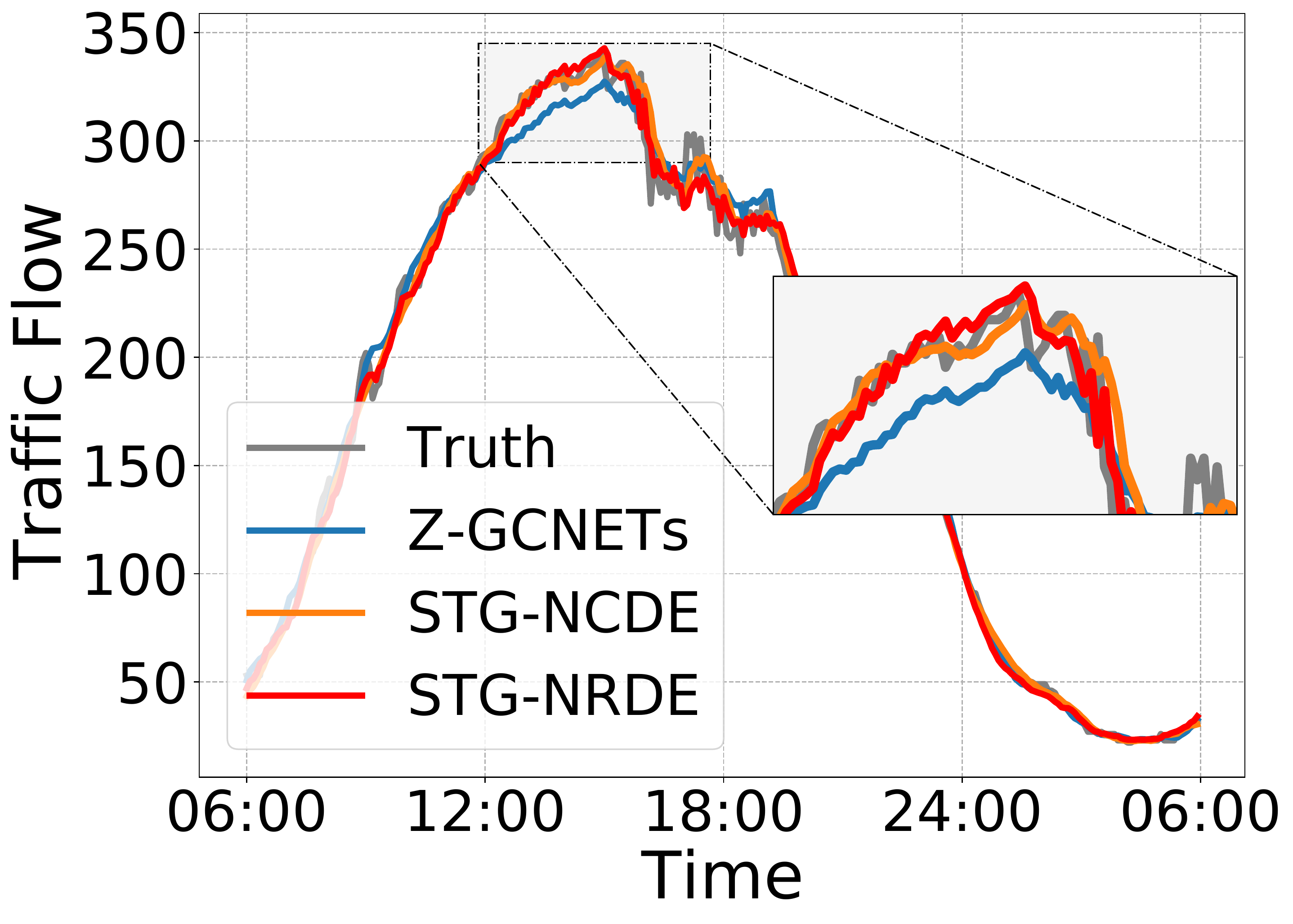}}
    \subfigure[Node 4 in PeMSD7]{\includegraphics[width=0.24\textwidth]{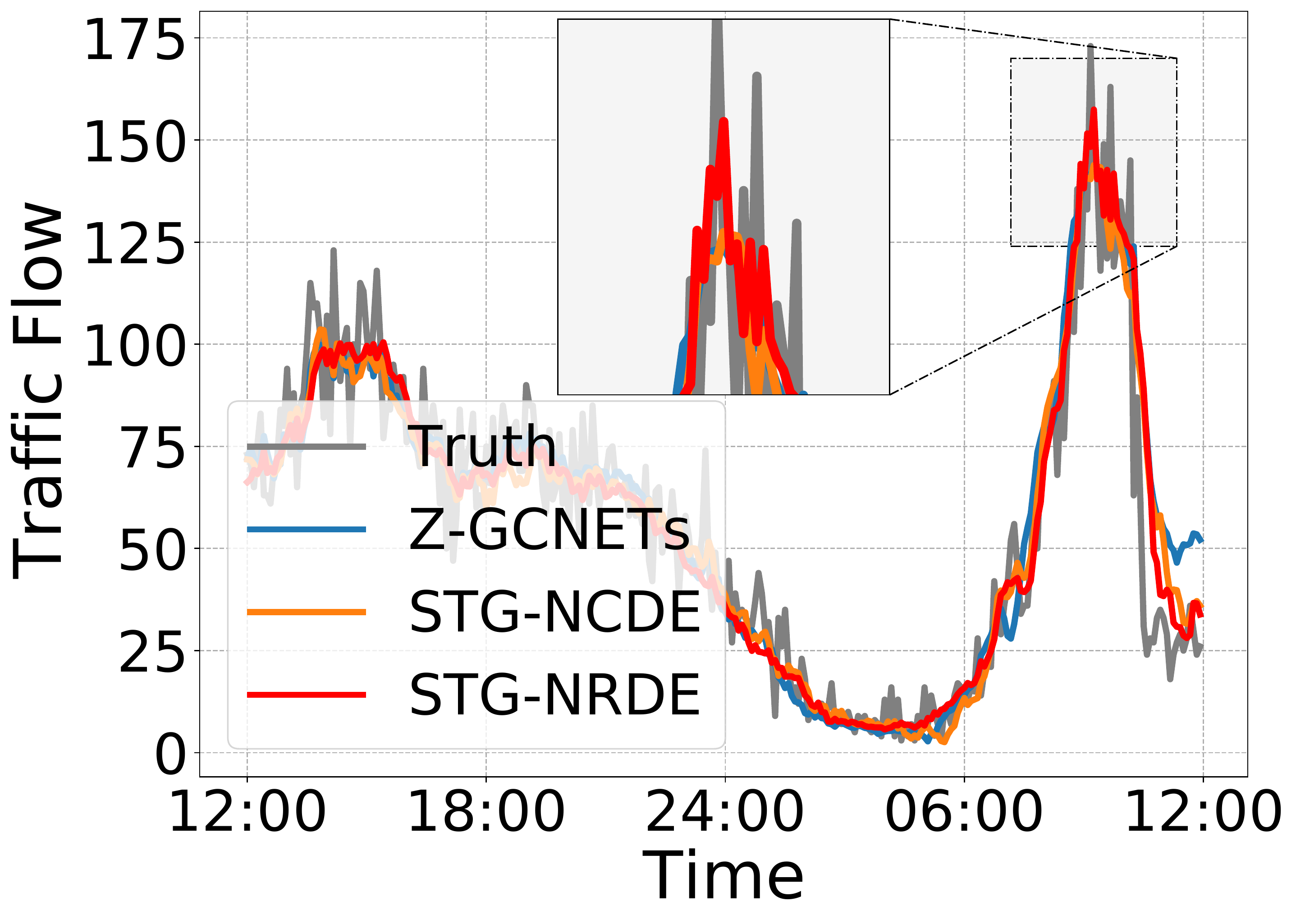}}
    \subfigure[Node 13 in PeMSD7]{\includegraphics[width=0.24\textwidth]{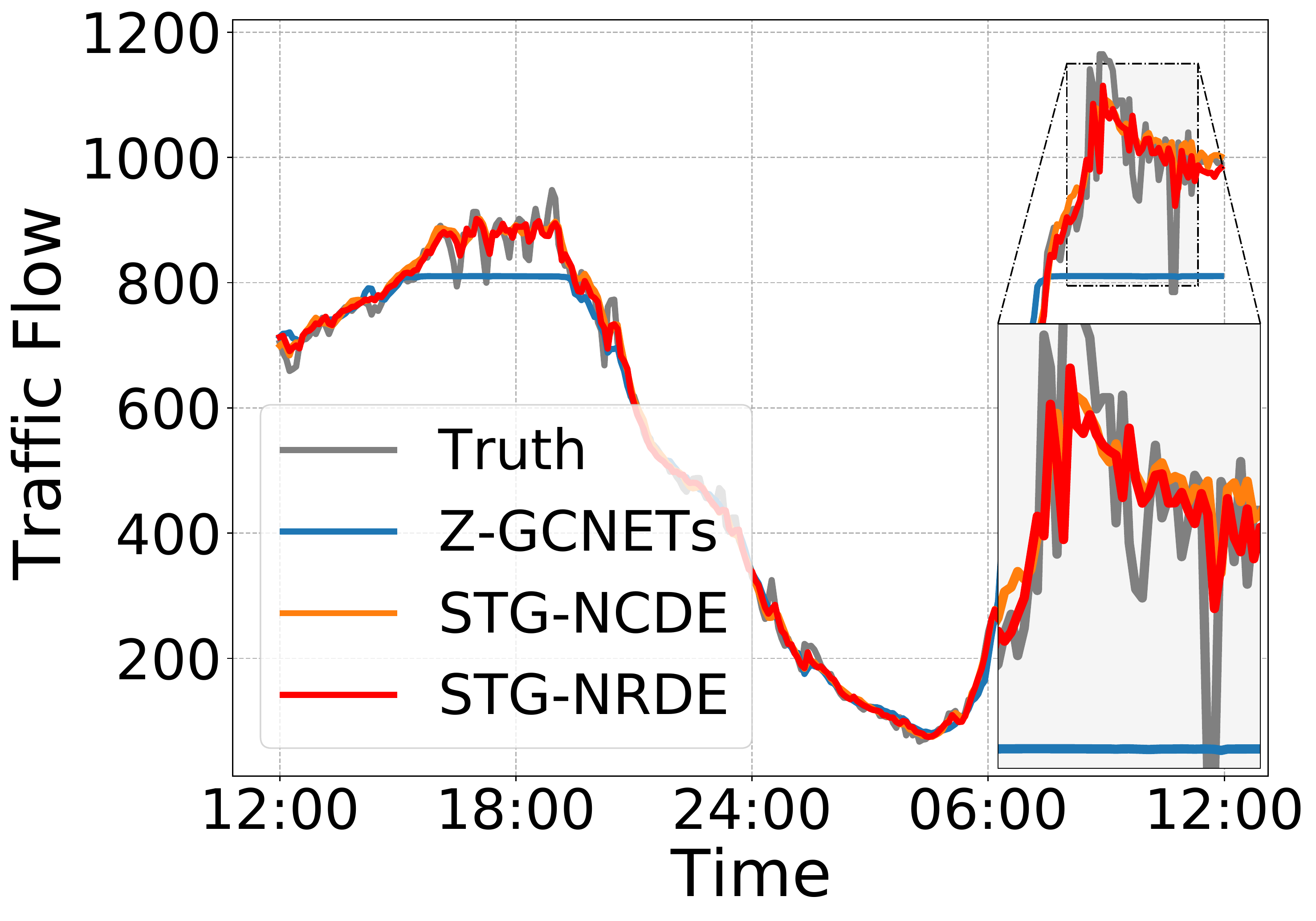}}
    \subfigure[Node 92 in PeMSD8]{\includegraphics[width=0.24\textwidth]{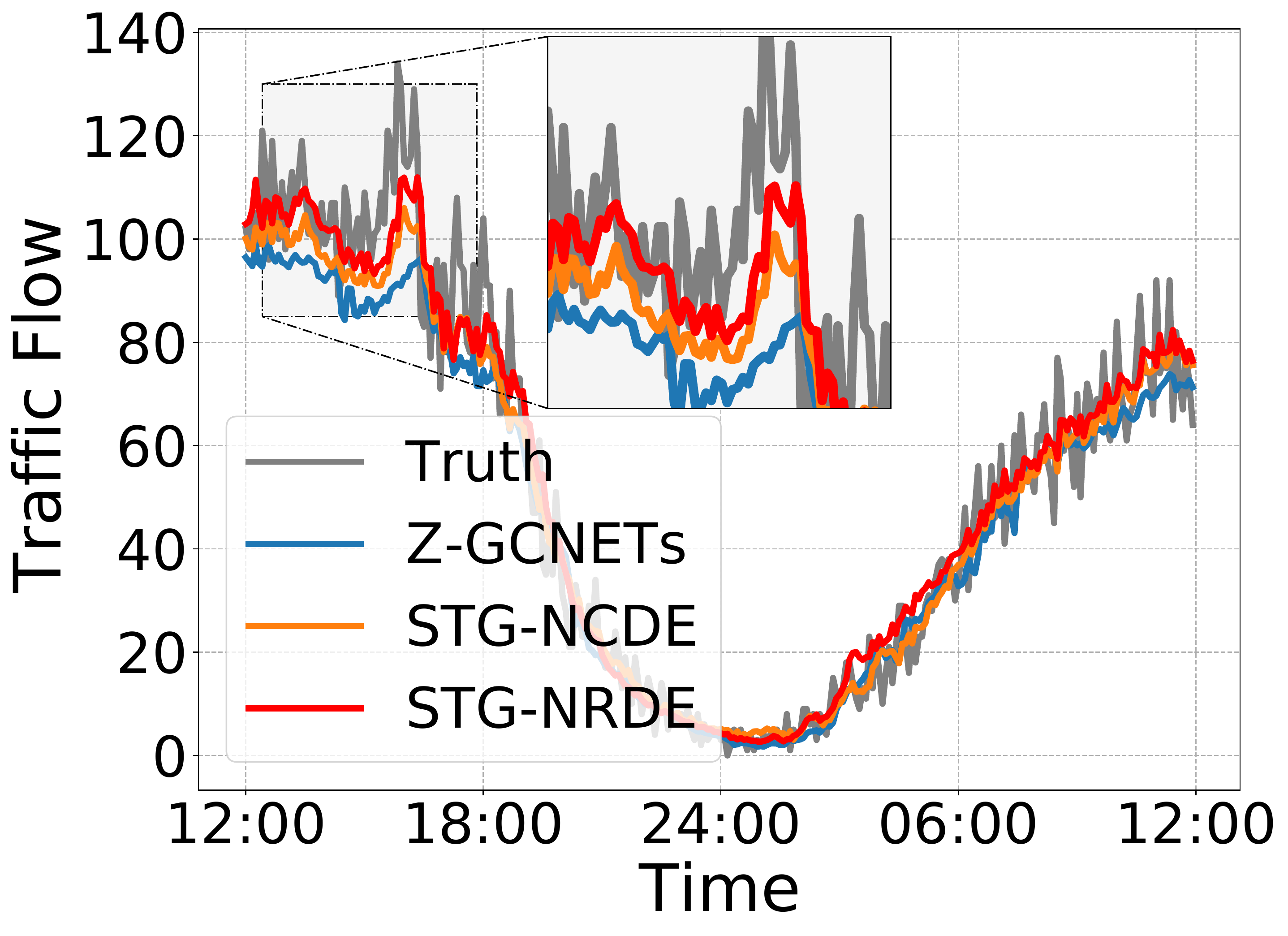}}
    \subfigure[Node 150 in PeMSD8]{\includegraphics[width=0.24\textwidth]{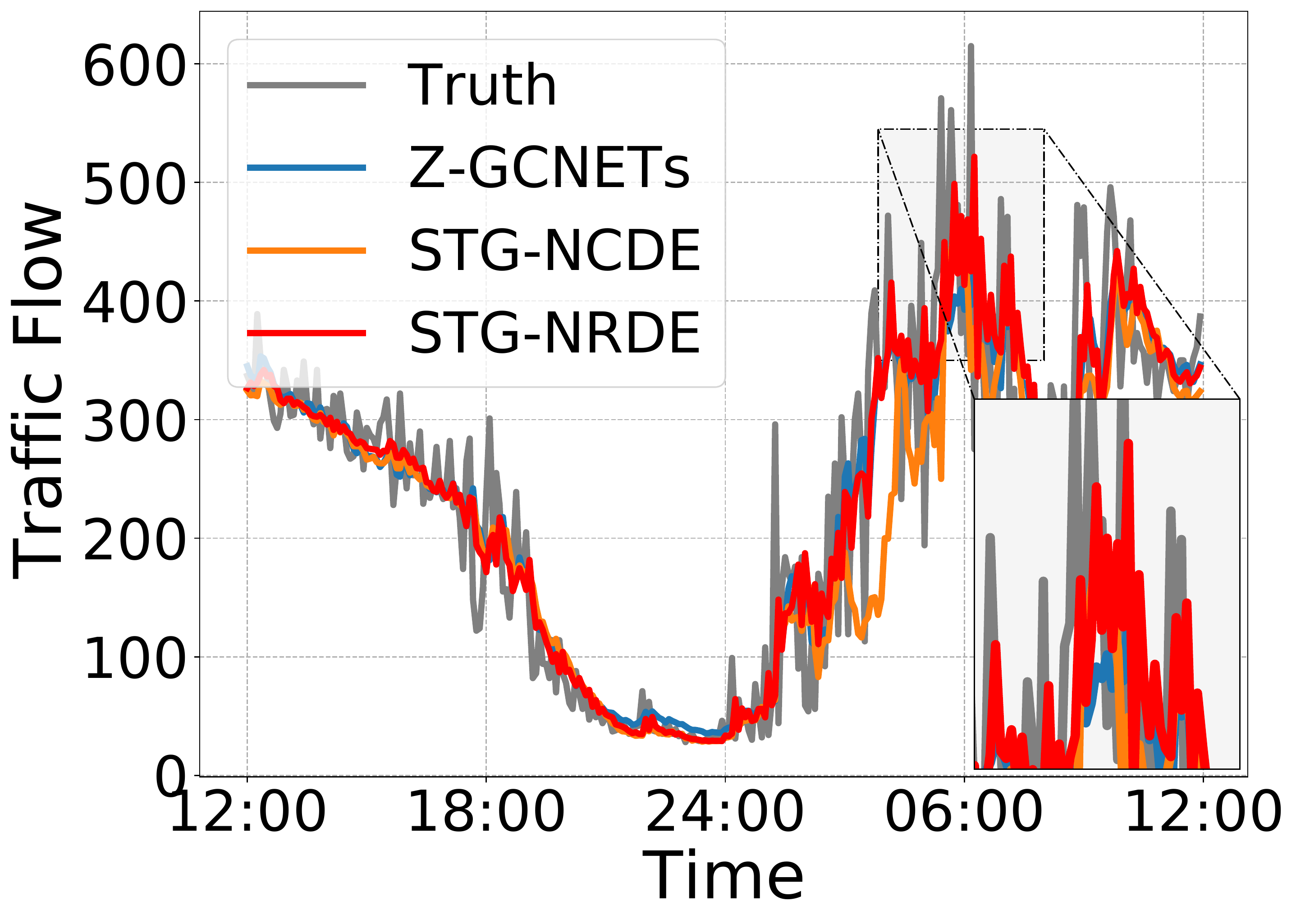}}
    \caption{Traffic forecasting visualization for PeMSD3, PeMSD4, PeMSD7 and PeMSD8 on a randomly selected day}
    \label{fig:pred_vis}
\end{figure*}

\begin{figure}[t]
    \centering
    \subfigure[Training curve in PeMSD3]{\includegraphics[width=0.33\textwidth]{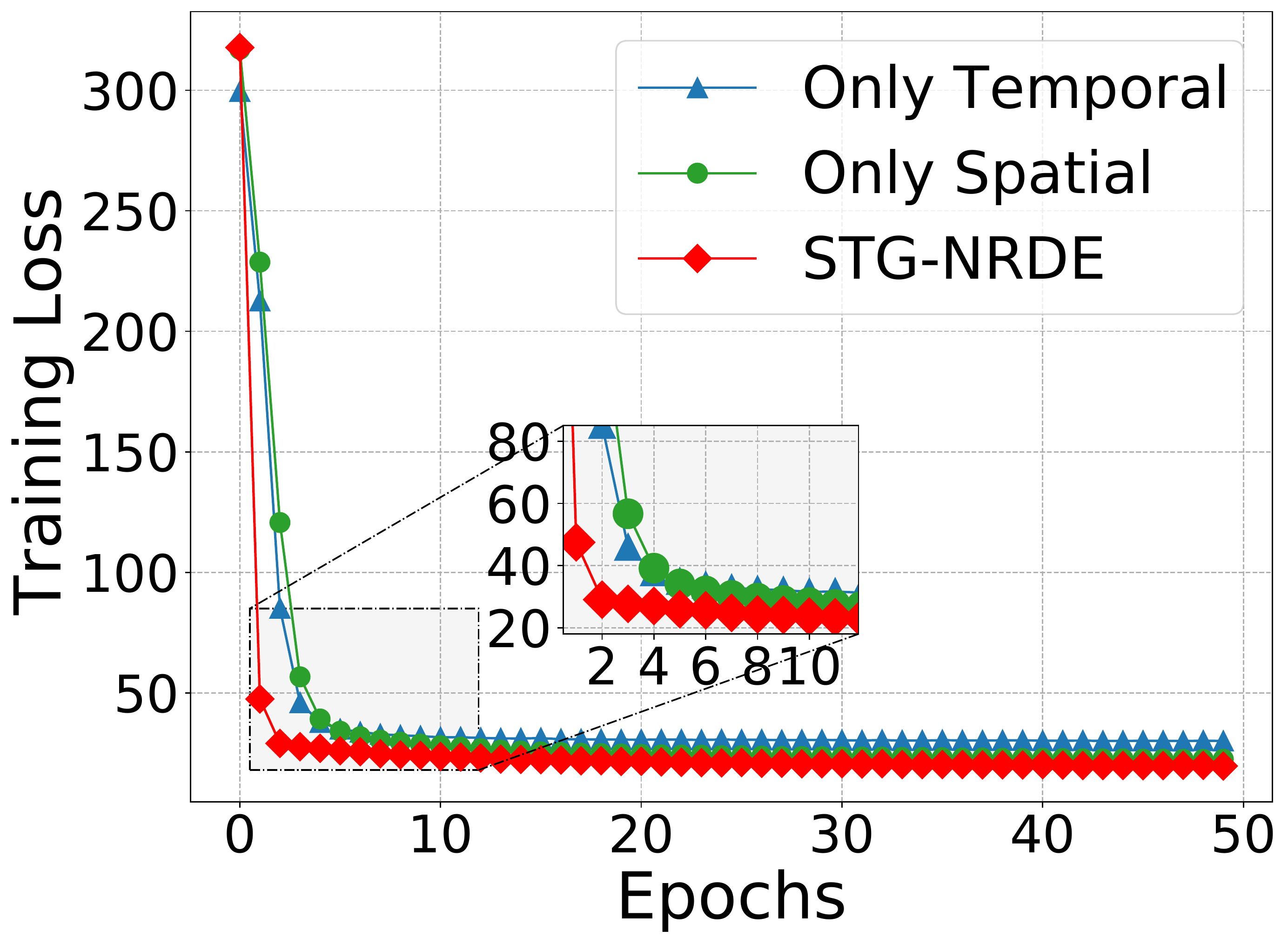}}
    \subfigure[Training curve in PeMSD7]{\includegraphics[width=0.33\textwidth]{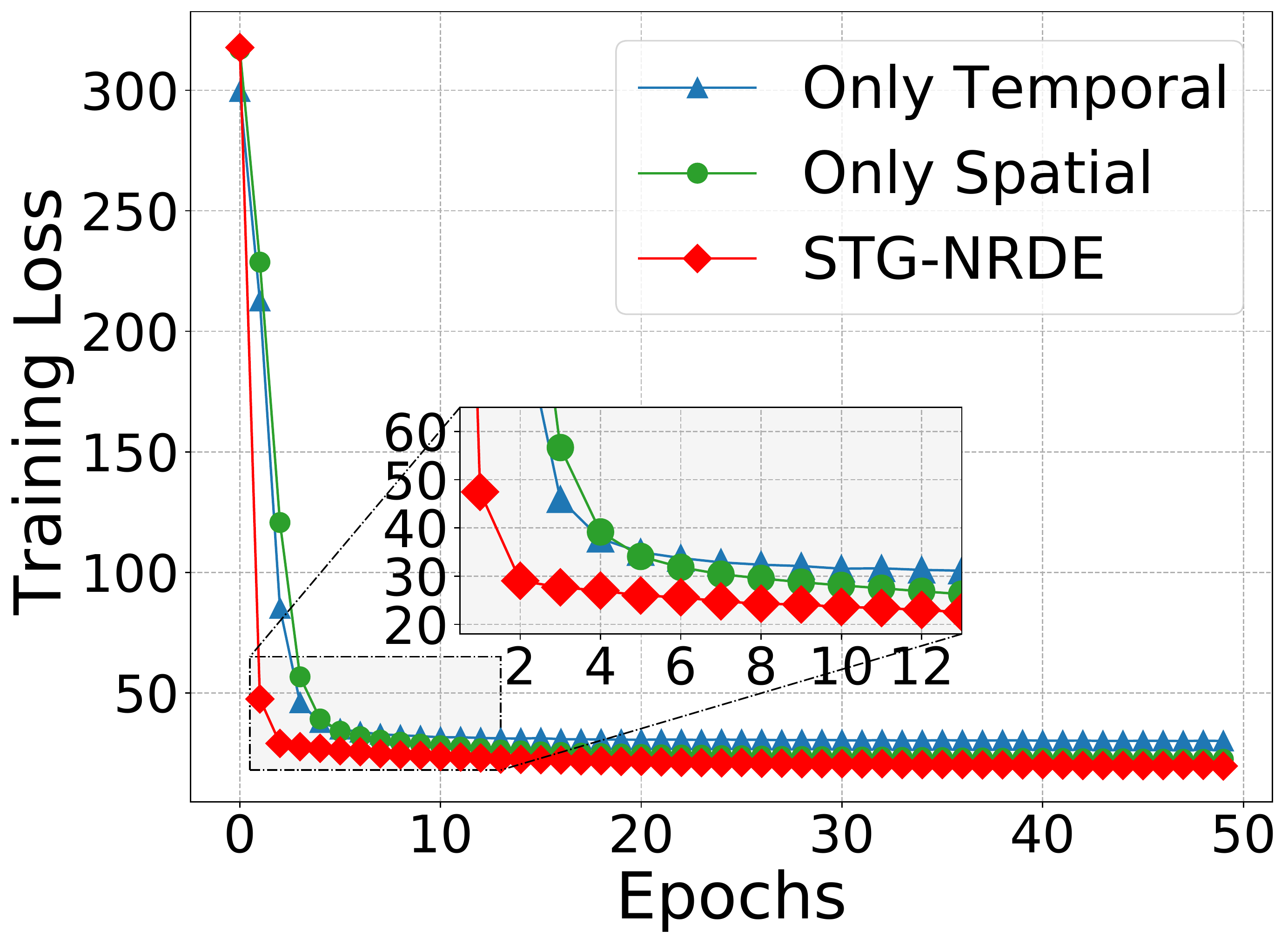}}
    \caption{Training curve}
    \label{fig:loss_pemsd7}
\end{figure}

%%%%%%%%%%%%%%%%%%%% SUBSECTION %%%%%%%%%%%%%%%%%%%%
% \subsection{Ablation, Sensitivity, and Additional Studies}\label{sec:abl}
\subsection{Ablation Study}\label{sec:abl}

% \subsubsection{Ablation Study} 
We define the following two models as ablation study models: i) The first ablation model contains only the temporal processing component, i.e., Equation~\ref{eq:type1-2}, and ii) the second ablation model contains only the spatial processing component, which can be written as follows:
\begin{align}\begin{split}
\bm{Z}(T) = \bm{Z}(0) + \int_{0}^{T} g(\bm{Z}(t);\bm{\theta}_g) \frac{\text{LogSig}^D_{r_i,r_{i+1}}(\bm{X})}{r_{i+1} - r_i} dt,\label{eq:abl}
\end{split}\end{align}where the trajectory $\bm{Z}(t)$ over time is controlled by the sequence of the time-derivative of the log-signature. As a result, for this ablation study model, we change the model architecture. The first (resp. second) model is denoted as ``Only temporal'' (resp. ``Only spatial'') in the tables.

In every case, the ablation study model using only spatial processing outperforms the model using only temporal processing, for example, an RMSE of 26.88 in PeMSD3 using only spatial processing vs. 33.22 using only temporal processing. On the other hand, STG-NRDE uses both temporal and spatial processing, and the effect on spatial processing is more significant. This demonstrates that we require both of them in order to achieve the highest level of model accuracy.

In PeMSD3 and PeMSD7, we compare their training curves in Fig.~\ref{fig:loss_pemsd7}. The loss curve of the STG-NRDE is stabilized after the second epoch, whereas the loss curves of the other two ablation models take longer to be stabilized.
    
We also show the results of using and not using log-signature for the ``Only spatial'' model in the Table~\ref{tbl:only_spatial}. A model that does not use log-signature can be called a ``Only spatial'' model of STG-NCDE. It can be seen that performance improves when log-signature is used in all traffic flow benchmark datasets except MAPE for PeMSD3 and PeMSD8.

\begin{table}[t]
    \centering
    \caption{Ablation study of log-signature on ``Only spatial'' model. LS denotes the log-signature.}
    \label{tbl:only_spatial}
    \setlength{\tabcolsep}{1pt}
    \resizebox{\textwidth}{!}{
    \begin{tabular}{l ccc ccc ccc ccc ccc ccc}
        \toprule
        \multirow{2}{*}{Model}  & \multicolumn{3}{c}{PeMSD3} & \multicolumn{3}{c}{PeMSD4} & \multicolumn{3}{c}{PeMSD7} & \multicolumn{3}{c}{PeMSD8} & \multicolumn{3}{c}{PeMSD7(M)} & \multicolumn{3}{c}{PeMSD7(L)}\\\cmidrule(lr){2-4} \cmidrule(lr){5-7} \cmidrule(lr){8-10} \cmidrule(lr){11-13} \cmidrule(lr){14-16} \cmidrule(lr){17-19}
                                & MAE & RMSE  & MAPE & MAE & RMSE & MAPE & MAE & RMSE & MAPE & MAE & RMSE & MAPE & MAE & RMSE & MAPE & MAE & RMSE & MAPE\\ \midrule
        w/o LS
                                & 15.92 & 27.17 & \textbf{15.14}\%
                                & 19.86 & 31.92 & 13.35\%
                                & 21.72 & 34.73 &  9.24\%
                                & 17.58 & 27.76 & \textbf{11.27}\%
                                &  2.77 & 5.40 & 7.00\%
                                &  2.99 & 5.85 & 7.60\%\\
        w/ LS
                                & \textbf{15.79} & \textbf{26.88} & 15.45\% 
                                & \textbf{19.46} & \textbf{31.41} & \textbf{13.13}\% 
                                & \textbf{21.01} & \textbf{34.09} &  \textbf{8.91}\%         
                                & \textbf{17.02} & \textbf{26.57} & 11.37\%
                                & \textbf{2.67}  & \textbf{5.37}  & \textbf{6.73}\%
                                & \textbf{2.90}  & \textbf{5.77}  & \textbf{7.40}\% \\
        \bottomrule
    \end{tabular}
    }
\end{table}

\begin{table}[t]
    \small
    \centering
    \caption{Forecasting error on irregular PeMSD3, PeMSD4, PeMSD7 and PeMSD8}
    \label{tbl:irregular}
    \setlength{\tabcolsep}{1pt}
    \resizebox{\textwidth}{!}{
    \begin{tabular}{cc ccc ccc ccc ccc}
        \toprule
        \multirow{2}{*}{Model}  & \multirow{2}{*}{\shortstack{Missing\\rate}} & \multicolumn{3}{c}{PeMSD3} & \multicolumn{3}{c}{PeMSD4} & \multicolumn{3}{c}{PeMSD7} & \multicolumn{3}{c}{PeMSD8}\\\cmidrule(lr){3-5} \cmidrule(lr){6-8} \cmidrule(lr){9-11} \cmidrule(lr){12-14}
                                && MAE & RMSE  & MAPE & MAE    & RMSE  & MAPE   & MAE    & RMSE  & MAPE & MAE    & RMSE  & MAPE\\ \midrule
        STG-NCDE                & \multirow{4}{*}{10\%} & 15.89 & 27.61 & 15.83\% & 19.36 & 31.28 & 12.79\% & 20.65 & 33.95 & 8.86\%  & 15.68  & 24.96 & 10.05\% \\
        \textbf{STG-NRDE}       &                       & \textbf{15.62} & \textbf{27.15} & \textbf{15.08}\% & \textbf{19.10} & \textbf{30.93} & 1\textbf{2.61}\% & \textbf{20.61} & \textbf{33.84} &  \textbf{8.75}\% & \textbf{15.57}  & \textbf{24.83} & \textbf{10.04}\% \\
        \textbf{Only Temporal}  &                       & 58.41 & 86.08 & 56.24\% & 26.19 & 40.62 & 18.17\% & 29.10 & 44.75 & 12.58\% & 21.09  & 32.81 & 13.19\% \\
        \textbf{Only Spatial }  &                       & 16.62 & 28.04 & 16.15\% & 19.62 & 31.64 & 13.13\% & 21.32 & 34.40 &  9.08\% & 16.07  & 25.38 & 10.57\% \\\midrule
        STG-NCDE                & \multirow{4}{*}{30\%} & 16.08 & 27.78 & 16.05\% & 19.40 & 31.30 & 13.04\% & 20.76 & 34.20 &  8.91\% & 16.21  & 25.64 & 10.43\% \\
        \textbf{STG-NRDE}       &                       & \textbf{15.64} & \textbf{27.30} & \textbf{15.38}\% & \textbf{19.39} & \textbf{31.38} & \textbf{12.69}\% & \textbf{20.74} & \textbf{34.06} & \textbf{8.83}\% & \textbf{15.47}  & \textbf{24.69} & \textbf{10.02}\% \\
        \textbf{Only Temporal}  &                       & 58.31 & 86.56 & 60.29\% & 26.66 & 41.13 & 18.31\% & 29.14 & 44.75 & 13.09\% & 21.68  & 33.55 & 13.87\% \\
        \textbf{Only Spatial }  &                       & 16.72 & 27.80 & 16.56\% & 19.54 & 31.55 & 13.06\% & 22.88 & 36.10 &  9.86\% & 16.78  & 26.26 & 10.83 \% \\\midrule
        STG-NCDE                & \multirow{4}{*}{50\%} & 16.50 & 28.52 & 16.03\% & 19.98 & 32.09 & 13.84\% & 21.51 & 34.91 &  9.25\% & 16.68  & 26.17 & 10.67\% \\
        \textbf{STG-NRDE}       &                       & \textbf{16.14} & \textbf{27.91} & \textbf{15.85}\% & \textbf{19.54} & \textbf{31.64} & \textbf{12.96}\% & \textbf{21.25} & \textbf{34.71} &  \textbf{9.04}\% & \textbf{15.95}  & \textbf{25.35} & \textbf{10.49}\% \\
        \textbf{Only Temporal}  &                       & 61.63 & 89.36 & 62.41\% & 27.75 & 42.78 & 19.10\% & 30.60 & 46.84 & 13.32\% & 22.65  & 35.05 & 14.03\% \\
        \textbf{Only Spatial }  &                       & 17.53 & 29.15 & 16.70\% & 19.98 & 32.19 & 13.77\% & 21.95 & 35.17 &  9.40\% & 16.99  & 26.71 & 10.94\% \\
        \bottomrule
    \end{tabular}
    }
\end{table}

\begin{table}[t]
    \small
    \centering
    \caption{Forecasting error on irregular PeMSD7(M) and PeMSD7(L)}
    \label{tbl:irregular_02}
    \setlength{\tabcolsep}{3pt}
    \begin{tabular}{cc ccc ccc}
        \toprule
        \multirow{2}{*}{Model}  & \multirow{2}{*}{\shortstack{Missing\\rate}} & \multicolumn{3}{c}{PeMSD7(M)} & \multicolumn{3}{c}{PeMSD7(L)}\\\cmidrule(lr){3-5} \cmidrule(lr){6-8}
                                &&                          MAE & RMSE & MAPE           & MAE & RMSE & MAPE   \\ \hline
        STG-NCDE                & \multirow{4}{*}{10\%} & 2.67 & 5.38 & 6.78\%                              & 2.90  & 5.78  & 7.34\% \\
        \textbf{STG-NRDE}       &                       & 2.66 & \textbf{5.31} &\textbf{ 6.65}\% & \textbf{2.86} & \textbf{5.77} & \textbf{7.16}\% \\
        \textbf{Only Temporal}  &                       & 3.30 & 6.64 & 8.23\% & 3.49 & 6.99 &  8.69\% \\
        \textbf{Only Spatial }  &                       & \textbf{2.65} & 5.34 & 6.70\% & 2.90 & 5.78 & 7.30\% \\\midrule
        STG-NCDE                & \multirow{4}{*}{30\%} & 2.72 & 5.45 & 6.81\% & 2.89 & 5.80 & 7.25\% \\
        \textbf{STG-NRDE}       &                       & \textbf{2.69} & \textbf{5.34} & \textbf{6.73}\%   & \textbf{2.87} & \textbf{5.80} & \textbf{7.21}\% \\
        \textbf{Only Temporal}  &                       & 3.33 & 6.67 & 8.31\% & 3.53 & 7.05 & 8.86\% \\
        \textbf{Only Spatial }  &                       & 2.69 & 5.41 & 6.82\% & 2.97 & 5.86 & 7.49\% \\\midrule
        STG-NCDE                & \multirow{4}{*}{50\%} & 2.75 & 5.53 & \textbf{6.79}\% & 3.03  & 5.97  & 7.63\% \\
        \textbf{STG-NRDE}       &                       & \textbf{2.74} & \textbf{5.48} & 6.90 \% & \textbf{2.93} & \textbf{5.85} & \textbf{7.32}\% \\
        \textbf{Only Temporal}  &                       & 3.45 & 6.89 & 8.66\% & 3.65 & 7.26 & 9.11\% \\
        \textbf{Only Spatial }  &                       & 2.83 & 5.55 & 7.32\% & 3.02 & 5.99 & 7.67\% \\
        \bottomrule
    \end{tabular}
    % }
\end{table}
\subsection{Irregular Traffic Forecasting} In the real-world, traffic sensors can be damaged, and we may be unable to collect data in some areas for a period of time. To reflect this situation, we randomly drop 10\% to 50\% of sensing values for each node independently. NRDEs are able to process irregular time-series since their cubic spline interpolation method is able to produce continuous paths from it. Therefore, STG-NRDE is able to do so without changing its model designs, which is one of its most distinguishing characteristics when compared to most of the existing baselines. Tables~\ref{tbl:irregular} and~\ref{tbl:irregular_02} summarize the results. In comparison with the results in Table~\ref{tab:main_exp}, our model's performance is not significantly degraded. We note that the other baselines listed in Table~\ref{tab:main_exp}, with the exception of STG-NCDE, cannot perform irregular forecasting, so we compare STG-NRDE with its ablation models and STG-NCDE in Tables~\ref{tbl:irregular} and~\ref{tbl:irregular_02}.

\subsection{Sensitivity Study}

\begin{figure}[t]
    \centering
    \subfigure[PeMSD3]{\includegraphics[width=0.32\textwidth]{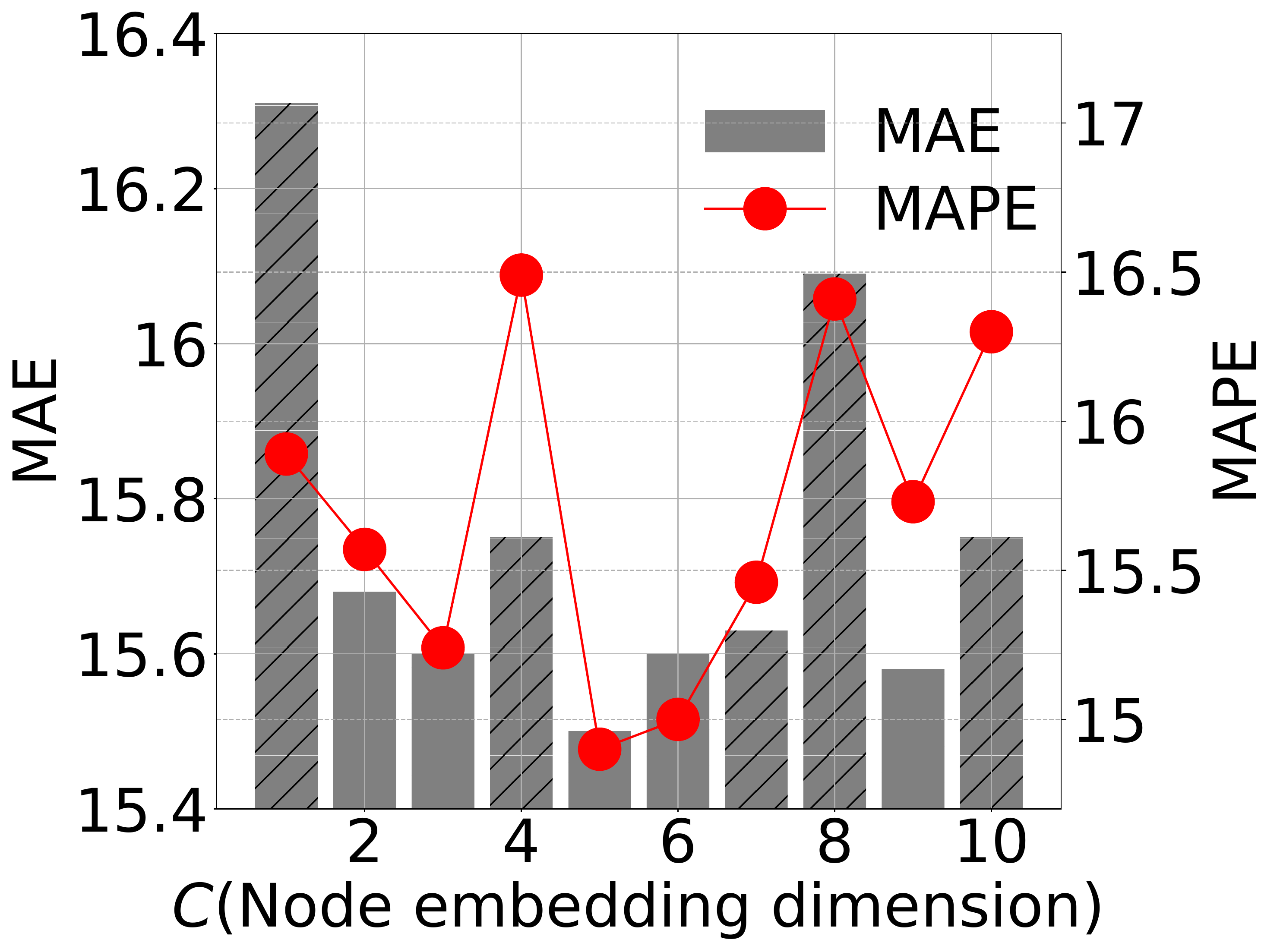}}
    \subfigure[PeMSD4]{\includegraphics[width=0.32\textwidth]{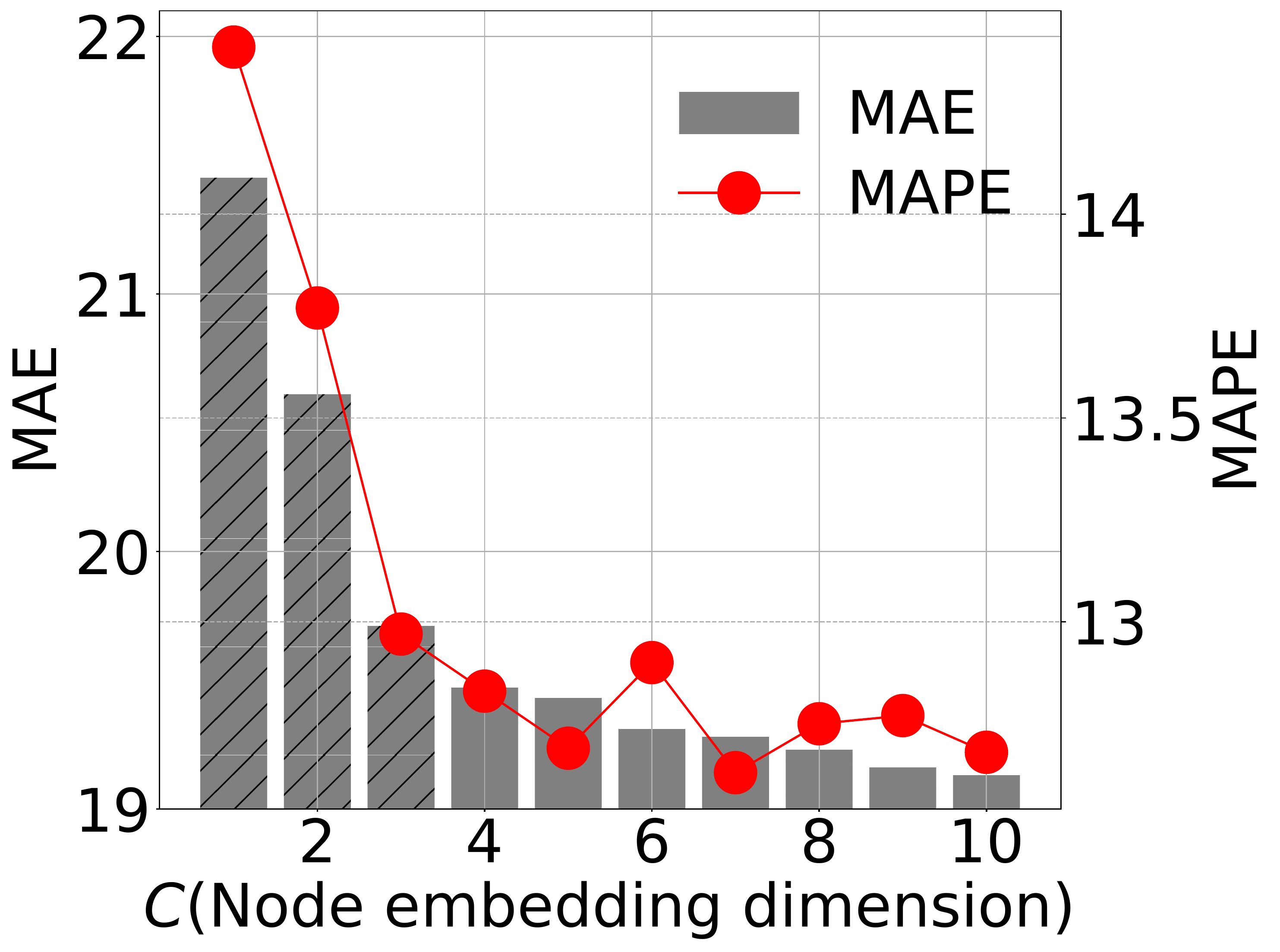}}
    \subfigure[PeMSD7]{\includegraphics[width=0.32\textwidth]{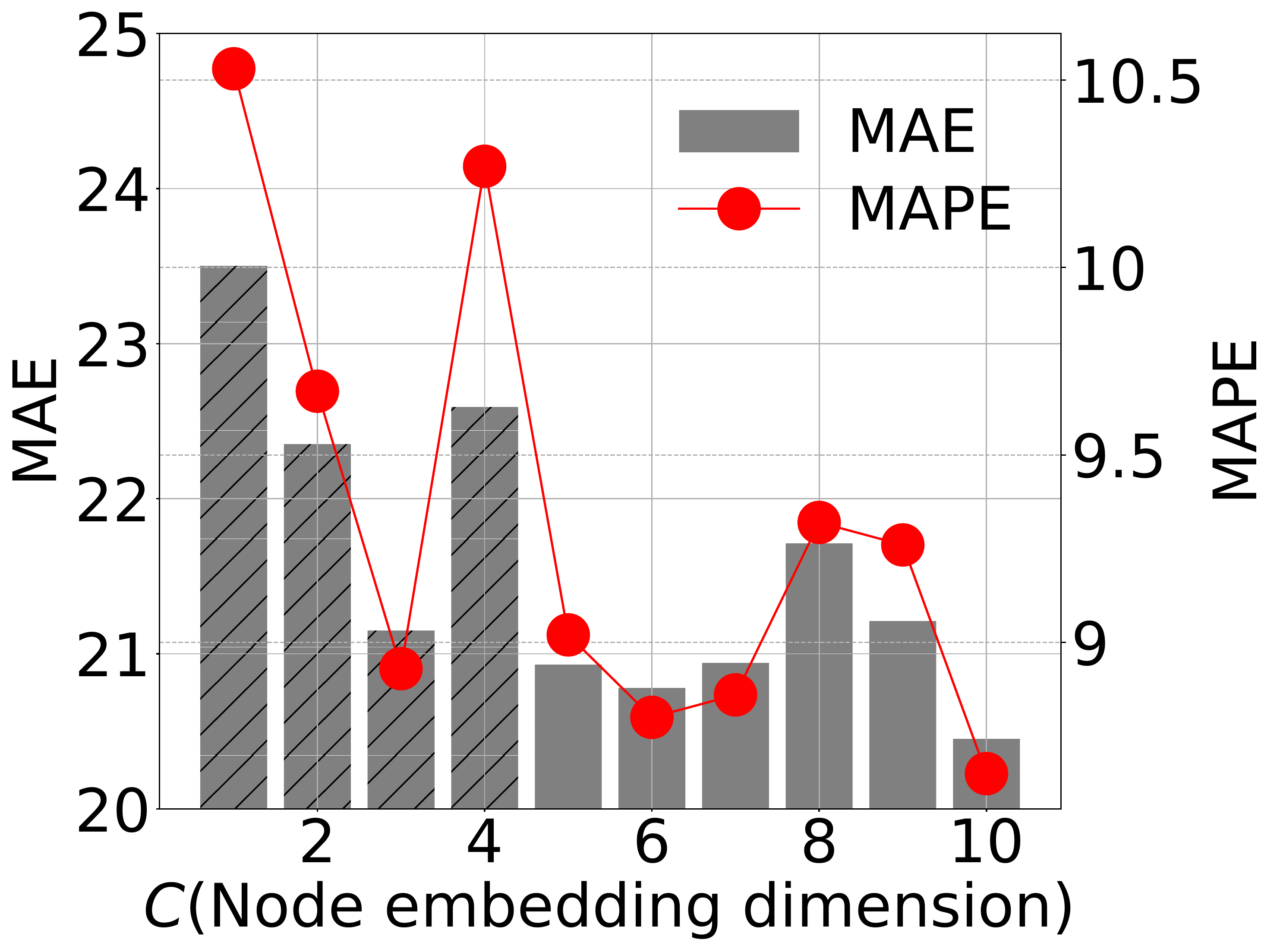}}
    \subfigure[PeMSD8]{\includegraphics[width=0.32\textwidth]{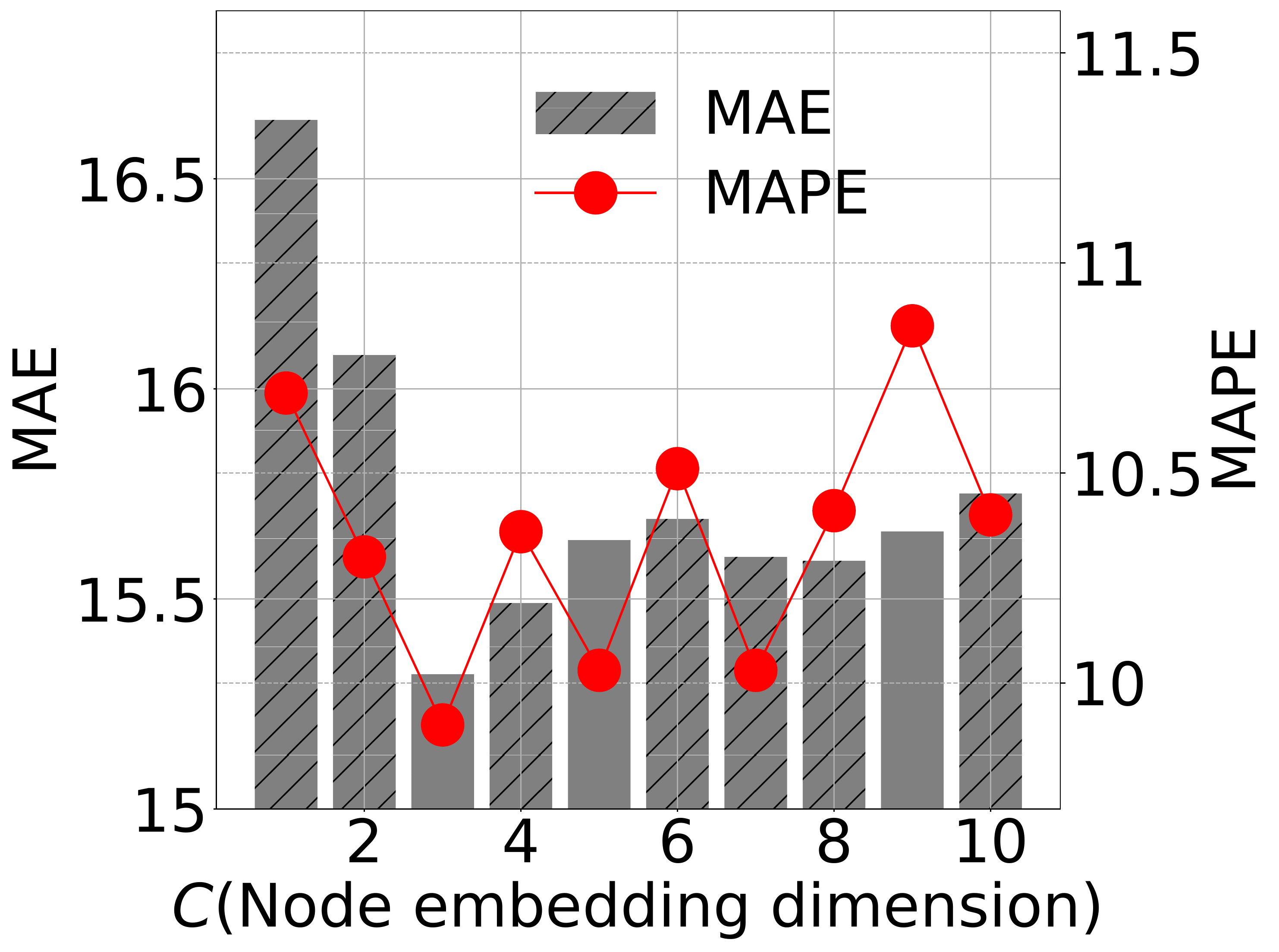}}
    \subfigure[PeMSD7(M)]{\includegraphics[width=0.32\textwidth]{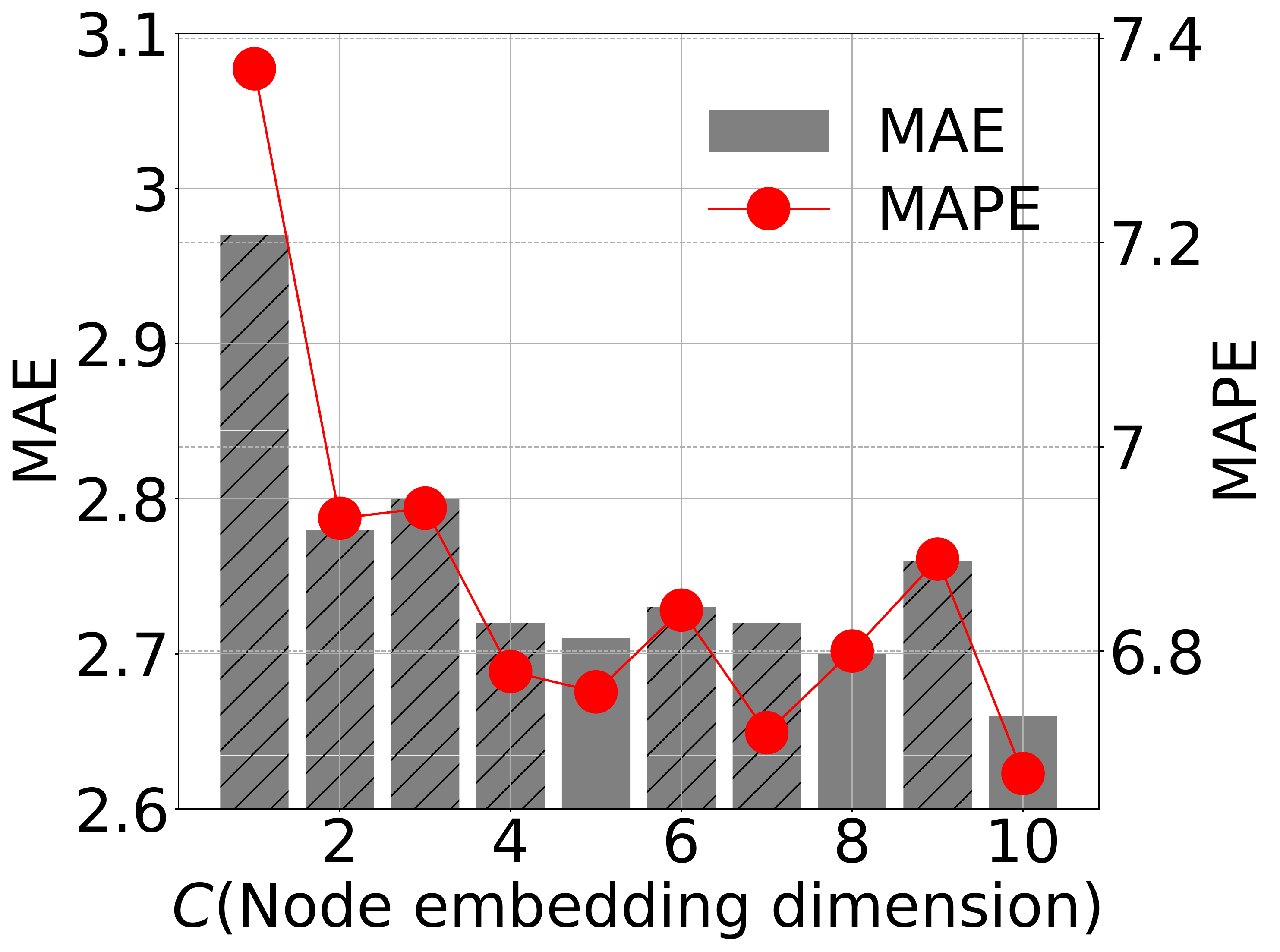}}
    \subfigure[PeMSD7(L)]{\includegraphics[width=0.32\textwidth]{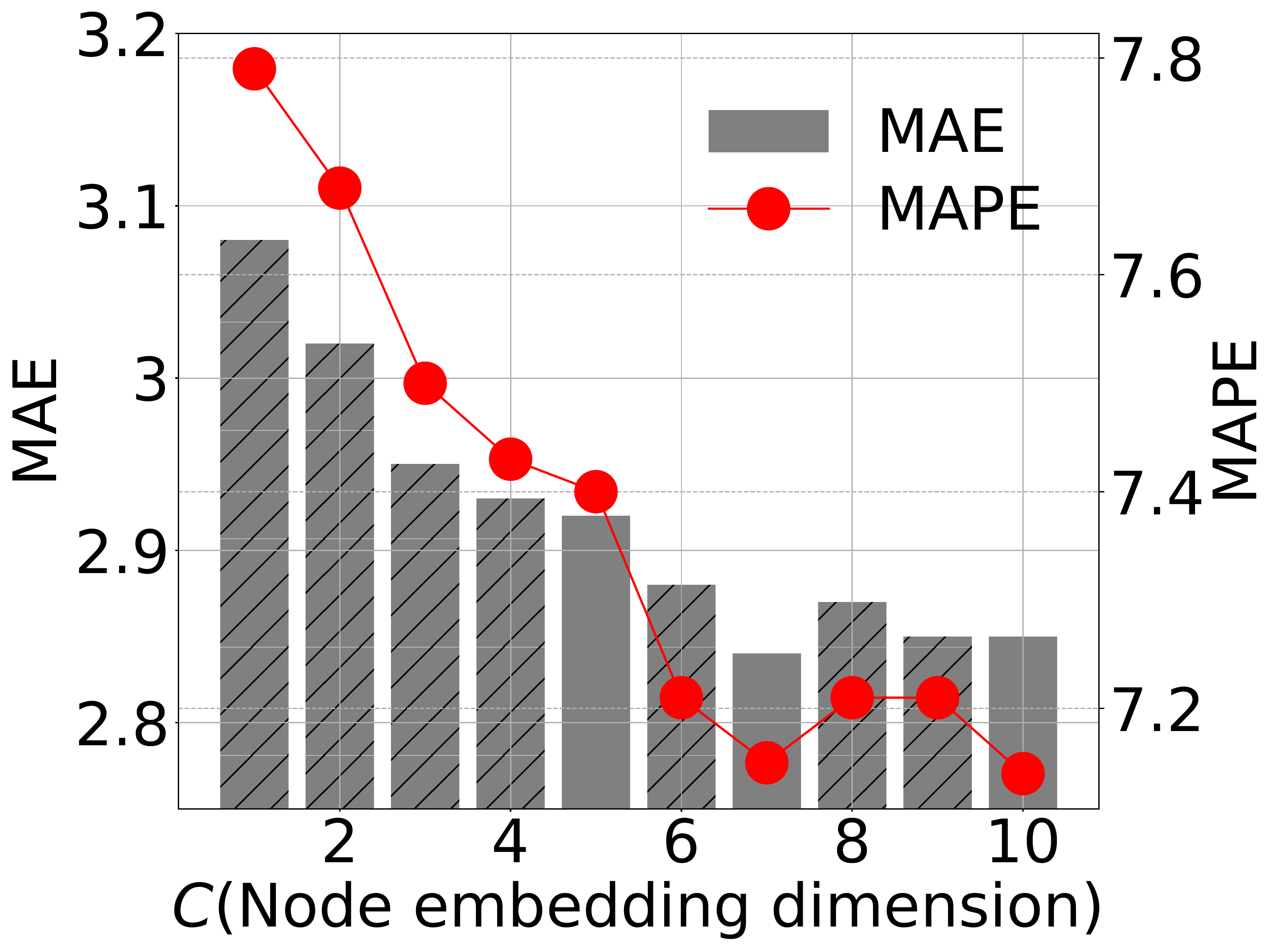}}
    \caption{Sensitivity to $C$}
    \label{fig:sensitivity_appendix}
\end{figure}
\subsubsection{Sensitivity to $C$} Fig.~\ref{fig:sensitivity_appendix} shows the MAE and MAPE by varying the node embedding size $C$. For PeMSD4, PeMSD7(M), abd PeMSD7(L), the two error metrics are stabilized after $C=6$, and then we can achieve the best accuracy with $C=10$. In the other datasets, we can find the best accuracy at a specific size with a different pattern. For instance, the error metrics are lowest when $C=3$ for PeMSD8.

\subsubsection{Sensitivity to $D$} Table~\ref{tbl:abl_depth} shows the forecasting error by varying the log-signature depth $D$. Our method performs best when the depth $D=2$, with the exception of the PeMSD3, PeMSD7(M), and PeMSD7(L) datasets. For PeMSD7(L), the depth $D=4$ shows the best result because the higher depth represents each sub-path accurately. This sensitivity study indicates that there can be performance differences based on the log-signature depth.

\begin{table}[t]
    \centering
    % \small
    \setlength{\tabcolsep}{1pt}
    \caption{Sensitivity analysis for $D$ on PeMSD3, PeMSD4, PeMSD7, PeMSD8, PeMSD7(M) and PeMSD7(L)}
    \label{tbl:abl_depth}
    \resizebox{\textwidth}{!}{
    \begin{tabular}{c ccc ccc ccc ccc ccc ccc}
        \toprule
        \multirow{2}{*}{$D$} & \multicolumn{3}{c}{PeMSD3} & \multicolumn{3}{c}{PeMSD4} & \multicolumn{3}{c}{PeMSD7} & \multicolumn{3}{c}{PeMSD8} & \multicolumn{3}{c}{PeMSD7(M)} & \multicolumn{3}{c}{PeMSD7(L)}\\\cmidrule(lr){2-4} \cmidrule(lr){5-7} \cmidrule(lr){8-10} \cmidrule(lr){11-13}  \cmidrule(lr){14-16}  \cmidrule(lr){17-19}
          & MAE & RMSE & MAPE & MAE & RMSE & MAPE & MAE & RMSE & MAPE & MAE & RMSE & MAPE & MAE & RMSE & MAPE & MAE & RMSE & MAPE \\ \midrule
        1 & 15.73 & 26.99 & 15.35\% & 19.39 & 31.26 & 12.96\% & 21.51 & 34.43 & 9.18\% & 15.70 & 24.89 & 10.40\% & 2.68 & 5.39 & 6.83\%    & 2.87 & 5.80 & 7.22\%  \\
        2 & 15.50 & 27.06 & \textbf{14.90}\% & \textbf{19.13} & \textbf{30.94} & \textbf{12.68}\% & \textbf{20.45} & \textbf{33.73} & \textbf{8.65}\% & \textbf{15.32} & 24.72 &  \textbf{8.90}\% & 2.67 & 5.35 & \textbf{6.65}\% & 2.85 & 5.81 & 7.21\%  \\
        3 & \textbf{15.43} & \textbf{27.06} & 15.33\% & 19.34 & 31.17 & 12.78\% & 21.36 & 34.25 & 9.22\% & 15.58 & \textbf{24.68} & 10.08\% &\textbf{2.66} & \textbf{5.31} & 6.68\% & 2.86 & 5.80 & 7.27\%  \\
        4 & 16.29 & 27.94 & 15.88\% & 19.19 & 31.00 & 12.72\% & 21.10 & 34.20 & 9.03\% & 15.63 & 24.78 & 10.44\% & 2.69 & 5.32 & 6.64\%    &  \textbf{2.85} &\textbf{ 5.76} & \textbf{7.14}\%  \\
        \bottomrule
    \end{tabular}
    }
\end{table}

\subsubsection{Sensitivity to the number of GNN layers} Table~\ref{tab:layer} shows the MAE by varying the number of layer of GNN. Model performance tends to decrease as the number of message passing layers increases. Both the baselines and STG-NRDE we compared are the best when the number of layers is two.

\begin{table}[t]
    \small
    \centering
    \caption{MAE with respect to the number of GNN layers on PeMSD7}
    \begin{tabular}{c cccc}\toprule
        \# of layer & STG-NRDE & STG-NCDE & AGCRN & ST-GCN\\ \midrule
        1     & 20.91    & 21.07    & 23.01 & 25.45\\
        2     & \textbf{20.45}    & \textbf{20.53}    & \textbf{22.37} & \textbf{25.33}\\
        3     & 20.86    & 21.05    & 22.94 & 25.35\\
        4     & 20.85    & 21.05    & 22.98 & 25.51\\
        \bottomrule
    \end{tabular}
    \label{tab:layer}
\end{table}

Table~\ref{tab:layer} shows that the propagation range of GNN is more effective in using a regional range than a wide range of neighbor information. As the number of layers of GNN increases, the performance decreases, indicating that traffic forecasting is also a well-known oversmoothing problem in the GNN community.

\subsection{Additional Studies}
\subsubsection{Various GNN functions} STG-NRDE is a flexible framework wherein any standard message-passing layer (such as GAT~\cite{velickovic2018GAT}, GCN~\cite{kipf2017GCN}, or ChebNet~\cite{Defferrard2016}) can be used as the graph function. In Eq.~\eqref{eq:fun_g2}, The adaptive graph convolution (AGC) is used to configure the message passing layer. We can rewrite Eq.~\eqref{eq:fun_g2} as follows:
\begin{align}
\bm{B}_1 &=  m(\bm{B}_0),\label{eq:fun_g2-a}
\end{align} where $m$ is any GNN (message-passing) function.

\begin{table}[t]
    \small
    \centering
    \caption{Various message passing (GNN) functions on PeMSD7}
    \begin{tabular}{c cccc}\toprule
                 & AGC & ChebNet & GAT   & GCN\\ \midrule
        MAE  & \textbf{20.45}  & 21.04   & 23.09 & 23.15\\
        RMSE & \textbf{33.73}  & 34.10   & 34.89 & 35.11\\
        MAPE & \textbf{8.65}\% &  9.01\% & 10.01\% & 10.25\%\\
        \bottomrule
    \end{tabular}
    \label{tab:graph}
\end{table}

Table~\ref{tab:graph} shows the experimental results for the PeMSD7 dataset for not only AGC but also ChebNet, GAT, and GCN for function $m$. The error of AGC was the lowest at 20.45, and GAT, which learns weights for edges, had the highest error, so it was ineffective. And since GCN and ChebNet showed worse performance than AGC, it is effective to improve the spatial dependency using AGC of our model.    

\begin{figure}[t]
    \centering
    \subfigure[The original adjacency matrix]{\includegraphics[width=0.3\textwidth]{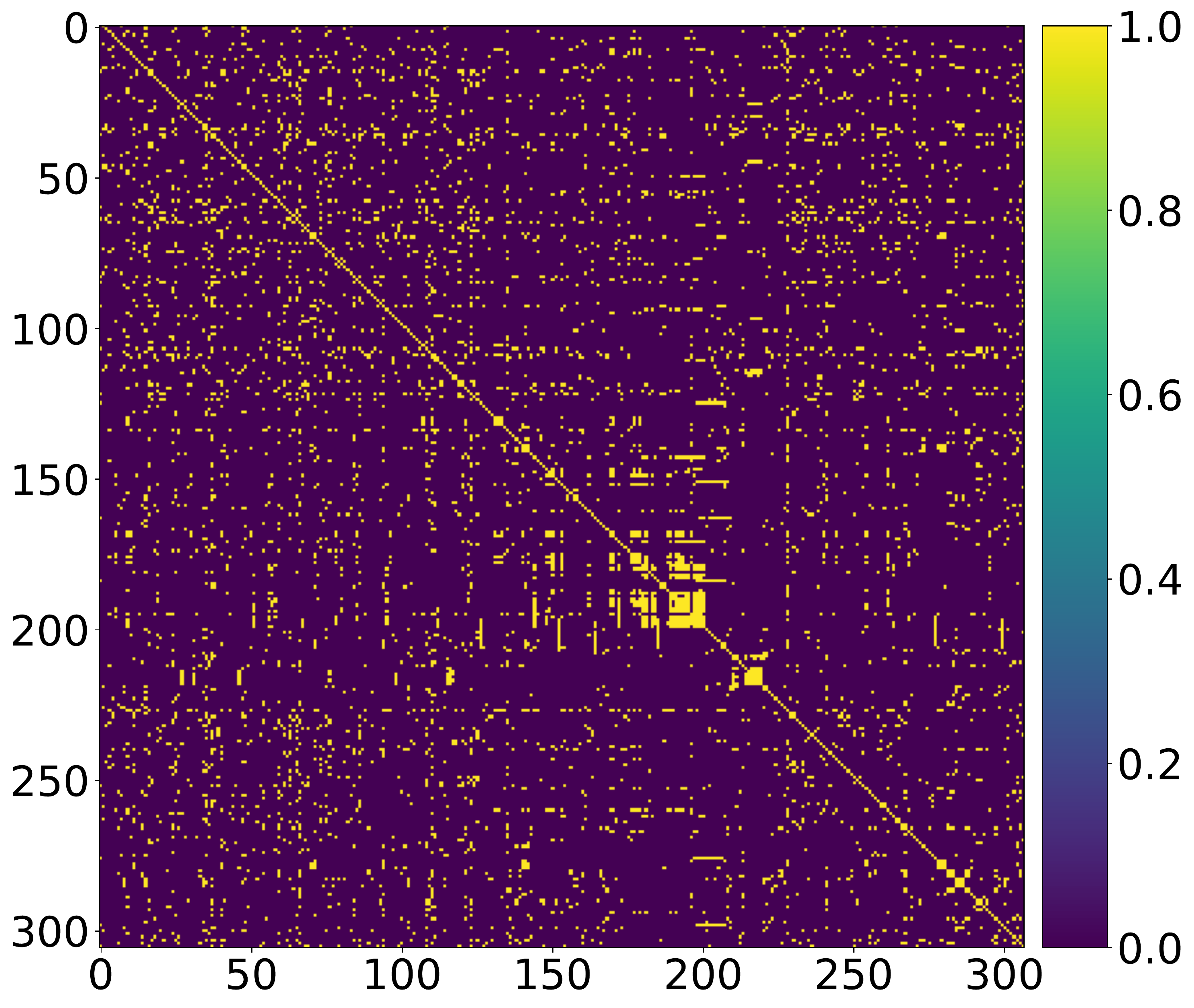}}
    \subfigure[The normalized adjacency matrix]{\includegraphics[width=0.3\textwidth]{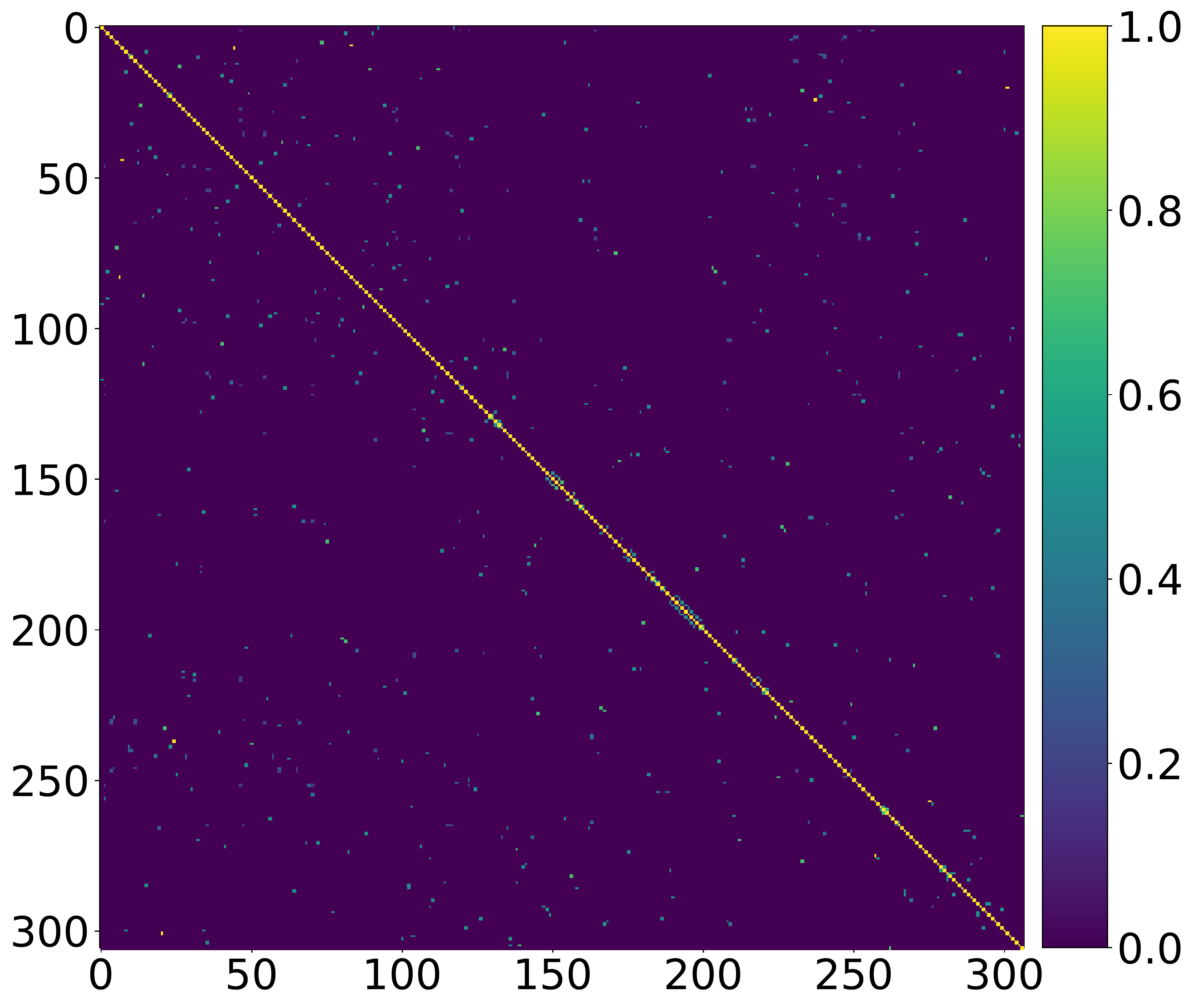}}
    \subfigure[The learned adjacency matrix]{\includegraphics[width=0.3\textwidth]{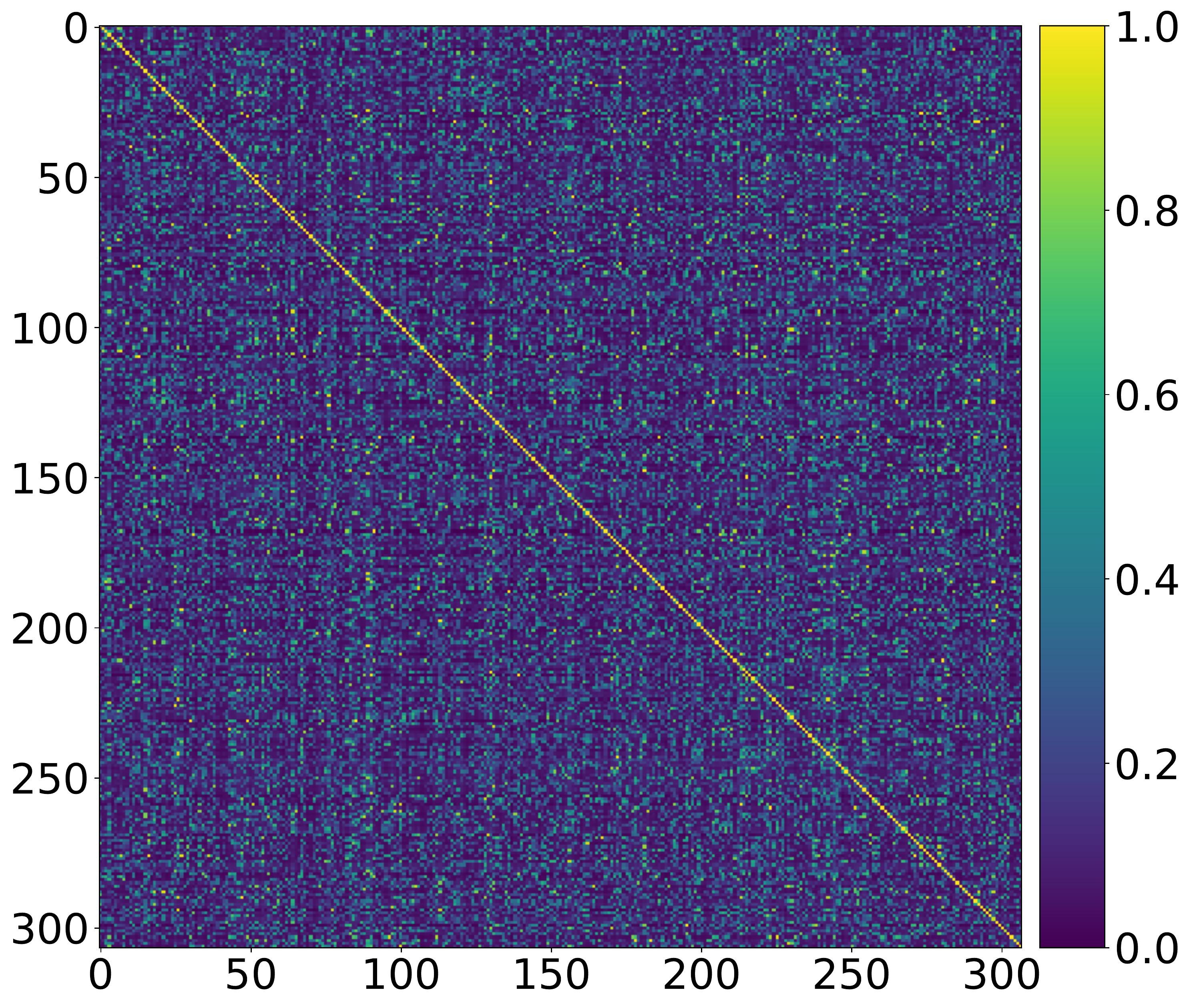}}
    % \subfigure[The histogram of the original adjacency matrix]{\includegraphics[width=0.3\textwidth]{img/adj/hist_original_adj.pdf}}
    % \subfigure[The histogram of the learned adjacency matrix]{\includegraphics[width=0.3\textwidth]{img/adj/hist_learned_adj.pdf}}
    \caption{Adjacency matrix comparison on PeMSD4}
    \label{fig:adj}
\end{figure}

We compare the learned adjacency matrix, the adjacency matrix, and the normalized adjacency matrix given by data. For instance, we show the two adjacency matrices for PeMSD4 in Fig.~\ref{fig:adj}. We note that the original adjacency matrix is a binary matrix whereas the learned matrix's elements range in $[0,1]$. The normalized adjacency matrix is the most commonly used form of adjacency matrix in GNNs e.g., GCN. As shown, the learned adjacency matrix has much more edges than the original and normalized one --- however, many of them are weak edges whose edge weights are around 0.1. There exist many latent weak edges in PeMSD4 that are missing in its original adjacency matrix and our learned adjacency matrix successfully found them.

\begin{figure}[t]
    \centering
    \subfigure[MAE on PeMSD3]{\includegraphics[width=0.24\textwidth]{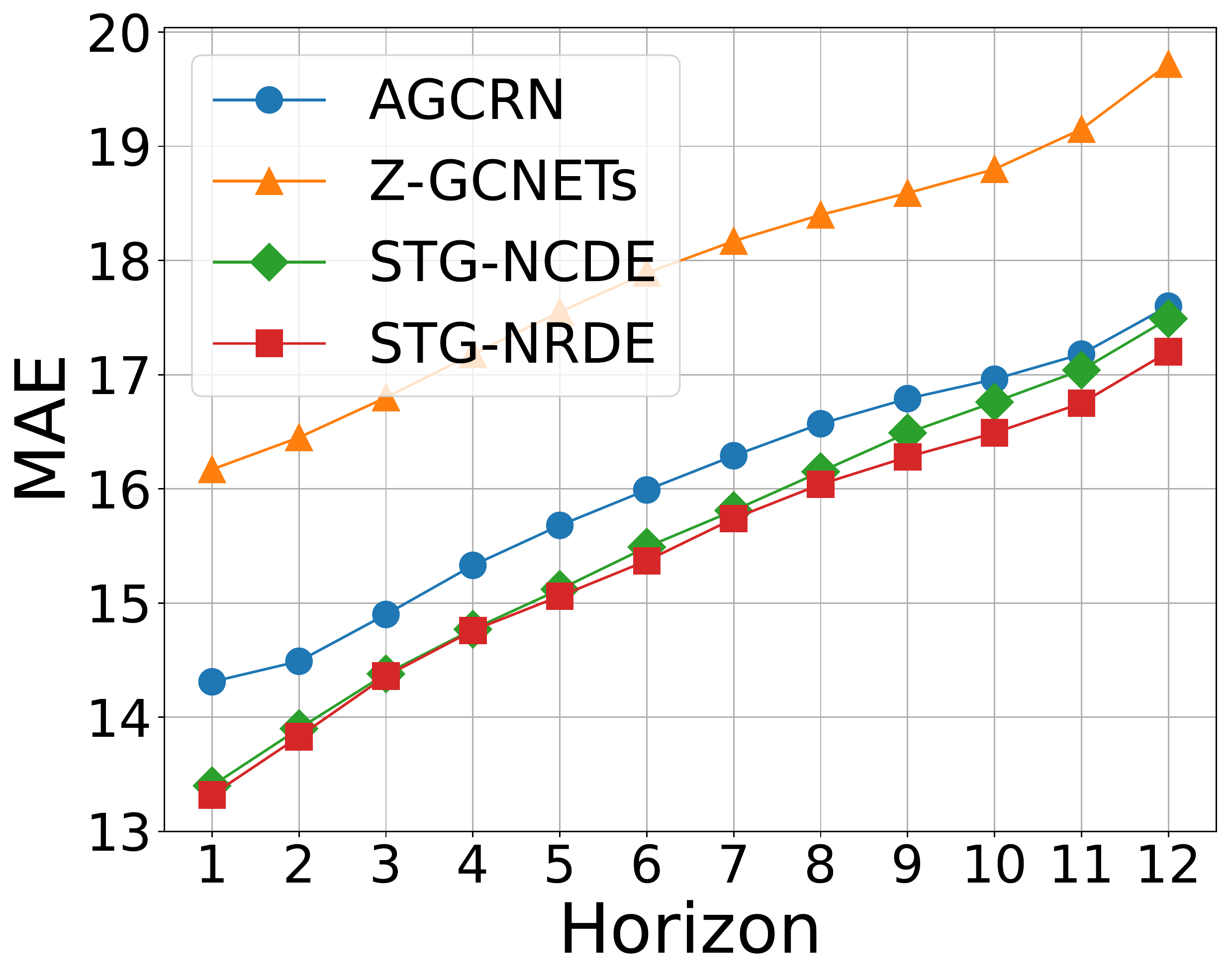}}
    % \subfigure[RMSE on PeMSD3]{\includegraphics[width=0.24\textwidth]{img/horizons/pemsd3_rmse.pdf}}
    \subfigure[MAE on PeMSD4]{\includegraphics[width=0.24\textwidth]{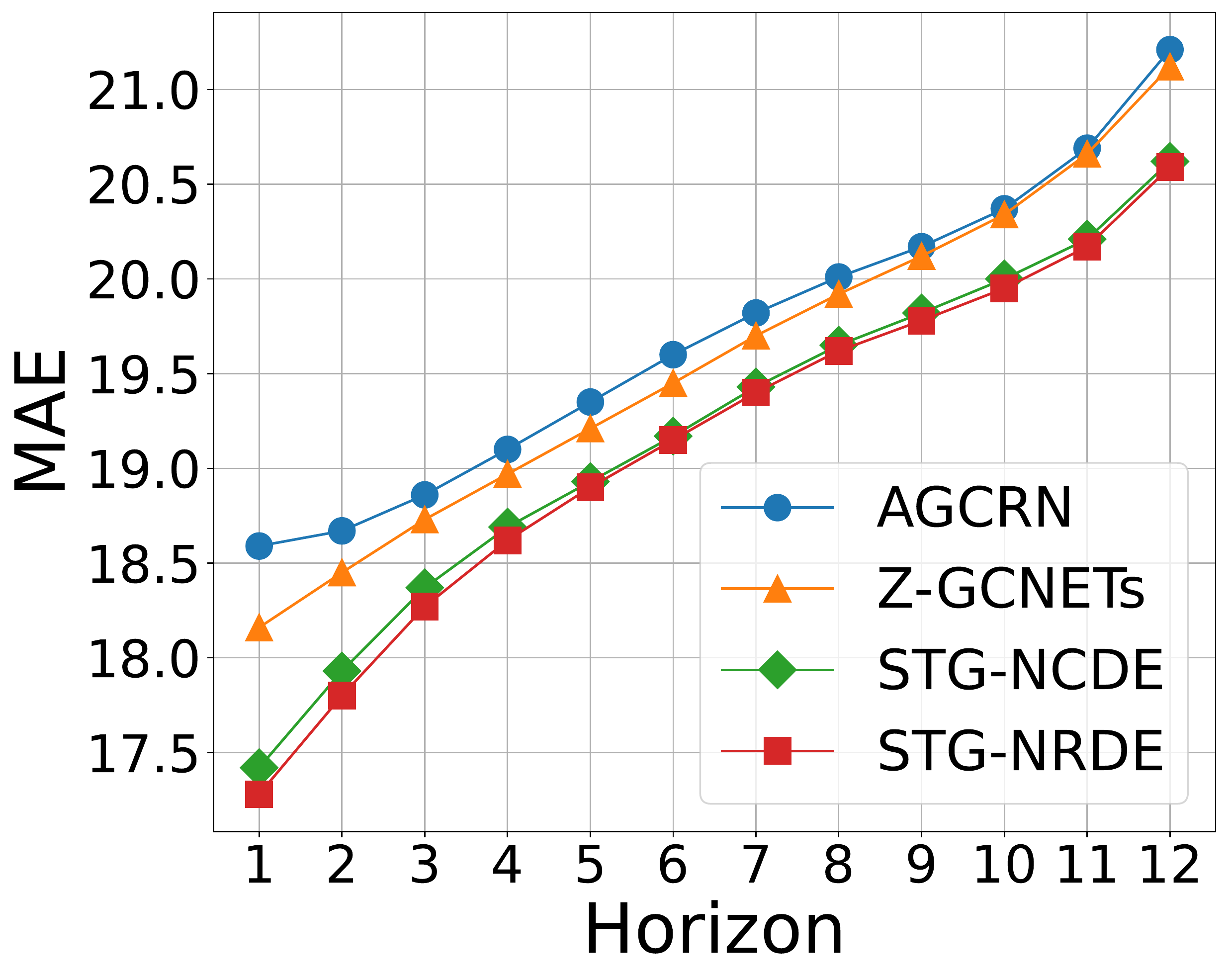}}
    % \subfigure[RMSE on PeMSD4]{\includegraphics[width=0.24\textwidth]{img/horizons/pemsd4_rmse.pdf}}
    \subfigure[MAE on PeMSD7]{\includegraphics[width=0.24\textwidth]{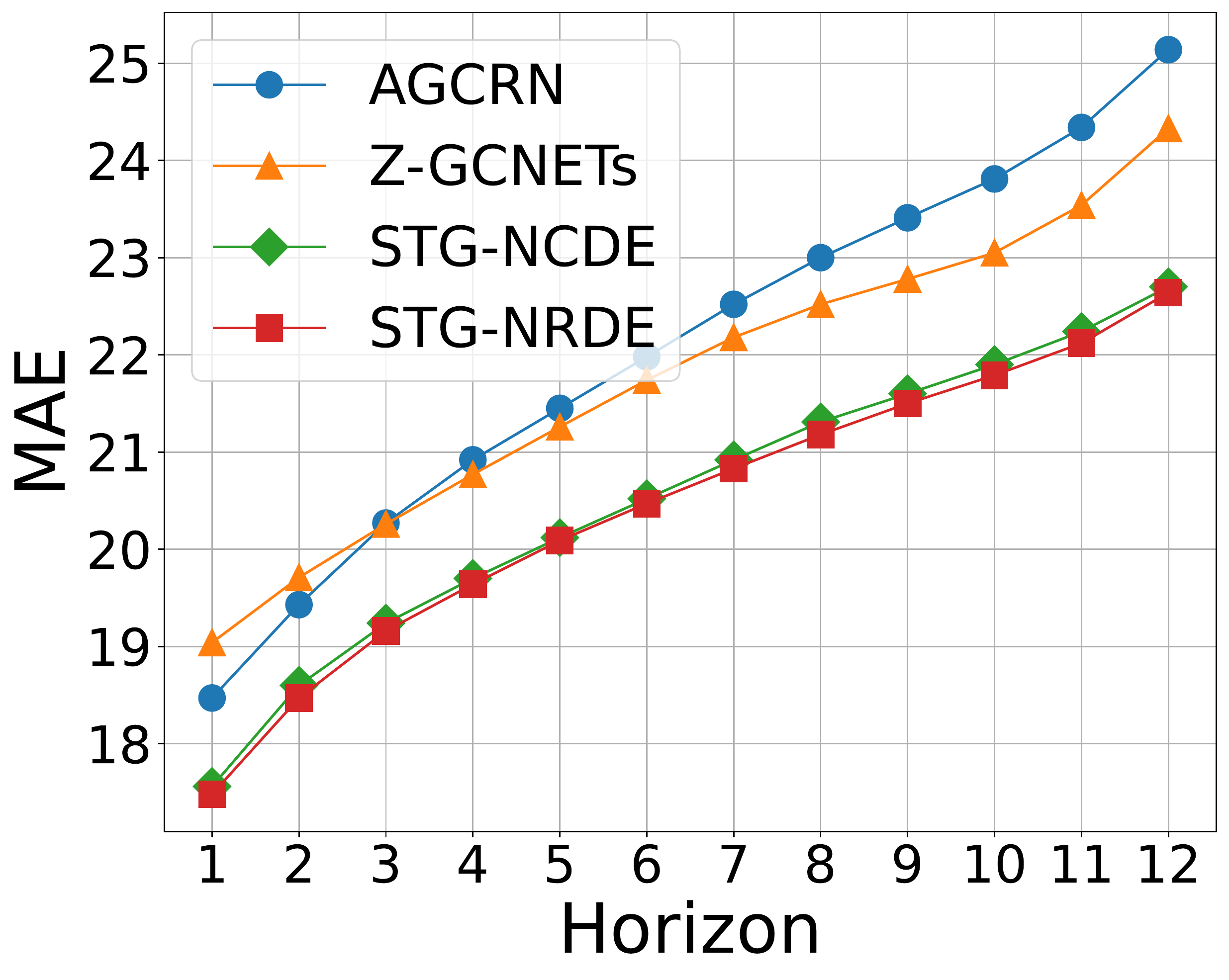}}
    % \subfigure[RMSE on PeMSD7]{\includegraphics[width=0.24\textwidth]{img/horizons/pemsd7_rmse.pdf}}
    \subfigure[MAE on PeMSD8]{\includegraphics[width=0.24\textwidth]{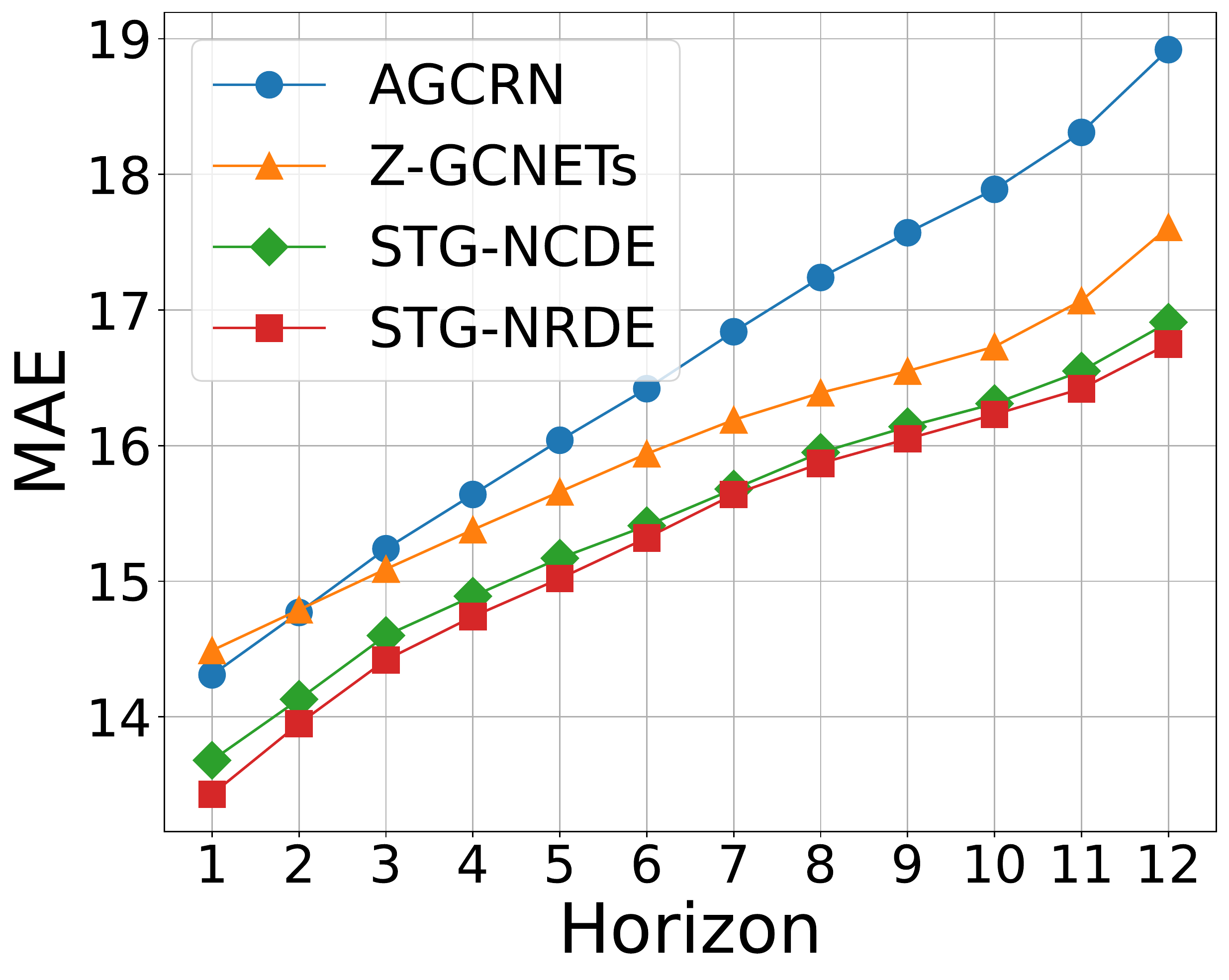}}
    % \subfigure[RMSE on PeMSD8]{\includegraphics[width=0.24\textwidth]{img/horizons/pemsd8_rmse.pdf}}
    \caption{Prediction error at each horizon}
    \label{fig:horizon}
\end{figure}

\subsubsection{Error for Each Horizon} In our notation, $S$ denotes the length of forecasting, i.e., the number of forecasting horizons. Since the benchmark dataset has a setting of $S=12$, we show the model error for each forecasting horizon in Fig.~\ref{fig:horizon}. The error levels show high correlations to $S$. For most horizons, STG-NRDE shows smaller errors than other baselines.

% \begin{figure}[ht!]
%     \centering
%     \includegraphics[width=0.48\textwidth]{img/horizons/pemsd4_long_mae.pdf}
%     \caption{MAEs of each horizon for longer sequences on PeMSD4}
%     \label{fig:long_horizon}
% \end{figure}

\begin{wrapfigure}{r}{0.35\textwidth}
  \begin{center}
    \includegraphics[width=0.35\textwidth]{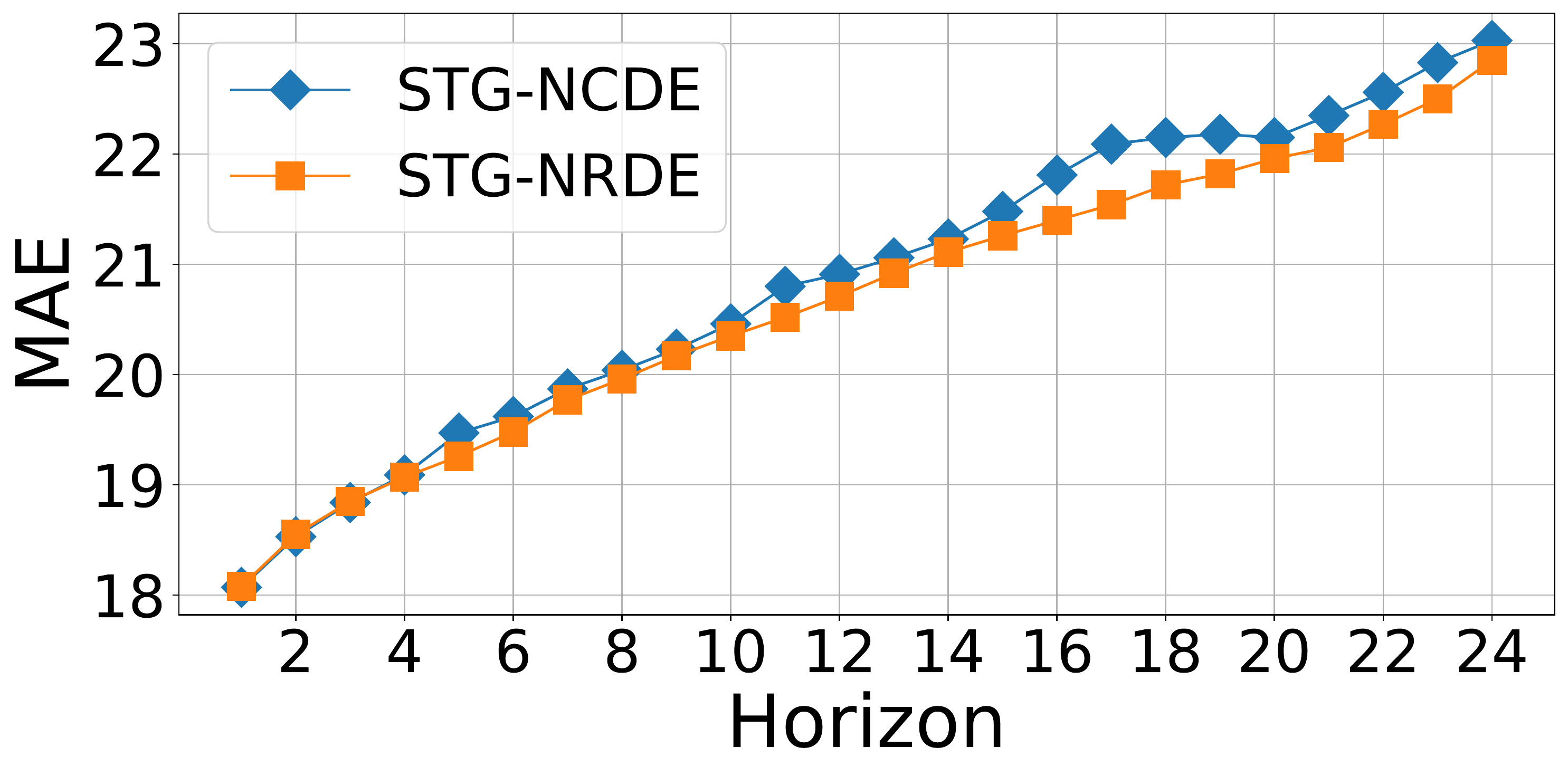}
  \end{center}
    \caption{MAEs of each horizon for longer sequences on PeMSD4}
    \label{fig:long_horizon}
\end{wrapfigure}

\begin{table}[t]
    \small
    \centering
    \caption{Longer sequence forecasting on PeMSD4 and PeMSD8}
    \begin{tabular}{cc ccc ccc ccc ccc}\toprule
        \multirow{2}{*}{Model}   & \multirow{2}{*}{$S$} &\multicolumn{3}{c}{PeMSD4} & \multicolumn{3}{c}{PeMSD8}\\\cmidrule(lr){3-5} \cmidrule(lr){6-8}
                &  & MAE & RMSE & MAPE &  MAE & RMSE & MAPE\\\midrule
        \multirow{2}{*}{STG-NCDE} 
            & 18 & 20.27 & 32.63 & 13.43\% 
                 & 16.75 & 26.78 & 11.42\% \\
            & 24 & 20.86 & 33.51 & 13.72\%
                 & 17.54 & 27.82 & 11.56\% \\\midrule
        \multirow{2}{*}{\textbf{STG-NRDE}}
            & 18 & \textbf{20.22} & \textbf{32.53} & \textbf{13.34\%} 
                 & \textbf{16.71} & \textbf{26.57} & \textbf{11.22\%} \\
            & 24 & \textbf{20.67} & \textbf{33.32} & \textbf{13.69\%}
                 &\textbf{17.48} & \textbf{27.56}  & \textbf{11.53\%}\\\bottomrule
    \end{tabular}
    \label{tab:long}
\end{table}

\subsubsection{Forecasting on Longer Horizons}
We experimented with longer sequence-to-sequence forecasting on PeMSD4 and PeMSD8 with settings $S=18$ and $S=24$. In the case of $S=24$, it reads 24 previous graph snapshots and forecasts the next 24 sequences, which is a longer setting than the standard benchmark setting in the traffic forecasting domain. In Table~\ref{tab:long} and Fig.~\ref{fig:long_horizon}, STG-NRDE shows better results than STG-NCDE. For $S=18$ on PeMSD8, the RMSE of STG-NRDE shows a performance of 26.78, and the STG-NRDE shows a lower error with an error of 26.57. We show that STG-NRDE shows a greater advantage over STG-NCDE in longer sequences, thanks to the log-signature transform of STG-NRDE.

\subsection{Robustness Analysis}
For robustness analysis, we perform time-series cross-validation and blocked cross-validation. In general cross-validation, temporal dependencies must be maintained during testing, so cross-validation in time series is not trivial, so it is performed on two cross-validation methods. For both methods, the number of folds is 4, and the train/valid/test ratio is 6:2:2. We use 5 baseline models, AGCRN, STFGNN, STGODE, Z-GCNETs, and STG-NCDE. The hyperparameter settings of our method and baselines are the same as mentioned in the Section~\ref{subsubsec:hyperparam}.

\begin{table}[t]
    \centering
    \caption{Time-series cross-validation on PeMSD3, PeMSD4, PeMSD7 and PeMSD8}
    \label{tab:time_cv}
    \setlength{\tabcolsep}{1pt}
    \resizebox{\textwidth}{!}{
    \begin{tabular}{c ccc ccc ccc ccc}
        \toprule
        \multirow{2}{*}{Model}  & \multicolumn{3}{c}{PeMSD3}    & \multicolumn{3}{c}{PeMSD4}      & \multicolumn{3}{c}{PeMSD7}      & \multicolumn{3}{c}{PeMSD8}\\\cmidrule(lr){2-4} \cmidrule(lr){5-7} \cmidrule(lr){8-10} \cmidrule(lr){11-13}
                                & MAE & RMSE & MAPE(\%) & MAE & RMSE & MAPE(\%) & MAE & RMSE & MAPE(\%) & MAE & RMSE & MAPE(\%) \\ \midrule        
        AGCRN                   
                                & 15.95\std{0.9} & 26.12\std{1.8} & 14.57\std{1.4}   
                                & 20.68\std{1.2} & 34.74\std{3.2} & 13.98\std{1.3}         
                                & 23.62\std{1.6} & 38.31\std{2.8} & 10.03\std{0.8}         
                                & 18.67\std{2.9} & 30.64\std{5.6} & 11.21\std{1.3} \\
        STFGNN                   
                                & 16.86\std{0.9} & 27.52\std{1.9} & 17.06\std{1.5}
                                & 20.80\std{0.7} & 32.86\std{0.7} & 13.39\std{1.0}
                                & 24.35\std{1.1} & 38.30\std{0.8} & 12.42\std{1.2}
                                & 18.05\std{0.8} & 27.20\std{0.5} & 12.09\std{0.5} \\
        STGODE                  
                                & 17.51\std{1.0} & 27.41\std{1.3} & 16.98\std{0.3}
                                & 21.56\std{0.5} & 33.58\std{0.6} & 16.29\std{1.6}         
                                & 23.99\std{0.9} & 37.34\std{0.5} & 11.14\std{0.8}         
                                & 19.21\std{0.9} & 28.54\std{0.2} & 11.65\std{0.7} \\
        Z-GCNETs                  
                                & 16.72\std{1.6} & 30.31\std{8.2} & 14.98\std{1.4}   
                                & 19.84\std{0.4} & 32.51\std{0.8} & 13.62\std{0.8}   
                                & 22.74\std{1.0} & 37.05\std{1.8} & 9.77\std{0.5}         
                                & 16.84\std{0.8} & 27.27\std{1.6} & 10.57\std{0.7} \\

        STG-NCDE                
                                & 15.21\std{0.4} & 24.69\std{1.6} & 14.33\std{0.6}         
                                & 19.39\std{0.5} & 31.56\std{0.4} & 13.09\std{0.3}         
                                & 21.22\std{0.4} & 34.58\std{0.7} & 9.33\std{0.4}          
                                & 16.78\std{1.1} & 27.05\std{1.8} & 10.66\std{0.7} \\
        \midrule 
        \textbf{STG-NRDE}       
                                & \textbf{15.12}\std{0.5} & \textbf{24.55}\std{1.7} & \textbf{13.92}\std{0.7} 
                                & \textbf{19.24}\std{0.1} & \textbf{31.31}\std{0.3} & \textbf{12.95}\std{0.3}   
                                & \textbf{20.85}\std{0.4} & \textbf{34.08}\std{0.4} & \textbf{9.05}\std{0.4}  
                                & \textbf{16.33}\std{1.0} & \textbf{26.53}\std{1.6} & \textbf{9.99}\std{1.1}\\
        \midrule
        \textit{improv.}        & \textit{0.64\%} & \textit{0.58\%} & \textit{2.82\%}
                                & \textit{0.76\%} & \textit{0.78\%} & \textit{1.10\%}
                                & \textit{1.77\%} & \textit{1.45\%} & \textit{2.94\%}
                                & \textit{2.71\%} & \textit{1.94\%} & \textit{6.22\%} \\
        \bottomrule
    \end{tabular}
    }
\end{table}

\begin{table}[t!]
    % \small
    \centering
    \caption{Time-series cross-validation on PeMSD7(M) and PeMSD7(L)}
    \label{tab:time_cv_2}
    \resizebox{0.65\textwidth}{!}{
    \begin{tabular}{c ccc ccc}
        \toprule
        \multirow{2}{*}{Model}  & \multicolumn{3}{c}{PeMSD7(M)}    & \multicolumn{3}{c}{PeMSD7(L)} \\\cmidrule(lr){2-4} \cmidrule(lr){5-7}
                                & MAE & RMSE & MAPE(\%) & MAE & RMSE & MAPE(\%) \\ \midrule        
        AGCRN                         			
                                & 2.95\std{0.1} & 5.87\std{0.2} & 7.27\std{0.2}
                                & 4.24\std{2.0} & 7.84\std{3.1} & 11.65\std{6.8} \\
        STFGNN                        			
                                & 3.01\std{0.2} & 5.99\std{0.3} & 7.55\std{0.2}
                                & 3.95\std{0.8} & 7.75\std{0.7} & 8.06\std{0.5} \\
        STGODE                        			
                                & 3.02\std{0.1} & 5.70\std{0.2} & 7.45\std{0.5}
                                & 3.28\std{0.3} & 6.11\std{0.5} & 7.94\std{0.7} \\
        Z-GCNETs                      			
                                & 3.04\std{0.2} & 5.90\std{0.2} & 7.27\std{0.2}
                                & 3.02\std{0.1} & 5.98\std{0.1} & 7.58\std{0.4} \\
        STG-NCDE                
                                & 2.86\std{0.2} & 5.72\std{0.3} & 7.00\std{0.3}
                                & 3.03\std{0.1} & 5.99\std{0.2} & 7.47\std{0.3} \\
        \midrule 
        \textbf{STG-NRDE}
                                & \textbf{2.72}\std{0.1} & \textbf{5.50}\std{0.2} & \textbf{6.79}\std{0.4} 
                                & \textbf{2.88}\std{0.1} & \textbf{5.75}\std{0.2} & \textbf{7.09}\std{0.5}\\
        \midrule
        \textit{improv.}        & \textit{5.07\%} & \textit{3.76\%} & \textit{2.95\%}
                                & \textit{4.87\%} & \textit{4.13\%} & \textit{5.18\%} \\
        \bottomrule
    \end{tabular}
    }
\end{table}
\subsubsection{Time-series Cross-validation}
The first method, time-series cross-validation, is on a rolling basis. This cross-validation starts with a small subset of data for training purposes, forecasts for the later data points, and then checks the accuracy of the forecasted data points. The same forecasted data points are then included as part of the next training dataset, and subsequent data points are forecasted. In other words, this time-series split divides the training set at each iteration on the condition that the test set is always ahead of the training set.

We report the result of time-series cross-validation with mean and standard deviation for 4 folds in Table~\ref{tab:time_cv} and \ref{tab:time_cv_2}. As a result of time-series cross-validation, we can see a noticeable performance improvement in MAPE. In the case of AGCRN, the error in time-series cross-validation is higher compared to our model. In particular, in the case of PeMSD8, it can be seen that STG-NRDE shows an MAE reduction of 12.53\% and 3.02\%, respectively, compared to AGCRN and ZGCNETs. Surprisingly, STG-NRDE shows a performance improvement of 6.22\% compared to STG-NCDE in PeMSD8. In PeMSD3, the performance improvement of MAE and RMSE of our model compared to STG-NCDE does not exceed 1\%, but MAPE's error drops by 2.82\%. On PeMSD7(M) and PeMSD7(L), the STG-NRDE significantly outperform the STG-NCDE. Since the MAPE errors of all datasets are reduced by more than 1\%, we show that our proposed model is effective through this robustness analysis. However, in the case of MAE and RMSE, PeMSD3 and PeMSD4 show a slight performance improvement, so there may be room for improvement through the exploration of log-signature transformation or advanced spatial dependency processing techniques.

\subsubsection{Blocked Cross-validation}
Unlike the time-series cross-validation method, the blocked cross-validation method fixes the train/valid/test data size at all cross-validation folds. In the time-series cross-validation method, the data size increases as the fold proceeds. For example, as the number of folds increases, the train data increases to include the test set of the previous fold. In this case, stability against errors is not guaranteed. In contrast, in blocked cross-validation, the next fold does not use the previous fold's test set as the train set, and each fold has the same data size. As a result, the blocked cross-validation produces robust error estimates.

In Table~\ref{tab:blocked_cv} and \ref{tab:blocked_cv_2}, we report the result of the blocked cross-validation with mean and standard deviation for 4 folds. For the blocked cross-validation, we can find that the improvement shows a significant gap. From Table~\ref{tab:blocked_cv}, the performance of STG-NRDE is significantly boosted, by 18.40\% to 30.38\% on PeMSD4. In PeMSD8, the MAE of our STG-NRDE is 3.16\% lower than that of STG-NCDE. 
In PeMSD7(L), STG-NRDE reduces the MAE, RMSE, and MAPE by 5.01\%, 3.28\%, and 5.71\%, respectively, over STG-NCDE. The results in Table~\ref{tab:blocked_cv_2} show that our proposed model's performance improvement is clear even in the traffic speed benchmark dataset. Although the improvement of a few datasets in Table~\ref{tab:main_exp_2} is marginal, the performance improvement is evident in the robustness analysis using the blocked cross-validation. These results suggest that the proposed method can be created a hidden vector with better expressiveness than other baselines thanks to the log-signature transform even when the cross-validation is performed using only a portion of the dataset as a training dataset.

\begin{table}[t]
    \centering
    \caption{Blocked cross-validation on PeMSD3, PeMSD4, PeMSD7 and PeMSD8}
    \label{tab:blocked_cv}
    \setlength{\tabcolsep}{1pt}
    \resizebox{\textwidth}{!}{
    \begin{tabular}{c ccc ccc ccc ccc}
        \toprule
        \multirow{2}{*}{Model}  & \multicolumn{3}{c}{PeMSD3}    & \multicolumn{3}{c}{PeMSD4}      & \multicolumn{3}{c}{PeMSD7}      & \multicolumn{3}{c}{PeMSD8}\\\cmidrule(lr){2-4} \cmidrule(lr){5-7} \cmidrule(lr){8-10} \cmidrule(lr){11-13}
                                & MAE & RMSE & MAPE(\%) & MAE & RMSE & MAPE(\%) & MAE & RMSE & MAPE(\%) & MAE & RMSE & MAPE(\%) \\ \midrule        
        AGCRN                   
                                & 16.33\std{1.3} & 27.17\std{2.7} & 14.27\std{0.8}   
                                & 24.88\std{1.8} & 44.26\std{3.1} & 18.23\std{3.3}         
                                & 25.93\std{0.3} & 43.47\std{0.8} & 10.78\std{0.4}
                                & 21.45\std{1.5} & 36.12\std{2.6} & 12.38\std{1.0} \\
        STFGNN                   
                                & 17.86\std{1.2} & 30.51\std{1.1} & 17.96\std{0.5}
                                & 21.12\std{0.8} & 33.20\std{0.9} & 17.49\std{1.0}   
                                & 24.87\std{0.6} & 38.73\std{0.3} & 10.65\std{0.2}
                                & 18.05\std{0.3} & 27.45\std{0.4} & 11.62\std{0.4} \\
        STGODE                   
                                & 17.51\std{1.0} & 27.41\std{1.3} & 16.98\std{0.3}   
                                & 21.56\std{0.5} & 33.58\std{0.6} & 16.29\std{1.6}   
                                & 23.99\std{0.9} & 37.34\std{0.5} & 11.14\std{0.0}         
                                & 18.11\std{0.9} & 28.54\std{1.7} & 11.67\std{0.7} \\
        Z-GCNETs                  
                                & 16.35\std{0.9} & 25.91\std{1.5} & 15.01\std{0.9}
                                & 21.40\std{0.7} & 35.29\std{1.1} & 15.18\std{1.5}
                                & 23.70\std{0.5} & 39.22\std{0.5} &  9.93\std{0.5}
                                & 17.38\std{0.5} & 28.17\std{0.9} & 11.09\std{0.6} \\
        STG-NCDE              
                                & 15.62\std{0.8} & 24.68\std{1.3} & 14.77\std{0.8}         
                                & 20.44\std{0.7} & 33.01\std{0.9} & 15.78\std{1.0}         
                                & 21.90\std{0.2} & 35.94\std{1.2} & 9.46\std{0.3} 
                                & 17.23\std{0.8} & 27.45\std{1.3} & 11.35\std{0.4} \\
        \midrule 
        \textbf{STG-NRDE}       
                                & \textbf{15.57}\std{0.9} & \textbf{24.55}\std{1.2} & \textbf{14.66}\std{0.8} 
                                & \textbf{16.68}\std{0.9} & \textbf{26.77}\std{1.6} & \textbf{10.98}\std{0.5}   
                                & \textbf{21.56}\std{0.3} & \textbf{35.15}\std{1.1} & \textbf{9.44}\std{0.3}  
                                & \textbf{16.68}\std{0.9} & \textbf{26.77}\std{1.6} & \textbf{10.98}\std{0.5}\\
        \midrule
        \textit{improv.}        & \textit{0.37\%} & \textit{0.54\%} & \textit{0.75\%}
                                & \textit{18.40\%} & \textit{18.92\%} & \textit{30.38\%}
                                & \textit{1.52\%} & \textit{2.22\%} & \textit{0.24\%}
                                & \textit{3.16\%} & \textit{2.48\%} & \textit{3.23\%} \\
        \bottomrule
    \end{tabular}
    }
\end{table}

\begin{table}[t]
    % \small
    \centering
    \caption{Blocked cross-validation on PeMSD7(M) and PeMSD7(L)}
    \label{tab:blocked_cv_2}
    \resizebox{0.65\textwidth}{!}{
    \begin{tabular}{c ccc ccc}
        \toprule
        \multirow{2}{*}{Model}  & \multicolumn{3}{c}{PeMSD7(M)}    & \multicolumn{3}{c}{PeMSD7(L)} \\\cmidrule(lr){2-4} \cmidrule(lr){5-7}
                                & MAE & RMSE & MAPE(\%) & MAE & RMSE & MAPE(\%) \\ \midrule        
        AGCRN                         			
                                & 3.10\std{0.1} & 6.16\std{0.1} & 7.93\std{0.5}         
                                & 7.18\std{2.3} & 12.32\std{3.4} & 22.77\std{8.7} \\
        STFGNN                         			
                                & 2.98\std{0.3} & 5.84\std{0.2} & 7.52\std{0.4}
                                & 3.20\std{0.2} & 6.25\std{0.2} & 8.15\std{0.5} \\
        STGODE                        			
                                & 3.02\std{0.1} & 5.86\std{0.2} & 7.55\std{0.5}
                                & 3.28\std{0.3} & 6.11\std{0.5} & 7.94\std{0.7} \\
        Z-GCNETs                      			
                                & 3.16\std{0.1} & 6.18\std{0.1} & 8.04\std{0.5}
                                & 3.28\std{0.1} & 6.44\std{0.3} & 8.38\std{0.5} \\
        STG-NCDE                
                                & 2.95\std{0.1} & 5.84\std{0.2} & 7.48\std{0.5}
                                & 3.15\std{0.1} & 6.22\std{0.3} & 8.01\std{0.6} \\
        \midrule 
        \textbf{STG-NRDE}       
                                & \textbf{2.89}\std{0.1} & \textbf{5.81}\std{0.2} & \textbf{7.25}\std{0.1} 
                                & \textbf{3.00}\std{0.2} & \textbf{6.02}\std{0.4} & \textbf{7.57}\std{0.8}\\
        \midrule
        \textit{improv.}        & \textit{2.08\%} & \textit{0.52\%} & \textit{3.17\%}
                                & \textit{5.01\%} & \textit{3.28\%} & \textit{5.71\%} \\
        \bottomrule
    \end{tabular}
    }
\end{table}

\begin{table}[t]
    \small
    \centering
    \caption{Training time (seconds per epoch). In addition, Z-GCNETs' preprocessing requires more than an hour. Our method's preprocessing with the natural cubic spline algorithm and the rough signature transform take less than a second.}
    \label{tab:time}
    \begin{tabular}{ccccccc}
        \hline
        Model                    & PeMSD3 & PeMSD4 & PeMSD7 & PeMSD8 & PeMSD7(M) & PeMSD7(L)\\\hline
        AGCRN                    &  23.65 &  13.44 &  63.80 &   8.80 &      7.52 &     35.15\\
        % STFGNN                   & 000.00 & 000.00 & 000.00 & 000.00 &    000.00 &     00000\\
        STGODE                   &  79.90 &  45.14 & 190.73 &  32.38 &     28.04 &    100.20\\
        Z-GCNETs                 & 219.65 &  28.66 & 114.90 & 147.47 &     17.99 &     61.25\\
        STG-NCDE                 & 185.85 & 118.64 & 131.70 &  62.97 &     37.16 &    153.66\\\hline
        \textbf{STG-NRDE}        & 170.20 & 118.64 & 125.70 &  54.34 &     35.53 &    140.15\\
        % \textbf{Only temporal}   &  29.30 &  17.91 &  70.48 &  11.09 &      7.85 &     21.27\\
        % \textbf{Only spatial}    &  43.23 &  25.58 & 106.26 &  33.76 &     13.78 &     44.37\\
        \hline
    \end{tabular}

\end{table}

\begin{table}[t]
    \small
    \centering
    \caption{Model size in terms of the number of parameters}
    \label{tab:size}
    \begin{tabular}{ccccccc}
        \hline
        Model                   & PeMSD3  & PeMSD4  & PeMSD7  & PeMSD8  & PeMSD7(M) & PeMSD7(L)\\\hline
        AGCRN                   & 749,320 & 748,810 & 151,538 & 150,112 &   150,228 &   151,824\\
        % STFGNN                  & 000.00  & 000.00  & 000.00  & 000.00  &    000.00 &     00000\\
        STGODE                  & 714,036 & 714,504 & 732,936 & 709,572 &   709,356 &   738,084\\
        Z-GCNETs                & 455,544 & 455,034 & 460,794 & 453,664 &   454,244 &   462,224\\
        STG-NCDE                & 373,764 & 376,904 & 388,424 & 560,436 &   162,652 &   178,612\\\hline
        \textbf{STG-NRDE}       & 314,852 & 381,192 & 388,552 & 700,392 &   162,652 &   195,636\\
        % \textbf{Only temporal}  &  17,032 &  20,682 &  26,442 &  25,728 &     9,044 &    17,024\\
        % \textbf{Only spatial}   & 103,172 & 376,904 & 113,672 & 288,724 &    93,884 &   109,844\\
        \hline
    \end{tabular}
\end{table}

\subsection{Model Efficency Analysis}
We report the training time of various methods in Table~\ref{tab:time} --- we also report the number of parameters in Table~\ref{tab:size}. In general, the training time of our method is sometimes longer than that of baseline models. We attribute this to the fact that it is still difficult to compute the gradient for the NRDE and solve the integration problem. However, since our model can reduce the integration time due to the rough path signature of NRDE, it can be confirmed that the training time is reduced more than that of STG-NCDE. In addition, it takes less than a second to run the natural cubic spline algorithm and the rough signature transform for our method. However, Z-GCNETs requires more than an hour for extracting time-aware zigzag persistence from data before starting training.

In the case of the number of parameters, the number of parameters slightly increases compared to STG-NCDE because the number of features increases as the depth of the rough signature deepens. However, in the case of PeMSD3, it has the advantage of being able to reduce the size of other hidden vectors as it is converted into a rough signature. In addition, in the case of PeMSD3, less than half the number of parameters can be used compared to the baselines of AGCRN and STGODE.

%%%%%%%%%%%%%%%%%%%%%%%%%%%%%%%%%%%%%%%%%%%%%%%%%%%%
%%%%%%%%%%%%%%%%%%%%  SECTION   %%%%%%%%%%%%%%%%%%%%
%%%%%%%%%%%%%%%%%%%%%%%%%%%%%%%%%%%%%%%%%%%%%%%%%%%%
\section{Conclusions}
\noindent We presented a spatio-temporal NRDE model to perform traffic forecasting. Our model has two NRDEs: one for temporal processing and the other for spatial processing. In particular, our NRDE for spatial processing can be considered as an NRDE-based interpretation of graph convolutional networks. In our experiments with 6 datasets and 27 baselines, our method clearly shows the best overall accuracy. In addition, our model can perform irregular traffic forecasting where some input observations can be missing, which is a practical problem setting but not actively considered by existing methods. We believe that the combination of NRDEs and GCNs is a promising research direction for spatio-temporal processing.

\begin{acks}
This work was supported by the Yonsei University Research Fund of 2022, and the Institute of Information \& Communications Technology Planning \& Evaluation (IITP) grant funded by the Korean government (MSIT) (No. 2020-0-01361, Artificial Intelligence Graduate School Program (Yonsei University), and No. 2021-0-00155, Context and Activity Analysis-based Solution for Safe Childcare).
\end{acks}

\bibliographystyle{ACM-Reference-Format}
\bibliography{ref}

\clearpage
\appendix

\section{Additional Details for Experiments}\label{app:detail}
\subsection{Baseline Implementations}
We used the official implementation released by the authors on GitHub for all baselines besides GRU-ED, TCN and etc. 
\begin{itemize}
    \item STG-NCDE: \url{https://github.com/jeongwhanchoi/STG-NCDE}
    \item Z-GCNETs: \url{https://github.com/Z-GCNETs/Z-GCNETs}
    \item AGCRN: \url{https://github.com/LeiBAI/AGCRN}
    \item DSTAGNN: \url{https://github.com/SYLan2019/DSTAGNN}
    \item STG-ODE: \url{https://github.com/square-coder/STGODE}
    \item STSGCN: \url{https://github.com/Davidham3/STSGCN}
    \item STFGNN: \url{https://github.com/MengzhangLI/STFGNN}
    \item LSGCN: \url{https://github.com/helanzhu/LSGCN}
    \item ASTGCN: \url{https://github.com/guoshnBJTU/ASTGCN-r-pytorch}
    \item MSTGCN: \url{https://github.com/guoshnBJTU/ASTGCN-r-pytorch}
    \item STG2Seq: \url{https://github.com/LeiBAI/STG2Seq}
    \item DSANet: \url{https://github.com/bighuang624/DSANet}
\end{itemize}

\subsection{Tuning the baseline models}
We perform hyperparameter tuning on some unknown accuracy of baselines. So, for each method, its performance on different benchmarks can be reported from different hyperparameters. We test the following command-line arguments for each baseline method:

\begin{itemize}
    \item STG-NCDE:
        \begin{itemize}
            \item \texttt{epoch}: 200
            \item \texttt{batch\_size}: 64
            \item \texttt{lr}: \{0.005, 0.004, 0.003, 0.002, 0.001, 0.0008, 0.0005, 0\}
            \item \texttt{weight\_decay}: \{0.03, 0.02, 0.01, 0.005\}
            \item \texttt{embed\_dim}: \{1, 2, 3, 4, 5, 6, 7, 8, 9, 10\}
            \item \texttt{agc\_iter}: \{1, 2, 3, 4\}
            \item \texttt{hid\_dim}: \{32, 64, 128, 256\}
            \item \texttt{hid\_hid\_dim}: \{32, 64, 128, 256\}
        \end{itemize}
    \item Z-GCNETs:
        \begin{itemize}
            \item \texttt{epoch}: 350
            \item \texttt{batch\_size}: 64
            \item \texttt{lr}: \{0.005, 0.004, 0.003, 0.002, 0.001, 0.0008, 0.0005, 0\}
            \item \texttt{weight\_decay}: \{0.03, 0.02, 0.01, 0.005\}
            \item \texttt{embed\_dim}: \{1, 2, 3, 4, 5, 6, 7, 8, 9, 10\}
            \item \texttt{cheb\_order}: \{1, 2, 3, 4\}
            \item \texttt{rnn\_units}: \{32, 64, 128, 256\}
            \item \texttt{number\_mixture}: 2
            \item \texttt{link\_len}: 2
        \end{itemize}
    \item AGCRN:
        \begin{itemize}
            \item \texttt{epoch}: 200
            \item \texttt{batch\_size}: 64
            \item \texttt{lr}: \{0.005, 0.004, 0.003, 0.002, 0.001, 0.0008, 0.0005, 0\}
            \item \texttt{weight\_decay}: \{0.03, 0.02, 0.01, 0.005\}
            \item \texttt{embed\_dim}: \{1, 2, 3, 4, 5, 6, 7, 8, 9, 10\}
            \item \texttt{cheb\_order}: \{1, 2, 3, 4\}
            \item \texttt{rnn\_units}: \{32, 64, 128, 256\}
        \end{itemize}
    \item DSTAGNN:
        \begin{itemize}
            \item \texttt{epoch}: 100
            \item \texttt{batch\_size}: \{12, 32, 64\}
            \item \texttt{lr}: \{0.001, 0.0005, 0.0001\}
            \item \texttt{d\_model}: 512
            \item \texttt{nb\_block}: 4
            \item \texttt{n\_head}: \{3,4\}
            \item \texttt{thres1}: 0.6
            \item \texttt{thres2}: 0.5
            \item \texttt{in\_channels}: \{32, 64\}
            \item \texttt{out\_channels}: \{32, 64\}
            \item \texttt{num\_layer}: \{1, 2, 3\}
        \end{itemize}
    \item STG-ODE:
        \begin{itemize}
            \item \texttt{epoch}: 200
            \item \texttt{batch\_size}: 16
            \item \texttt{lr}: \{0.005, 0.002, 0.001, 0.0005\}
            \item \texttt{weight\_decay}: \{0.03, 0.02, 0.01, 0.005, 0\}
            \item \texttt{embed\_dim}: \{1, 2, 3, 4, 5, 6, 7, 8, 9, 10\}
            \item \texttt{sigma1}: 0.1
            \item \texttt{sigma2}: 10
            \item \texttt{thres1}: 0.6
            \item \texttt{thres2}: 0.5
            \item \texttt{in\_channels}: \{32, 64\}
            \item \texttt{out\_channels}: \{32, 64\}
            \item \texttt{num\_layer}: \{1, 2, 3\}
        \end{itemize}
    \item STSGNN:
        \begin{itemize}
            \item \texttt{epoch}: 200
            \item \texttt{batch\_size}: 32
            \item \texttt{lr}: \{0.005, 0.002, 0.001, 0.0005\}
            \item \texttt{first\_layer\_embedding\_size}: \{32, 64\}
            \item \texttt{use\_mask}: \{True, False\}
            \item \texttt{temporal\_emb}: \{True, False\}
            \item \texttt{spatial\_emb}: \{True, False\}
            \item \texttt{ctx}: \{0, 1\}
            \item \texttt{max\_update\_factor}: 1
        \end{itemize}
    \item STFGNN:
        \begin{itemize}
            \item \texttt{epoch}: 200
            \item \texttt{batch\_size}: 32
            \item \texttt{lr}: \{0.005, 0.002, 0.001, 0.0005\}
            \item \texttt{first\_layer\_embedding\_size}: \{32, 64\}
            \item \texttt{ctx}: \{0, 1, 2, 3\}
            \item \texttt{max\_update\_factor}: 1
        \end{itemize}
    \item ASTGCN:
        \begin{itemize}
            \item \texttt{epoch}: 80
            \item \texttt{batch\_size}: 32
            \item \texttt{lr}: \{0.005, 0.002, 0.001, 0.0005\}
            \item \texttt{nb\_block}: \{1, 2, 3\}
            \item \texttt{nb\_chev\_filter}: \{32, 64, 128\}
            \item \texttt{nb\_time\_filter}: \{32, 64, 128\}
            \item \texttt{time\_strides}: 3
            \item \texttt{K}: \{1,2,3,4\}
        \end{itemize}
    \item MSTGCN:
        \begin{itemize}
            \item \texttt{epoch}: 80
            \item \texttt{batch\_size}: 32
            \item \texttt{lr}: \{0.005, 0.002, 0.001, 0.0005\}
            \item \texttt{nb\_block}: \{1, 2, 3\}
            \item \texttt{nb\_chev\_filter}: \{32, 64, 128\}
            \item \texttt{nb\_time\_filter}: \{32, 64, 128\}
            \item \texttt{time\_strides}: 3
            \item \texttt{K}: \{1,2,3,4\}
        \end{itemize}
    \item LSGCN:
        \begin{itemize}
            \item \texttt{epoch}: 100
            \item \texttt{batch\_size}: 32
            \item \texttt{lr}: \{0.005, 0.002, 0.001, 0.0005\}
            \item \texttt{C1}: 64
            \item \texttt{C2}: 64
            \item \texttt{ks}: 3
            \item \texttt{kt}: 3
        \end{itemize}
    \item STG2Seq:
        \begin{itemize}
            \item \texttt{epoch}: 400
            \item \texttt{batch\_size}: 64
            \item \texttt{lr}: \{0.005, 0.002, 0.001, 0.0008, 0.0005\}
            \item \texttt{num\_units}: \{32, 64, 128\}
            \item \texttt{num\_layers}: \{2, 3, 4, 5\}
            \item \texttt{delta}: 0.5
            \item \texttt{beta1}: 0.8
            \item \texttt{beta2}: 0.999
        \end{itemize}
    \item GRU-ED:
        \begin{itemize}
            \item \texttt{epoch}: 200
            \item \texttt{batch\_size}: 64
            \item \texttt{lr}: \{0.005, 0.002, 0.001, 0.0008, 0.0005\}
            \item \texttt{weight\_decay}: \{0.001, 0.002, 0.005\}
            \item \texttt{enc\_dim}: \{32, 64, 128, 256\}
            \item \texttt{dec\_dim}: \{32, 64, 128, 256\}
            \item \texttt{n\_layers}: \{1, 2, 3, 4\}
            \item \texttt{drop\_prob}: \{0, 0.1, 0.2, 0.3, 0.4, 0.5\}
        \end{itemize}
    \item TCN:
        \begin{itemize}
            \item \texttt{epoch}: 200
            \item \texttt{batch\_size}: 64
            \item \texttt{lr}: \{0.005, 0.002, 0.001, 0.0008, 0.0005\}
            \item \texttt{weight\_decay}: \{0.001, 0.002, 0.005\}
            \item \texttt{num\_channels}: \{32, 64, 128\}
            \item \texttt{kernel\_size}: 2
            \item \texttt{drop\_prob}: \{0, 0.1, 0.2, 0.3, 0.4, 0.5\}
        \end{itemize}
    \item DSANet:
        \begin{itemize}
            \item \texttt{epoch}: 200
            \item \texttt{batch\_size}: 64
            \item \texttt{lr}: \{0.005, 0.002, 0.001, 0.0008, 0.0005\}
            \item \texttt{weight\_decay}: \{0.001, 0.002, 0.005\}
            \item \texttt{d\_model}: \{32, 64, 128\}
            \item \texttt{d\_k}: 8
            \item \texttt{d\_v}: 8
            \item \texttt{n\_head}: \{2, 4, 6, 8\}
            \item \texttt{d\_inner}: \{32, 64, 128\}
            \item \texttt{n\_layers}: \{1, 2, 3, 4\}
            \item \texttt{drop\_prob}: \{0, 0.1, 0.2, 0.3, 0.4, 0.5\}
        \end{itemize}
\end{itemize}

\end{document}